\documentclass[lettersize,journal]{IEEEtran}
\usepackage{amsmath,amsfonts}
\usepackage{algorithmic}
\usepackage{algorithm}
\usepackage{array}
\usepackage[caption=false,labelfont=sf,textfont=sf]{subfig}
\usepackage{textcomp}
\usepackage{stfloats}
\usepackage{url}
\usepackage{verbatim}
\usepackage{graphicx}
\graphicspath{{figures/}}
\usepackage{cite}
\usepackage{tcolorbox}
\usepackage{subcaption}
\usepackage{caption}

\usepackage{amssymb}
\usepackage{amsthm}
\usepackage{amsmath}
\usepackage{enumerate}
\usepackage{amsfonts}
\usepackage{mathrsfs}
\usepackage{bm}
\usepackage{multirow}
\usepackage{threeparttable}
\usepackage{siunitx}
\usepackage{tikz}
\usetikzlibrary{arrows.meta,positioning,fit}

\usepackage{nomencl}
\usepackage{booktabs}
\theoremstyle{plain}

\newtheorem{lem}{Lemma}
\newtheorem{thm}{Theorem}
\newtheorem{assum}{Assumption}
\newtheorem{rem}{Remark}
\newtheorem{prop}{Proposition}
\newtheorem{defn}{Definition}
\newtheorem{cor}{Corollary}

\AtBeginDocument{%
  \setlength{\abovedisplayskip}{5pt plus 2pt minus 2pt}%
  \setlength{\abovedisplayshortskip}{2pt plus 1pt}%
  \setlength{\belowdisplayshortskip}{3pt plus 2pt minus 2pt}%
  \setlength{\belowdisplayskip}{5pt plus 2pt minus 2pt}%
}

\begin{document}

\title{Decentralized Nonconvex Composite Federated Learning with Gradient Tracking and Momentum}

\author{Yuan Zhou, Xinli Shi, \IEEEmembership{Senior Member, IEEE}, Xuelong Li, \IEEEmembership{Fellow, IEEE}, Jiachen Zhong,  \\
Guanghui Wen, \IEEEmembership{Senior Member, IEEE}, and Jinde Cao, \IEEEmembership{Fellow, IEEE}
\thanks{
This work was supported in part by the National Key Research and Development Project of China under Grant 2020YFA0714300; in part by the National Natural Science Foundation of China under Grant 62473098; in part by the SEU Innovation Capability Enhancement Plan for Doctoral Students under Grant CXJH\_SEU\_25243.
}
\thanks{Yuan Zhou, Xinli Shi and Jiachen Zhong are with the School of Cyber Science \& Engineering, Southeast University, Nanjing 210096, China (e-mail: zhouxyz@seu.edu.cn; xinli\_shi@seu.edu.cn; jiachen\_zhong@seu.edu.cn).}
\thanks{Xuelong Li is with the Institute of Artificial Intelligence (TeleAI) of China Telecom (e-mail: xuelong\_li@ieee.org).}
\thanks{Guanghui Wen is with the School of Mathematics, Southeast University, Nanjing 210096, China (e-mail: wenguanghui@gmail.com).}
\thanks{Jinde Cao is with the School of Mathematics, Southeast University, Nanjing 210096, China, and also with the Purple Mountain Laboratories, Nanjing 211111, China (email: jdcao@seu.edu.cn).}
}



\maketitle

\begin{abstract}
Decentralized Federated Learning (DFL) enables collaborative model training without relying on a central server.
When local objectives are nonconvex and coupled with nonsmooth weakly convex regularization, DFL gives rise to a challenging decentralized nonconvex composite optimization problem involving data heterogeneity, stochastic-gradient noise, and consensus error.
We propose DEPOSITUM, a decentralized composite optimization algorithm for this problem.
DEPOSITUM maintains momentum-filtered stochastic gradient estimates via a tracking mechanism, which accommodates both Polyak and Nesterov momentum.
It further allows multiple local updates between communication rounds to improve communication efficiency.
Theoretical analysis demonstrates that it achieves an expected $\epsilon$-stationary point with an iteration complexity of $\mathcal{O}(1/\epsilon^2)$ without imposing bounded gradient heterogeneity or mean-squared smoothness assumptions.
With an appropriate stepsize and momentum schedule, the averaged stationarity measure further achieves a rate of \(\mathcal{O}(1/\sqrt{nT})\) after a network-dependent transient, using a mini-batch size independent of $T$.
Experiments on different benchmark datasets validate the effectiveness of DEPOSITUM and demonstrate competitive performance against representative federated composite optimization methods.
\end{abstract}

\begin{IEEEkeywords}
Decentralized federated learning, Nonconvex composite optimization, Gradient tracking, Momentum method
\end{IEEEkeywords}

\section{Introduction}\label{sec:introduction}
Federated learning (FL) trains a shared model across multiple clients while keeping their raw data local.
In the standard Federated Averaging (FedAvg) framework \cite{mcmahan2017communication}, a central server broadcasts the current model, multiple clients perform several local training steps, and the server aggregates their model updates.
Performing multiple local steps between communication rounds reduces communication relative to conventional distributed optimization algorithms \cite{gao2023decentralized}.
However, the central coordinator creates two limitations: it is a single point of failure, and it can become a communication bottleneck as the number of participating clients grows \cite{kumar2023impact,kairouz2021advances}.

In decentralized FL (DFL), clients communicate directly with their neighbors over a connected communication graph rather than relying on a central coordinator.
Each client exchanges information only with its neighbors and updates its local model accordingly \cite{sun2022decentralized,liu2022decentralized,beltran2023decentralized}.
This decentralized architecture distributes the communication load in large-scale systems and removes the server bottleneck in FedAvg.
It also avoids dependence on a single coordinator, although its performance depends on the connectivity of the underlying network.
DFL is therefore suitable for settings in which a central server is unavailable or undesirable, such as edge and vehicular networks \cite{gao2023decentralized,kairouz2021advances,beltran2023decentralized}.

Most existing FL/DFL methods consider unconstrained smooth objectives.
In many applications, however, a regularizer is introduced to promote sparsity or low-rank structure, or an indicator function is used to encode constraints.
These settings lead to Federated Composite Optimization (FCO) \cite{yuan2021federated,zhang2024composite,bao2022fast}.
We consider the following Nonconvex Composite Optimization Problem (NCOP):
\begin{equation}\label{equ:OriginalProblem}
\min_{x\in\mathbb{R}^d}  \   \phi(x)=f(x)+\beta h(x), \ f(x)\triangleq\frac{1}{n}\sum_{i=1}^nf_i(x),
\end{equation}
where each local function $f_i$ is smooth but need not be convex, and $h$ is a common weakly convex function that may be nondifferentiable.
The parameter $\beta>0$ controls the relative strength of the regularization term, allowing the tradeoff between fitting the local data and imposing the structure encoded by $h$ to be adjusted without changing the smooth loss functions.
Typical choices of $h$ include the $\ell_1$ norm, the $\ell_2$ norm, the Minimax Concave Penalty (MCP), the Smoothly Clipped Absolute Deviation (SCAD), and indicator functions \cite{zeng2018nonconvex,hajinezhad2016nestt,bohm2021variable}.
Existing studies of related composite problems remain centralized and require a server to maintain the global model \(x\) \cite{yuan2021federated,zhang2024composite}.
We therefore develop a Decentralized Nonconvex Composite FL (DNCFL) method that allows multiple local updates between communication rounds.

\noindent\textbf{Related Work.}
Stochastic Gradient Descent (SGD) replaces the full gradient with a stochastic estimate computed from a mini-batch, making it suitable for large-scale learning problems with limited computational resources \cite{yu2019linear,liu2020accelerated}.
In distributed settings, FedAvg can be regarded as parallel SGD with multiple local updates \cite{yu2019parallel,lian2017can}, commonly referred to as local SGD \cite{koloskova2020unified}.
Removing the central server yields Decentralized SGD (DSGD).
For nonconvex optimization, DSGD can attain convergence rates close to those of centralized methods, with additional terms that depend on network connectivity \cite{lian2017can,xin2021improved,koloskova2020unified,yuan2020influence,george2020distributed}.
Standard DSGD nevertheless has limitations.
With a fixed stepsize, even under deterministic gradients, it typically converges only to a neighborhood of an optimal solution in convex problems or of a stationary point in nonconvex problems \cite{yuan2016convergence,zeng2018nonconvex}.

Due to the distribution of training data, data heterogeneity is common in distributed machine learning \cite{huang2024federated,vettoruzzo2024advances,kwon2023tighter}.
Such heterogeneity can arise from differences across clients in feature distributions, label distributions, and dataset sizes \cite{mcmahan2017communication,kairouz2021advances}.
To control its effect, analyses of related work often impose a bounded gradient heterogeneity condition, for example,
\begin{equation}\label{equ:GradientHeterogeneity}
\frac{1}{n}\sum_{i=1}^{n}\|\nabla f_i(x)-\nabla f(x)\|^2 \leq \zeta^2,\quad \forall x\in\mathbb{R}^d.
\end{equation}
This condition bounds the discrepancy between local and global gradients and is used to control the bias caused by heterogeneous local updates \cite{yu2019linear,lian2017can,koloskova2020unified,yuan2020influence,bao2022fast,li2018simple}.
Without a correction mechanism, this bias can cause ``client drift" and slow convergence.
Bounds on local gradient norms are another commonly used assumption, but they are conceptually distinct from bounded gradient heterogeneity \cite{li2024problem,yu2019parallel,sun2022decentralized}.
To avoid relying on such assumptions in convergence analyses, decentralized stochastic gradient tracking (DSGT) methods maintain an auxiliary variable that tracks the average of the local gradients \cite{li2024problem,xin2021improved,nguyen2023performance,gao2023distributed,koloskova2021improved,liu2024decentralized}.
Related bias-correction methods include EXTRA \cite{shi2015extra} and Exact-Diffusion/\(\text{D}^2\) \cite{yuan2020influence,alghunaim2024local,tang2018d}.
The unified framework in \cite{alghunaim2022unified} connects several of these methods.
Model-update tracking provides another correction mechanism: Global Update Tracking (GUT) reconstructs neighboring models from communicated updates \cite{aketi2023global}, and FedNMUT extends this approach to imperfect information sharing caused by additive communication noise \cite{chellapandi2024decentralized}.

Momentum is commonly used in stochastic optimization to accelerate training and reduce the effect of stochastic gradient noise \cite{yu2019linear,gao2023distributed,yan2018unified,liu2020improved,mai2020convergence}.
Two standard stochastic momentum schemes are Polyak momentum, also known as stochastic heavy-ball (SHB) momentum, and Nesterov momentum, also known as stochastic Nesterov accelerated gradient (SNAG) momentum \cite{yan2018unified,liu2023last,liang2023stochastic}.
The unified framework in \cite{yan2018unified} shows that the momentum term enhances training stability and can improve generalization performance.
Recent decentralized methods combine momentum with gradient tracking \cite{huang2024accelerated,takezawa2023momentum,xin2019distributed,gao2023distributed}, aiming to leverage the strengths of both approaches.
They differ in the quantity tracked by the auxiliary variable: the methods in \cite{gao2023distributed,xin2019distributed} update it using stochastic gradients, whereas those in \cite{huang2024accelerated,takezawa2023momentum} track momentum variables.
For composite optimization, \cite{gao2024non} shows that proximal SGD with Polyak momentum can attain optimal convergence rates for NCOPs without mega-batches or exact proximal subproblem solutions.
In FL, \cite{cheng2023momentum} shows that momentum can avoid the bounded gradient heterogeneity assumption in the analysis and accelerate SCAFFOLD proposed in \cite{karimireddy2020scaffold}.

Linear speedup is an important scalability property in distributed learning.
Under suitable conditions, using \(n\) clients reduces the per-client computation required to reach a given accuracy by a factor proportional to \(n\) \cite{yu2019linear,li2024problem,gao2023distributed,qu2021federated}.
For nonconvex DSGD, \cite{lian2017can} establishes a convergence rate of \(\mathcal{O}(1/\sqrt{nT})\) when \(T\) (the number of iterations) is sufficiently large.
Compared with the single-agent SGD rate \(\mathcal{O}(1/\sqrt{T})\), this rate is consistent with linear speedup \cite{yu2019linear,lian2017can}.
Similar guarantees have been established for some FL algorithms \cite{qu2021federated,bao2022fast,sun2023efficient} and a range of decentralized methods, including momentum-based DSGD, gradient-tracking methods, and their momentum variants \cite{yu2019linear,xin2021improved,gao2023distributed,takezawa2023momentum,liu2024decentralized,li2024problem,alghunaim2022unified}.

FCO has mainly been studied in the centralized FL setting \cite{yuan2021federated,tran2021feddr,wang2022fedadmm,bao2022fast,zhang2024composite}.
In \cite{yuan2021federated}, FedMiD extends FedAvg to composite objectives through proximal updates.
The same work notes that server aggregation can suffer from the ``curse of primal averaging'' in the presence of regularization and introduces FedDualAvg as an alternative.
FedDR \cite{tran2021feddr} and FedADMM \cite{wang2022fedadmm} apply Douglas-Rachford splitting and the alternating direction method of multipliers (ADMM), respectively, to FCO.
In addition, \cite{bao2022fast,zhang2024composite} establish convergence guarantees under smooth and strongly convex empirical losses.
These methods remain centralized and require a server to maintain the global model.
Moreover, smooth strongly convex assumptions exclude many learning problems in which neural-network training induces nonconvex objectives.
Decentralized methods for nonconvex composite FL, especially those supporting local updates, remain limited.

Research on NCOPs related to \eqref{equ:OriginalProblem} includes both centralized and decentralized methods.
Centralized approaches combine proximal SGD with variance reduction \cite{li2018simple,wang2019spiderboost}, stochastic ADMM \cite{huang2018mini}, or Polyak momentum \cite{gao2024non}, but do not address decentralized networks.
Among decentralized methods, \cite{wang2021distributed} develops a stochastic primal-dual method with momentum, while \cite{xin2021stochastic} introduces the ProxGT framework based on proximal gradient tracking.
The stochastic primal-dual method and ProxGT-SA require batch sizes that depend on the target accuracy, whereas ProxGT-SR-O/E improves sample complexity through periodic mega-batch or full-gradient refreshes.
DEEPSTORM \cite{mancino2023proximal} instead uses recursive momentum to enable mini-batch updates under mean-squared smoothness.
Prox-DASA-GT \cite{xiao2023one} and DProxSGT \cite{yan2023compressed} further provide mini-batch guarantees without this stronger smoothness assumption.
These decentralized methods nevertheless exchange information at every optimization iteration, and some use additional consensus rounds to reduce network dependence.
None provides a convergence guarantee for the decentralized local-update setting considered in this paper.

\noindent\textbf{Contributions.}
Our main contributions are as follows.
\begin{itemize}
\item We propose DEPOSITUM, a decentralized proximal stochastic gradient-tracking method for NCOP.
It supports smooth nonconvex local objectives and nonsmooth weakly convex regularizers, incorporates both Polyak and Nesterov momentum, and permits multiple local updates between communication phases.
The resulting framework does not require bounded gradient heterogeneity and accounts for stochastic-gradient noise, while multiple local iterations between communication phases reduce communication frequency without a central server.

\item  
We establish convergence guarantees for both Polyak and Nesterov momentum variants without bounded gradient heterogeneity or mean-squared smoothness assumptions.
DEPOSITUM obtains an expected $\epsilon$-stationary point with an iteration complexity of $\mathcal{O}(1/\epsilon^2)$. 
With an appropriate stepsize and momentum schedule, it further attains an averaged stationarity rate of $\mathcal{O}(1/\sqrt{nT})$ after a network-dependent transient, with a mini-batch size independent of $T$.
The analysis also quantifies the tradeoff among the communication period, admissible stepsize, and network connectivity.


\item  
We conduct extensive experiments on five datasets spanning tabular learning, image classification, and decentralized parameter-efficient language-model fine-tuning.
Across different models, regularizers, data partitions, network topologies, communication periods, and client counts, the results support the theoretical findings and demonstrate the effectiveness of DEPOSITUM relative to representative methods.

\end{itemize}

\noindent\textbf{Notation.} 
We use \(\mathbb{R}^n\) and \(\mathbb{R}^{n\times m}\) to denote the spaces of \(n\)-dimensional real vectors and \(n\)-by-\(m\) real matrices, respectively.
Unless otherwise specified, \(\|\cdot\|\) and \(\|\cdot\|_1\) denote the Euclidean and \(\ell_1\) norms.
The symbols \(\mathbf{0}\), \(\mathbf{1}\), and \(\mathbf{I}_n\) denote all-zero vectors, all-one vectors, and the \(n\)-by-\(n\) identity matrix, respectively.
We use \(\nabla\) and \(\partial\) to denote the gradient and subdifferential operators, respectively.

For a client-wise variable, bold lowercase notation denotes its stacked form.
For example, $
\mathbf{x}=[x_1;\cdots; x_n]\in\mathbb{R}^{nd}$.
Let $\mathbf{J}=(\frac{1}{n}\mathbf{1}_{n}\mathbf{1}_{n}^{\top})\otimes \mathbf{I}_d\in\mathbb{R}^{nd\times nd}$ denote the averaging matrix.
Then \(\mathbf{J}\mathbf{x}=[\bar{x};\ldots;\bar{x}]\), where \(\bar{x}=\frac{1}{n}\sum_{i=1}^{n}x_i\).
We use the same convention for $\mathbf{y}$, $\bm{\mu}$, $\bm{\nu}$, and $\mathbf{g}$.
For compact notation, define $\mathbf{f}(\mathbf{x})\triangleq\sum_{i=1}^nf_i(x_i)$ and $\mathbf{h}(\mathbf{x})\triangleq\sum_{i=1}^nh(x_i)$.
The gradient of $\mathbf{f}(\mathbf{x})$ and the average of all its local gradients are
$$\nabla\mathbf{f}(\mathbf{x})=\left[\nabla f_1(x_1);\cdots;\nabla f_n(x_n)\right], \overline{\nabla \mathbf{f}}(\mathbf{x})=\frac{1}{n}\sum_{i=1}^n\nabla f_i({x}_i).$$
For $\alpha>0$, the proximal operator of $\mathbf{h}$ has the stacked form
$$
\textbf{prox}_{\mathbf{h}}^{\alpha}\{\mathbf{x}\}=\left[\textbf{prox}_{h}^{\alpha}\{x_1\};\cdots;\textbf{prox}_{h}^{\alpha}\{x_n\}\right],
$$
where the local proximal operator is defined by $\textbf{prox}_{h}^{\alpha}\{x\}=\arg \min_z\{h(z)+\frac{\alpha}{2}{\Vert z-x \Vert}^2\}$.
 
\section{Problem Formulation and Preliminaries}
We consider \(n\) clients connected by an undirected and connected graph \(\mathcal{G}=(\mathcal{V},\mathcal{E})\), where \(\mathcal{V}=\{1,\ldots,n\}\) is the client set and \(\mathcal{E}\subseteq\mathcal{V}\times\mathcal{V}\) is the edge set.
Without a central server, client \(i\) communicates only with its neighbors \(\mathcal{N}_i=\{j\in\mathcal{V}:(i,j)\in\mathcal{E},\,j\neq i\}\).
Accordingly, NCOP \eqref{equ:OriginalProblem} is equivalently written as the decentralized consensus problem
\begin{equation}\label{equ:CompactProblem}
\begin{split}
\min  \quad  &\frac{1}{n}\sum_{i=1}^n\phi_i(x_i)\triangleq f_i(x_i)+\beta h(x_i), \\
s.t. \quad  & x_i=x_j, \quad \forall (i,j)\in\mathcal{E}.
		\end{split}
	\end{equation}
In the stochastic online or streaming setting, the local loss is
$$
f_i(x_i)\triangleq\mathbb{E}_{\xi_i\sim\mathcal{D}_i}[F_i(x_i,\xi_i)], \ i\in\mathcal{V},
$$
where $\xi_i$ is a random sample from the local data distribution $\mathcal{D}_i$.
Generally, the distributions ${\{\mathcal{D}_i\}}_{i=1}^n$ are different, reflecting the heterogeneity of local private data.
The consensus constraints ensure that the local variables \(x_1=x_2=\cdots=x_n\), thereby recovering the original problem \eqref{equ:OriginalProblem}.

We first recall the following definition.
\begin{defn}[Weak Convexity]
A function $h:\mathbb{R}^d\rightarrow\mathbb{R}\cup\{+\infty\}$ is defined as $\rho$-weakly convex if $h(\cdot)+\frac{\rho}{2}{\Vert \cdot\Vert}^2$ is convex for some \(\rho\geq 0\).
	\end{defn}
When $\rho=0$, a weakly convex function is convex.
For \(\rho>0\), this class also includes certain nonconvex regularizers.
In particular, for a proper, closed, \(\rho\)-weakly convex function \(h\), the proximal operator \(\textbf{prox}_{h}^{\alpha}\) is well defined and single-valued when \(\alpha>\rho\), since the corresponding proximal subproblem is strongly convex \cite{liu2023proximal}.
Typical examples in machine learning include MCP and SCAD, whose expressions and corresponding proximal operators can be found in \cite{bohm2021variable}.

\noindent\textbf{Stochastic Momentum Methods.}
Let \(\nabla F(x^t,\xi^t)\) denote an unbiased, bounded-variance estimator of \(\nabla f(x^t)\).
For $t\geqslant 1$, the update of SHB is given by
\begin{equation}\label{equ:SHB}
    x^{t+1}=x^t-\alpha(1-\gamma)\nabla F(x^t,\xi^t)+\gamma(x^t-x^{t-1}),
\end{equation}
where $\alpha>0$ is the stepsize, $\gamma\in[0,1)$ is the momentum parameter, and $x^0=x^1\in\mathbb{R}^d$.
The update of SNAG consists of two successive steps for $t\geqslant 1$:
\begin{subequations}\label{equ:SNAG}
\begin{align}
    z^{t+1}&=x^t-\alpha(1-\gamma) \nabla F(x^t,\xi^t), \\
    x^{t+1}&=z^{t+1}+\gamma(z^{t+1}-z^t),
    \end{align}
\end{subequations}
where \(z^1=x^1\).
Both \eqref{equ:SHB} and \eqref{equ:SNAG} admit the following momentum-based parameter update:
\begin{subequations}\label{equ:SMM}
\begin{align}
    \text{SHB} &\qquad\nu^{t+1}=\gamma\nu^t+(1-\gamma)\nabla F(x^t,\xi^t); \label{equ:SMM1}\\
    \text{SNAG} &\qquad\begin{cases}
        \mu^{t+1}=\gamma\mu^t+(1-\gamma)\nabla F(x^t,\xi^t), \\
        \nu^{t+1}=\gamma\mu^{t+1}+(1-\gamma)\nabla F(x^t,\xi^t);
    \end{cases}  \label{equ:SMM2}\\
    \text{SGD} &\qquad x^{t+1}=x^t-\alpha\nu^{t+1}. \label{equ:SMM3}
\end{align}
\end{subequations}
The momentum estimate \(\nu^{t+1}\) replaces the stochastic gradient in the update.
Combining \eqref{equ:SMM1} with \eqref{equ:SMM3} recovers \eqref{equ:SHB}, whereas combining \eqref{equ:SMM2} with \eqref{equ:SMM3} recovers \eqref{equ:SNAG} \cite{gao2023distributed,mai2020convergence,liu2020improved}.
Setting \(\gamma=0\) reduces both methods to vanilla SGD.

\noindent\textbf{Decentralized Stochastic Gradient Tracking.}
For the smooth case \(h\equiv0\) of \eqref{equ:CompactProblem}, a standard adapt-then-combine (ATC) DSGT scheme \cite{pu2021distributed,koloskova2021improved,xin2020general} updates the variables at client \(i\in\mathcal{V}\) as
\begin{subequations}\label{alg:GT}
    \begin{align}
    x_i^{t+1}&=\sum_{j=1}^nw_{ij}(x_j^t-\alpha y_j^t), \label{alg:GTEq1}\\
    y_i^{t+1}&=\sum_{j=1}^nw_{ij}y_j^t\!+\!\nabla F_i(x_i^{t+1},\xi_i^{t+1})\!-\!\nabla F_i(x_i^t,\xi_i^t), \label{alg:GTEq2}
\end{align}
\end{subequations}
where \(y_i^t\) is the local gradient-tracking variable and \(W_n=[w_{ij}]\) is a mixing matrix associated with \(\mathcal{G}\).
All weights $w_{ij}\geqslant0$ satisfy $\sum_{i=1}^nw_{ij}=1$ and $\sum_{j=1}^nw_{ij}=1$, $\forall i,j\in\mathcal{V}$.
For the composite optimization problem \eqref{equ:CompactProblem}, replacing the local gradient step in \eqref{alg:GTEq1} with a proximal gradient step yields
\begin{equation*}
x_i^{t+1}=\sum_{j=1}^nw_{ij}\textbf{prox}_{h}^{1/(\alpha\beta)}\{x_j^t-\alpha y_j^t\},
\end{equation*}
where $0<\alpha\beta\rho<1$.
As in \cite{xin2021improved,xin2019distributed,xin2021stochastic,yan2023compressed}, \eqref{alg:GTEq2} can also be replaced by
\begin{equation}
y_i^{t+1}\!=\!\sum_{j=1}^nw_{ij}\!\left[y_j^t\!+\!\nabla F_j(x_j^{t+1},\xi_j^{t+1})\!-\!\nabla F_j(x_j^{t},\xi_j^{t})\right]. \label{alg:GTEq3}
\end{equation}
Both forms can be used to update the gradient-tracking variable $y$ \cite{xin2019distributed}.
If $y_i^0=\nabla F_i(x_i^{0},\xi_i^{0})$ for all $i\in\mathcal{V}$, induction using \eqref{alg:GTEq2} or \eqref{alg:GTEq3} gives $$\bar{y}^t=\frac{1}{n}\sum_{i=1}^ny_i^t=\frac{1}{n}\sum_{i=1}^n\nabla F_i(x_i^{t},\xi_i^{t}).$$ Thus, $\bar{y}$ can be used to track the global gradient as consensus errors decrease \cite{xin2021improved,xin2021stochastic,pu2021distributed}.
\section{Algorithm Development}
We impose the following assumptions regarding the objective function, mixing matrix, and stochastic gradient.

\begin{assum}\label{ass:AssumptionFunction}
The components of \eqref{equ:OriginalProblem} satisfy:
		\begin{enumerate}[i.]
\item The objective function $\phi$ is lower bounded, i.e., $\phi^*=\inf_x \phi(x)>-\infty$;\label{ass:AssumptionFunctioni}
\item Each local function $f_i:\mathbb{R}^d\rightarrow\mathbb{R}$ is differentiable and \(L\)-smooth for some \(L>0\), i.e., \label{ass:AssumptionFunctionii}
 	$$\Vert\nabla  f_i(x)-\nabla  f_i(y)\Vert\leqslant L\Vert x-y\Vert, \forall x, y\in\mathbb{R}^d.$$
\item The regularizer $h:\mathbb{R}^d\rightarrow\mathbb{R}\cup\{+\infty\}$ is proper, closed, $\rho$-weakly convex, and its proximal operator $\textbf{prox}_{h}^{\tau}$ is easy to obtain for $\tau>\rho\geqslant0$.
\label{ass:AssumptionFunctioniii}
		\end{enumerate}
	\end{assum}
 
\begin{assum}\label{ass:AssumptionMixingMatrix}
The mixing matrix $W_n=[w_{ij}]$ satisfies:
\begin{enumerate}[i.]
\item
$W_n=W_n^{\top}$ and $W_n\mathbf{1}_{n}=\mathbf{1}_{n}$.
\item
\(w_{ii}>0\) for all \(i\in\mathcal{V}\).
For \(i\neq j\), \(w_{ij}>0\) if \((i,j)\in\mathcal{E}\), and \(w_{ij}=0\), otherwise.
		\end{enumerate}
	\end{assum}
 
Because the graph \(\mathcal G\) is connected and Assumption \ref{ass:AssumptionMixingMatrix} holds, the eigenvalues of $W_n$ satisfy $1=\lambda_1>\lambda_2\geqslant\cdots\geqslant\lambda_n>-1$.
We define
$$
\lambda\triangleq\max\{|\lambda_2|,|\lambda_n|\}=\left\Vert W_n-\frac{1}{n}\mathbf{1}_{n}\mathbf{1}_{n}^{\top} \right\Vert_2\in [0,1).
$$
If \(W_n=\frac{1}{n}\mathbf{1}_n\mathbf{1}_n^\top\), which corresponds to uniform averaging over a complete graph, then \(\lambda=0\).
More generally, a smaller \(\lambda\), or equivalently a larger absolute
\textit{spectral gap} \(1-\lambda\) \cite{mancino2023proximal}, indicates faster consensus mixing.
\begin{assum}\label{ass:AssumptionStochastic}
Let $\mathcal{F}_t$ denote the sigma-algebra generated by all randomness revealed before the mini-batches at iteration $t$ are sampled.
Conditional on $\mathcal{F}_t$, the samples \(\{\xi_{i,b}^t: i\in\mathcal{V},\ b=1,\ldots,B\}\) are mutually independent, and each $\xi_{i,b}^t$ is distributed according to $\mathcal{D}_i$.
Moreover, for every $x\in\mathbb{R}^d$,
\begin{align*}
&\mathbb{E}_{\xi_i\sim\mathcal{D}_i}
    [\nabla F_i(x,\xi_i)]
    = \nabla f_i(x),\\
&\mathbb{E}_{\xi_i\sim\mathcal{D}_i}
    \bigl[
    \|\nabla F_i(x,\xi_i)-\nabla f_i(x)\|^2
    \bigr]
    \leq \sigma^2.
\end{align*}
\end{assum}
To mitigate the impact of stochastic-gradient variance, one can employ
mini-batch stochastic approximation \cite{ghadimi2016mini}. For a
mini-batch size $B\in\mathbb{N}_+$, define
\[
g_i^t
=
\frac{1}{B}
\sum_{b=1}^B
\nabla F_i(x_i^t,\xi_{i,b}^t).
\]
Under Assumption \ref{ass:AssumptionStochastic}, the mini-batch estimator
remains conditionally unbiased, and its conditional variance is bounded by
$\sigma^2/B$ \cite{huang2018mini,xin2021stochastic}.


We now present the \underline{D}ecentralized f\underline{E}derated \underline{P}r\underline{O}ximal \underline{S}tochastic grad\underline{I}ent \underline{T}racking with moment\underline{UM} (DEPOSITUM) algorithm.
Given a common initial model $x_0$, client $i$ forms
\[
g_i^0=\frac{1}{B}\sum_{b=1}^B\nabla F_i(x_0,\xi_{i,b}^0),
\qquad \bar g^0=\frac{1}{n}\sum_{i=1}^n g_i^0,
\]
and initializes $x_i^0=x_0$, $\mu_i^0=\nu_i^0=g_i^0$, and $y_i^0=\bar g^0$.
Let \(\mathcal{T}\triangleq\{kT_0:k=0,1,\ldots\}\) be the communication schedule, where \(T_0\in\mathbb{N}_+\), and define a time-varying matrix
\[
\mathbf{W}^t=
\begin{cases}
\mathbf{W}\triangleq W_n\otimes\mathbf{I}_d, & t\in\mathcal{T},\\
\mathbf{I}_{nd}, & t\notin\mathcal{T}.
\end{cases}
\]
At iteration $t$, DEPOSITUM first updates the local models by
\begin{equation}\label{equ:Update4}
\mathbf{x}^{t+1}=\mathbf{W}^t\textbf{prox}_{\mathbf{h}}^{1/(\alpha\beta)}
\{\mathbf{x}^t-\alpha\mathbf{y}^{t}\}.
\end{equation}
Each client then evaluates the fresh mini-batch gradient $g_i^{t+1}$.
DEPOSITUM has two fixed variants, using Polyak momentum and Nesterov momentum, respectively.
For Polyak momentum,
\begin{equation}\label{equ:Update1}
\bm{\nu}^{t+1}=\gamma\bm{\nu}^{t}+(1-\gamma)\mathbf{g}^{t+1},
\end{equation}
whereas Nesterov momentum uses
\begin{subequations}\label{equ:Update23}
\begin{align}
\bm{\mu}^{t+1}&=\gamma\bm{\mu}^{t}+(1-\gamma)\mathbf{g}^{t+1},\label{equ:Update2} \\
\bm{\nu}^{t+1}&=\gamma\bm{\mu}^{t+1}+(1-\gamma)\mathbf{g}^{t+1}. \label{equ:Update3}
\end{align}
\end{subequations}
Finally, the gradient-tracking variable follows the change in the momentum estimate:
\begin{equation}\label{equ:Update5}
\mathbf{y}^{t+1}=\mathbf{W}^t(\mathbf{y}^{t}+\bm{\nu}^{t+1}-\bm{\nu}^{t}).
\end{equation}
Here $\alpha>0$ is the stepsize, $\beta>0$ is the regularization weight in \eqref{equ:OriginalProblem}, and $\gamma\in[0,1)$ is the momentum parameter.
Every iteration follows the order $\mathbf{x}^{t+1}\rightarrow\mathbf{g}^{t+1}\rightarrow\bm\nu^{t+1}\rightarrow\mathbf{y}^{t+1}$, with $\bm\mu^{t+1}$ updated immediately before $\bm\nu^{t+1}$ in the Nesterov case.
Clients exchange information with their neighbors only when \(t\in\mathcal{T}\); otherwise, they perform local updates.
When $T_0=1$, clients communicate at every iteration.
Algorithm \ref{alg1} gives the client-wise implementation.

\begin{algorithm}[t]
    \caption{DEPOSITUM for client $i$}
    \renewcommand{\algorithmicrequire}{\textbf{Initialization:}}
    \begin{algorithmic}[0]\label{alg1}
        \REQUIRE $x_i^0=x_0$, $g_i^0=\frac{1}{B}\sum_{b=1}^B\nabla F_i(x_0,\xi_{i,b}^0)$, $\mu_i^0=\nu_i^0=g_i^0$, $y_i^0=\bar g^0$, $\alpha>0$, $\beta>0$, $\gamma\in[0,1)$, $T_0,B\in\mathbb{N}_{+}$, $\mathcal{T}=\{kT_0:k=0,1,\ldots\}$. 
        \FOR{\(t=0,1,\ldots,T\)}
            \STATE \(\displaystyle x_i^{t+1}=\begin{cases}
                \sum_{j\in\mathcal{N}_i\cup\{i\}}w_{ij}\textbf{prox}_{h}^{1/(\alpha\beta)}\{x_j^t-\alpha y_j^t\}, & t\in\mathcal{T},\\
                \textbf{prox}_{h}^{1/(\alpha\beta)}\{x_i^t-\alpha y_i^t\}, &  t\notin\mathcal{T}.
            \end{cases}\)
            \STATE \(\displaystyle g_i^{t+1}=\frac{1}{B}\sum_{b=1}^B\nabla F_i(x_i^{t+1},\xi_{i,b}^{t+1}).\)
            \STATE \(\displaystyle \left\{\begin{aligned}
                \text{Polyak:}\quad &\nu_i^{t+1}=\gamma\nu_i^t+(1-\gamma)g_i^{t+1},\\
                \text{Nesterov:}\quad &\mu_i^{t+1}=\gamma\mu_i^t+(1-\gamma)g_i^{t+1},\\[-1pt]
                &\nu_i^{t+1}=\gamma\mu_i^{t+1}+(1-\gamma)g_i^{t+1}.
            \end{aligned}\right.\)
            \STATE \(\displaystyle y_i^{t+1}=\begin{cases}
                \sum_{j\in\mathcal{N}_i\cup\{i\}}w_{ij}[y_j^t+\nu_j^{t+1}-\nu_j^t], & t\in\mathcal{T},\\
                y_i^t+\nu_i^{t+1}-\nu_i^t, &  t\notin\mathcal{T}.
            \end{cases}\)
        \ENDFOR
    \end{algorithmic}
\end{algorithm}

\begin{rem}
For both \(\mathbf{W}^t=\mathbf{W}\) and \(\mathbf{W}^t=\mathbf{I}_{nd}\), we have \(\mathbf{J}\mathbf{W}^t=\mathbf{J}\).
Hence, \eqref{equ:Update5} gives
\[
\mathbf{J}\mathbf{y}^{t+1}=\mathbf{J}\mathbf{y}^t+\mathbf{J}(\bm{\nu}^{t+1}-\bm{\nu}^t).
\]
Since \(\mathbf{J}\mathbf{y}^0=\mathbf{y}^0=\mathbf{J}\bm{\nu}^0\), it follows that \(\mathbf{J}\mathbf{y}^t=\mathbf{J}\bm{\nu}^t\) for all \(t\).
Thus, the local iterations between communication phases preserve the network-average tracking property.
\end{rem}

\begin{rem}
Prior to the iterations, the common initialization $y_i^0=\bar g^0$ can be established by a single synchronization. 
If this synchronization is omitted, each client may initialize $y_i^0=g_i^0$ locally. 
Although this alternative preserves the network-average tracking property, it produces a nonzero initial tracking consensus error. 
The associated initialization-dependent term decays through network mixing and therefore affects only the transient behavior. 
For analytical convenience, we set $y_i^0=\bar g^0$ in the next section.
\end{rem}

\section{Theoretical Analysis}
Before analyzing the convergence of DEPOSITUM, we introduce the following definitions.

\begin{defn}[Proximal-Gradient Mappings]\label{def:ProxGrad} 
For NCOP \eqref{equ:OriginalProblem}, let $x,y\in\mathbb{R}^d$, $\alpha>0$, $\beta>0$, and $\alpha\beta\rho<1$.
Define
    \[
G^\alpha(x,y)\triangleq\frac{1}{\alpha}
        \left[x-\textbf{prox}_{h}^{1/(\alpha\beta)}
        \{x-\alpha y\}\right]
    \]
as the surrogate proximal-gradient mapping induced by the direction $y$.
When $y=\nabla f(x)$, it reduces to the exact proximal-gradient mapping $G^\alpha(x)\triangleq G^\alpha(x,\nabla f(x))$.
For $\mathbf{x}=[x_1;\ldots;x_n]$ and $\mathbf{y}=[y_1;\ldots;y_n]$, define
    \[
        \mathbf{G}^\alpha(\mathbf{x},\mathbf{y})
        \triangleq
[G^\alpha(x_1,y_1);\ldots;G^\alpha(x_n,y_n)],
    \]
    \[
        \mathbf{G}^\alpha(\mathbf{x})
        \triangleq
[G^\alpha(x_1,\nabla f(x_1));\ldots; G^\alpha(x_n,\nabla f(x_n))].
    \]
Accordingly, $\mathbf{G}^\alpha(\mathbf{x},\mathbf{y})$ and $\mathbf{G}^\alpha(\mathbf{x})$ are the stacked surrogate and exact proximal-gradient mappings, respectively.
\end{defn}

The exact proximal-gradient mapping is a standard stationarity measure for composite optimization \cite{mancino2023proximal,hajinezhad2016nestt,xiao2023one}.
In particular, $G^{\alpha}(x^*)=\mathbf{0}$ is equivalent to the first-order condition $\mathbf{0}\in\nabla f(x^*)+\beta\partial h(x^*)$.
When $h=0$, it reduces to $\nabla f(x)$, and centralized proximal gradient descent satisfies $x^{t+1}=x^t-\alpha G^\alpha(x^t)$.
For convex \(h\), the relation below follows from the nonexpansiveness of the proximal operator.
For weakly convex \(h\), it follows from the \(1/(1-\alpha\beta\rho)\)-Lipschitz continuity of the proximal operator:
\begin{equation}\label{equ:ProximalGradientRelationMain}
\Vert G^{\alpha}(x,y)-G^{\alpha}(x)\Vert^2
\leqslant\frac{\Vert y-\nabla f(x)\Vert^2}{(1-\alpha\beta\rho)^2}.
\end{equation}

\begin{defn}[Expected $\epsilon$-Stationary Point] \label{def:StationaryPoint}
Let $\mathbf{x}\in\mathbb{R}^{nd}$ and $\bar{\nu}=\frac{1}{n}\sum_{i=1}^n\nu_i\in\mathbb{R}^d$ be generated by Algorithm 1.
Define
    \begin{equation*}
    \mathbf{s}(\mathbf{x},\bar{\nu}):={\Vert \mathbf{G}^{\alpha}(\mathbf{x})\Vert}^2+L^2{\Vert \mathbf{J}\mathbf{x}-\mathbf{x}\Vert}^2+n{\Vert \overline{\nabla \mathbf{f}}(\mathbf{x})-\bar{\nu}\Vert}^2.
    \end{equation*}
We call $\mathbf{x}$ an expected $\epsilon$-stationary point if $\frac{1}{n}\mathbb{E}[\mathbf{s}(\mathbf{x},\bar{\nu})]\leqslant\epsilon$.
\end{defn}
The second and third terms in $\mathbf{s}(\mathbf{x},\bar{\nu})$ measure the model consensus error and the averaged gradient-estimation error, respectively.
When the local models reach consensus and \(\bar{\nu}=\overline{\nabla\mathbf f}(\mathbf{x})\), the measure reduces to the standard proximal-gradient stationarity measure for the global objective.

Throughout this analysis, let
\(\{\mathbf x^t,\mathbf y^t,\bm\nu^t\}_{t=0}^{T+1}\)
be generated by Algorithm \ref{alg1} with either Polyak or Nesterov momentum.
The following proposition establishes the basic descent relation for DEPOSITUM.
Because the local update \eqref{equ:Update4} uses the gradient-tracking variable \(\mathbf y^t\), the descent term is expressed through the surrogate proximal-gradient mapping \(\mathbf G^\alpha(\mathbf x^t,\mathbf y^t)\).
The proposition relates this surrogate mapping to \(\mathbf{s}(\mathbf x^t,\bar\nu^t)\).
\begin{prop}\label{prop1}
Suppose that Assumptions \ref{ass:AssumptionFunction} and \ref{ass:AssumptionMixingMatrix} hold.
If $\alpha L\leqslant1/16$ and $\alpha\beta\rho\leqslant1/48$, then
\begin{align}
&\sum_{t=0}^{T}\mathbf s(\mathbf x^t,\bar\nu^t)
\leqslant{} \frac{8n}{\alpha}[\phi(x_0)-\phi^*]
-\sum_{t=0}^{T}\Vert\mathbf G^\alpha(\mathbf x^t,\mathbf y^t)\Vert^2 \notag\\
&\qquad+142n\sum_{t=0}^{T}\Vert\overline{\nabla\mathbf f}(\mathbf x^t)-\bar\nu^t\Vert^2 +\frac{5}{\alpha^2}\sum_{t=0}^{T}\Vert\mathbf J\mathbf x^t-\mathbf x^t\Vert^2 \notag\\
&\qquad+11\sum_{t=0}^{T}\Vert\mathbf J\mathbf y^t-\mathbf y^t\Vert^2.
\label{equ:Proposition1}
\end{align}
\end{prop}

Apart from the initial objective gap, its right-hand side contains the four accumulated quantities shown in the upper row of Fig. \ref{fig:proofroadmap}.
The first auxiliary result bounds the averaged gradient-estimation error by the accumulated differences between successive model iterates and a stochastic-variance term.

\begin{figure*}[ht]
\centering
\begin{tikzpicture}[
    font=\scriptsize,
    box/.style={draw, rounded corners=1pt, align=center,
        inner xsep=1.2pt, inner ysep=0.8pt, fill=black!3},
    module/.style={box, line width=0.6pt, fill=black!6,
        minimum height=9.6mm},
    auxiliary/.style={box, dashed, fill=white, minimum height=10.2mm},
    result/.style={box, rounded corners=2.2pt, line width=0.55pt,
        double, double distance=0.45pt, fill=white,
        minimum height=10.2mm},
    arr/.style={-{Latex[length=2mm,width=1.35mm]}, line width=0.65pt,
        shorten >=0.8pt},
    looparr/.style={arr, line width=0.78pt},
    outarr/.style={arr, line width=0.82pt},
    lab/.style={fill=white, fill opacity=0.92, text opacity=1,
        inner xsep=0.6pt, inner ysep=0.2pt, align=center},
    node distance=4mm]
\node[module, text width=4.2cm] (mapping)
    {\textbf{Surrogate proximal-gradient term}\\[1.8pt]
     $\displaystyle-\sum\nolimits_{t=0}^{T}\mathbb E
     \Vert\mathbf G^\alpha(\mathbf x^t,\mathbf y^t)\Vert^2$};
\node[module, right=3.5mm of mapping, text width=3.8cm] (estimation)
    {\textbf{Gradient-estimation error}\\[1.8pt]
     $\displaystyle\sum\nolimits_{t=0}^{T}\mathbb E
     \Vert\overline{\nabla\mathbf f}(\mathbf x^t)-\bar\nu^t\Vert^2$};
\node[module, right=3.5mm of estimation, text width=3.5cm] (model)
    {\textbf{Model consensus error}\\[1.8pt]
     $\displaystyle\sum\nolimits_{t=0}^{T}\mathbb E
     \Vert\mathbf J\mathbf x^t-\mathbf x^t\Vert^2$};
\node[module, right=4.5mm of model, text width=3.5cm] (tracking)
    {\textbf{Tracking consensus error}\\[1.8pt]
     $\displaystyle\sum\nolimits_{t=0}^{T}\mathbb E
     \Vert\mathbf J\mathbf y^t-\mathbf y^t\Vert^2$};

\node[result, below=5.6mm of mapping, text width=5.2cm] (positive)
    {\textbf{Positive surrogate proximal-gradient term}\\[1.8pt]
     $\displaystyle+\sum\nolimits_{t=0}^{T}\mathbb E
     \Vert\mathbf G^\alpha(\mathbf x^t,\mathbf y^t)\Vert^2$};
\node[auxiliary, below=5.6mm of model, xshift=-2.00cm,
      text width=4.2cm] (movement)
    {\textbf{Accumulated model movement}\\[1.8pt]
     $\displaystyle\sum\nolimits_{t=0}^{T}\mathbb E
     \Vert\mathbf x^{t+1}-\mathbf x^t\Vert^2$};
\node[auxiliary, below=5.6mm of tracking, text width=4.2cm] (variation)
    {\textbf{Accumulated momentum variation}\\[1.8pt]
     $\displaystyle\sum\nolimits_{t=0}^{T}\mathbb E
     \Vert\bm\nu^{t+1}-\bm\nu^t\Vert^2$};

\draw[arr] (estimation.south) --
    node[lab, left=0.6mm, pos=0.48]{\eqref{equ:PropositionP1}}
    (estimation.south |- movement.north);
\draw[arr] (movement.west) --
    node[lab, above=0.6mm]{\eqref{equ:PropositionSuccessiveX}}
    (positive.east);
\draw[looparr] (model.south |- movement.north) --
    node[lab, right=0.6mm, pos=0.56]{\eqref{equ:PropositionSuccessiveX}}
    (model.south);
\draw[looparr] (model.east) --
    node[lab, above=0.6mm]{\eqref{equ:PropositionX}}
    (tracking.west);
\draw[looparr] (tracking.south) --
    node[lab, right=0.6mm]{\eqref{equ:PropositionY}}
    (variation.north);
\draw[looparr] (variation.west) --
    node[lab, below=0.6mm, pos=0.25]{\eqref{equ:MomentumDifference}}
    (movement.east);
\coordinate (eqtwentyturn) at
    ([xshift=2.5mm,yshift=-4.0mm]variation.south east);
\coordinate (eqtwentyleft) at ([yshift=-4.0mm]positive.south);
\draw[outarr] (tracking.east) -| (eqtwentyturn) --
    node[lab, above=0.55mm, pos=0.50]
    {\textit{feedback closure: }
     \eqref{equ:PropositionSuccessiveX}--\eqref{equ:PropositionY}
     $\Rightarrow$ \eqref{equ:PropositionYG1}}
    (eqtwentyleft) -- (positive.south);
\draw[arr] (positive.north) --
    node[lab, left=0.6mm]{absorbed}
    (mapping.south);
\end{tikzpicture}
\caption{Proof roadmap for deriving Theorem \ref{t1} from Proposition \ref{prop1}.
Scalar coefficients and stochastic-variance terms are omitted. 
}
\label{fig:proofroadmap}
\end{figure*}

\begin{lem}\label{lem:GradientEstimation}
Suppose that Assumptions \ref{ass:AssumptionFunction} and \ref{ass:AssumptionStochastic} hold, and let $0\leqslant\gamma<1$.
For either Polyak or Nesterov momentum,
\begin{align}
\sum_{t=0}^{T}\mathbb E\Vert\overline{\nabla\mathbf f}(\mathbf x^t)-\bar\nu^t\Vert^2&\leqslant\frac{\gamma^2L^2}{n(1-\gamma)^2}
\sum_{t=0}^{T}\mathbb E\Vert\mathbf x^{t+1}-\mathbf x^t\Vert^2 \notag\\
&+\left[\frac{1}{1-\gamma}+\frac{5T}{4}(1-\gamma)\right]\frac{\sigma^2}{nB}.
\label{equ:PropositionP1}
\end{align}
\end{lem}

It remains to bound the successive-iterate differences together with the model consensus and tracking consensus errors in Proposition \ref{prop1}.
To state these bounds, define
\begin{equation*}
    \delta_1=\left\{ \begin{array}{ll}
\lambda(1-\lambda)[(1-\alpha\beta\rho)^2-\lambda^{\frac{1}{T_0}}], & 0<\lambda<1, \\
\frac{T_0^{T_0}(1-\alpha\beta\rho)^{2T_0+2}}{(1+T_0)^{T_0+1}}, & \lambda=0, 
\end{array} \right.
\end{equation*}
\begin{equation*} 
\delta_2=\left\{ \begin{array}{ll}
    \lambda(1-\lambda)(1-\lambda^{\frac{1}{T_0}}), & 0<\lambda<1, \\
    \frac{T_0^{T_0}}{(1+T_0)^{T_0+1}}, & \lambda=0. 
    \end{array} \right.
\end{equation*}
For \(0\leqslant\alpha\beta\rho<1-\lambda^{1/(2T_0)}\), both quantities are positive.
The factor \(\delta_1\) is the contraction margin in the model consensus recursion; it accounts for periodic mixing and the possible expansion of the weakly convex proximal operator \eqref{equ:ProximalGradientRelationMain}.
The factor \(\delta_2\) is the corresponding contraction margin in the tracking consensus recursion.
The following lemma collects the coupled bounds shown in Fig. \ref{fig:proofroadmap}.
\begin{lem}\label{lem:CoupledRelations}
Suppose that Assumptions \ref{ass:AssumptionFunction}, \ref{ass:AssumptionMixingMatrix}, and \ref{ass:AssumptionStochastic} hold.
For either Polyak or Nesterov momentum, let $0\leqslant\gamma<1$ and $0\leqslant\alpha\beta\rho<1-\lambda^{1/(2T_0)}$.
Then
\begin{align}
\sum_{t=0}^{T}\Vert\mathbf x^{t+1}-\mathbf x^t\Vert^2
\leqslant{}&\alpha^2\sum_{t=0}^{T}\Vert\mathbf G^\alpha(\mathbf x^t,\mathbf y^t)\Vert^2 \notag\\
&\qquad+4\sum_{t=0}^{T+1}\Vert\mathbf J\mathbf x^t-\mathbf x^t\Vert^2,
\label{equ:PropositionSuccessiveX}
\end{align}
\begin{align}
\sum_{t=0}^{T+1}\Vert\mathbf J\mathbf x^t-\mathbf x^t\Vert^2
\leqslant\frac{T_0\alpha^2}{\delta_1}
\sum_{t=0}^{T}\Vert\mathbf J\mathbf y^t-\mathbf y^t\Vert^2,
\label{equ:PropositionX}
\end{align}
\begin{align}
&\sum_{t=0}^{T}\mathbb E\Vert\bm\nu^{t+1}-\bm\nu^t\Vert^2 \leqslant2L^2\sum_{t=0}^{T}\mathbb E\Vert\mathbf x^{t+1}-\mathbf x^t\Vert^2 \notag\\
&\qquad\qquad+\frac{2n(1-\gamma)\sigma^2}{B}[1+8(T+1)(1-\gamma)],
\label{equ:MomentumDifference}
\end{align}
\begin{align}
\sum_{t=0}^{T+1}\mathbb E\Vert\mathbf J\mathbf y^t-\mathbf y^t\Vert^2
\leqslant\frac{T_0}{\delta_2}
\sum_{t=0}^{T}\mathbb E\Vert\bm\nu^{t+1}-\bm\nu^t\Vert^2.
\label{equ:PropositionY}
\end{align}
\end{lem}

Substituting \eqref{equ:MomentumDifference} and \eqref{equ:PropositionSuccessiveX} into \eqref{equ:PropositionY}, and then applying \eqref{equ:PropositionX}, gives an inequality with the tracking consensus error on both sides.
Under the stepsize condition below, moving its right-hand occurrence to the left gives the feedback closure.
\begin{lem}\label{lem:TrackingConsensusBound}
Under the assumptions and parameter conditions of Lemma \ref{lem:CoupledRelations}, if
$\alpha^2<\delta_1\delta_2/(8L^2T_0^2)$, then
\begin{align}
&\sum_{t=0}^{T+1}\mathbb E\Vert\mathbf J\mathbf y^t-\mathbf y^t\Vert^2 \leqslant\frac{2T_0\delta_1\alpha^2L^2}
{\delta_1\delta_2-8T_0^2(\alpha L)^2}
\sum_{t=0}^{T}\mathbb E\Vert\mathbf G^\alpha(\mathbf x^t,\mathbf y^t)\Vert^2 \notag\\
&\qquad+\frac{2nT_0\delta_1(1-\gamma)[1+8(T+1)(1-\gamma)]\sigma^2}
{B[\delta_1\delta_2-8T_0^2(\alpha L)^2]}.
\label{equ:PropositionYG1}
\end{align}
\end{lem}

The preceding analysis requires neither the bounded gradient heterogeneity condition \eqref{equ:GradientHeterogeneity} nor a bounded-gradient condition such as \(\Vert\nabla f_i(x)\Vert\leqslant G\).

After taking expectations in \eqref{equ:Proposition1}, we substitute Lemma \ref{lem:GradientEstimation}, \eqref{equ:PropositionSuccessiveX}, \eqref{equ:PropositionX}, and Lemma \ref{lem:TrackingConsensusBound}.
The right-hand side then contains only the initial objective gap, stochastic-variance terms, and a multiple of the accumulated squared norm of the surrogate proximal-gradient mapping.
Under the parameter conditions below, this multiple is absorbed by the negative surrogate proximal-gradient term in \eqref{equ:Proposition1}, as indicated by the final arrow in Fig. \ref{fig:proofroadmap}; dividing both sides by \(n(T+1)\) then gives \eqref{equ:T1i}.

\begin{thm}\label{t1}
Suppose that Assumptions \ref{ass:AssumptionFunction}, \ref{ass:AssumptionMixingMatrix} and \ref{ass:AssumptionStochastic} hold.
For either Polyak or Nesterov momentum, let $\alpha\beta\rho\leqslant\frac{1}{48}$, $\alpha\beta\rho<1-\lambda^{\frac{1}{2T_0}}$, and $8\alpha^2L^2\leqslant\min\{\frac{1}{32},\frac{(1-\gamma)^2}{72},\frac{3\delta_1\delta_2}{T_0(11\delta_1+9T_0)}\}$.
Then
\begin{align}
&\frac{1}{n(T+1)}\sum_{t=0}^T\mathbb{E}[\mathbf{s}(\mathbf{x}^t,\bar{\nu}^{t})]
\leqslant\frac{8[\phi(x_0)-\phi^*]}{\alpha(T+1)} \notag\\
&\quad
+\frac{142\sigma^2}{nB(1-\gamma)(T+1)}+\frac{2T_0(11\delta_1+6T_0)(1-\gamma)\sigma^2}
{B[\delta_1\delta_2-8T_0^2(\alpha L)^2](T+1)} \notag\\
&\quad+\frac{355(1-\gamma)\sigma^2}{2nB} +\frac{16T_0(11\delta_1+6T_0)(1-\gamma)^2\sigma^2}
{B[\delta_1\delta_2-8T_0^2(\alpha L)^2]}.
\label{equ:T1i}
\end{align}
\end{thm}

All proofs can be found in the supplementary materials.
The last two terms in \eqref{equ:T1i} show why the momentum parameter cannot simply be fixed as the total number of iterations increases: unlike the preceding stochastic terms, they contain no explicit factor \(1/(T+1)\).
Letting \(\gamma\) approach $1$ reduces these two terms, but also enlarges the first stochastic term through the factor \(1/(1-\gamma)\).
Corollary \ref{c1} resolves this trade-off by choosing \(1-\gamma=\sqrt{n/(T+1)}\), together with a batch size independent of \(T\) and an appropriate stepsize.

\begin{cor} \label{c1}
Fix $\beta>0$.
Let $T+1\geqslant n$, $B=\lceil\sqrt{n}\rceil$, $1-\gamma=\sqrt{n/(T+1)}$, and $\alpha=\sqrt{n}/[24L\sqrt{T+1}]$, subject to the conditions of Theorem \ref{t1}.
Then,
\begin{align}
&\frac{1}{n(T+1)}\sum_{t=0}^T\mathbb{E}[\mathbf{s}(\mathbf{x}^t,\bar{\nu}^{t})]
\leqslant\frac{192L[\phi(x_0)-\phi^*]}{\sqrt{n(T+1)}}+\frac{320\sigma^2}{n\sqrt{T+1}} \notag\\
&\qquad+\frac{T_0(11\delta_1+6T_0)\sigma^2}{\delta_1\delta_2-\frac{nT_0^2}{72(T+1)}} \left[\frac{2}{(T+1)^{3/2}}+\frac{16\sqrt{n}}{T+1}\right].
\label{equ:C1i}
\end{align}
Moreover, fix $L$, \(\beta\), \(\rho\), \(T_0\), and \(\lambda<1\).
For sufficiently large \(T\) satisfying the conditions of Theorem \ref{t1}, the right-hand side of \eqref{equ:C1i} is \(\mathcal{O}(1/\sqrt{nT})\) when \(T=\Omega(n^2)\), where the hidden constant depends on these fixed parameters.
Consequently, there exists an index \(\eta\in\{0,\ldots,T\}\) satisfying \(\frac{1}{n}\mathbb{E}[\mathbf{s}(\mathbf{x}^{\eta},\bar{\nu}^{\eta})]=\mathcal{O}(1/\sqrt{nT})\).
In this post-transient regime, reaching expected \(\epsilon\)-stationarity requires \(\mathcal{O}(1/(n\epsilon^2))\) iterations.
\end{cor}

The first two terms in \eqref{equ:C1i} represent the initial objective gap and network-independent stochastic variance, respectively.
The final term arises from the coupled recursions for model consensus and tracking consensus.
If its denominator remains a fixed positive fraction of \(\delta_1\delta_2\), this term is \(\mathcal{O}(T^{-3/2}+\sqrt{n}/T)\).
Since \(\sqrt{n}/T\) becomes no larger than \(1/\sqrt{nT}\) for \(T=\Omega(n^2)\), the network-related term may dominate before this threshold, up to constants depending on the fixed parameters.
Choosing \(1-\gamma=\sqrt{n/(T+1)}\) makes the normalized gradient-estimation error decay with \(T\) while keeping \(B=\lceil\sqrt n\rceil\) independent of \(T\).
With \(\gamma=0\) and fixed \(B\), \eqref{equ:PropositionP1} instead leaves an \(\mathcal{O}(\sigma^2/(nB))\) variance term.
If \(\sigma=0\), Theorem \ref{t1} gives an \(\mathcal{O}(1/T)\) deterministic rate with a fixed admissible stepsize.

With \(T_0\) local iterations between communication phases, DEPOSITUM performs \(\mathcal{O}(T/T_0)\) such phases, each comprising two elementary neighbor-mixing operations.
Increasing \(T_0\) reduces communication frequency but decreases the contraction margins \(\delta_1\) and \(\delta_2\), restricting the admissible stepsize and enlarging the network-related coefficient in Corollary \ref{c1}.
Consequently, more iterations are required before the \(1/\sqrt{nT}\) term dominates.
For example, when \(\rho=0\) and \(\lambda\) is close to $1$, \(\delta_1=\delta_2=\Theta((1-\lambda)^2/T_0)\).
Once the denominator correction \(nT_0^2/[72(T+1)]\) is negligible, the network-related coefficient can scale as \(\Theta(T_0^4/(1-\lambda)^4)\).
Periodic local updates therefore save communication phases but incur a larger network-dependent constant, particularly for large \(T_0\) or a small spectral gap \(1-\lambda\).

\begin{table*}[t]
\centering
\setlength{\tabcolsep}{2.3pt}
\renewcommand{\arraystretch}{1.1}
\begin{threeparttable}
\caption{Comparison of assumptions, complexity, and communication schedules for related methods.}
\label{tab:complexity}
\begin{tabular*}{\textwidth}{@{\extracolsep{\fill}}lccccccccc@{}}
\toprule
\multirow{2}{*}{Method} & \multirow{2}{*}{\shortstack[c]{Mean-squared\\[-1pt]smoothness}} & \multirow{2}{*}{\shortstack[c]{Bounded gradient\\[-1pt]heterogeneity}} & \multicolumn{3}{c}{Complexity} & \multirow{2}{*}{\shortstack[c]{Mega-\\[-1pt]batch}} & \multirow{2}{*}{\shortstack[c]{Linear\\speedup}} & \multicolumn{2}{c}{Communication schedule} \\[-2pt]
\cmidrule(lr){4-6}\cmidrule(l){9-10}
& & & Batch & Communication & Gradient evals./client & & & Interval & Mixing ops. \\[-1pt]
\midrule
SPPDM \cite{wang2021distributed}
& No
& No
& $\mathcal O(1/\epsilon)$
& $\mathcal O(1/\epsilon)$
& $\mathcal O(1/\epsilon^{2})$
& Yes
& No
& $1$
& $1$ \\

ProxGT-SA \cite{xin2021stochastic}
& No
& No
& $\mathcal O(1/(n\epsilon))$
& $\widetilde{\mathcal O}(1/\epsilon)$
& $\mathcal O(1/(n\epsilon^{2}))$
& Yes
& Yes
& $1$
& $2K$ \\

ProxGT-SR-O \cite{xin2021stochastic}
& Yes
& No
& $\mathcal O(1/(n\epsilon))$\tnote{*}
& $\widetilde{\mathcal O}(1/\epsilon)$
& $\mathcal O(1/(n\epsilon^{3/2}))$
& Yes
& Yes
& $1$
& $2K$ \\

DEEPSTORM \cite{mancino2023proximal}\tnote{\textdagger}
& Yes
& No
& $\mathcal O(1/\sqrt{\epsilon})/\mathcal O(1)$
& $\mathcal O(1/(n\epsilon^{3/2}))$
& $\mathcal O(1/(n\epsilon^{3/2}))$
& Yes
& Yes
& $1$
& $2K$ \\

Prox-DASA \cite{xiao2023one}
& No
& Yes
& $\mathcal O(1)$
& $\mathcal O(1/(n\epsilon^{2}))$
& $\mathcal O(1/(n\epsilon^{2}))$
& No
& Yes
& $1$
& $2K$ \\

Prox-DASA-GT \cite{xiao2023one}
& No
& No
& $\mathcal O(1)$
& $\mathcal O(1/(n\epsilon^{2}))$
& $\mathcal O(1/(n\epsilon^{2}))$
& No
& Yes
& $1$
& $3K$ \\

DProxSGT \cite{yan2023compressed}
& No
& No
& $\mathcal O(1)$
& $\mathcal O(1/\epsilon^{2})$
& $\mathcal O(1/\epsilon^{2})$
& No
& No
& $1$
& $2$ \\

\textbf{DEPOSITUM}
& No
& No
& $\mathcal O(\sqrt n)$
& $\mathcal O(1/(nT_0\epsilon^{2}))$
& $\mathcal O(1/(\sqrt n\,\epsilon^{2}))$
& No
& Yes
& $T_0$
& $2$ \\
\bottomrule
\end{tabular*}
\begin{tablenotes}[flushleft]
\scriptsize
\item Complexity orders are converted to the stationarity measure in Definition \ref{def:StationaryPoint} and report the leading dependence on $n$ and $\epsilon$.
Gradient evals./client counts component-gradient evaluations at each client, while Mixing ops. counts elementary neighbor-mixing operations within one communication phase; $K$ denotes the number of repeated mixing steps in each recursion.
All orders suppress problem- and network-dependent factors, while $\widetilde{\mathcal O}$ also suppresses logarithmic factors.
\item[*] ProxGT-SR-O uses the displayed large batch periodically within its double-loop recursion.
\item[\textdagger] For DEEPSTORM, the two batch sizes correspond to initialization and subsequent iterations.
The displayed linear-speedup result corresponds to its constant-stepsize regime.
\end{tablenotes}
\end{threeparttable}
\end{table*}

This schedule differs from representative decentralized stochastic composite methods \cite{wang2021distributed,xin2021stochastic,mancino2023proximal,xiao2023one,yan2023compressed}, which communicate at every iteration and often employ multi-step mixing.
Table \ref{tab:complexity} compares their assumptions, complexity, and communication schedules.
SPPDM, the ProxGT variants, and DEEPSTORM use accuracy-dependent large batches; ProxGT-SR-O and DEEPSTORM additionally require mean-squared smoothness.
Prox-DASA avoids large batches under bounded gradient heterogeneity, whereas Prox-DASA-GT removes this condition through an additional mixing recursion.
DProxSGT avoids these assumptions but does not establish linear speedup in \(n\).
In contrast, DEPOSITUM uses a mini-batch size independent of $T$, permits periodic local updates, requires neither mean-squared smoothness nor bounded gradient heterogeneity, and achieves linear speedup with respect to $n$ after a network-dependent transient.

\section{Numerical Experiments}\label{sec:experiments}

\begin{figure*}[t]
  \centering
  \includegraphics[width=0.58\linewidth]{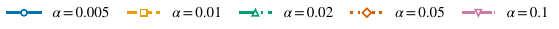}\\[-8pt]
  \subfloat[Polyak: Loss]{\includegraphics[width=0.242\linewidth]{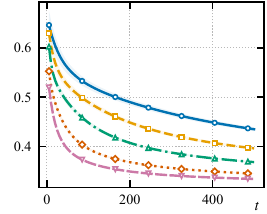}\label{fig:e1-loss}}\hfill
  \subfloat[Polyak: $\|\mathbf{G}^{\alpha}(\mathbf{x}^t)\|^2$]{\includegraphics[width=0.242\linewidth]{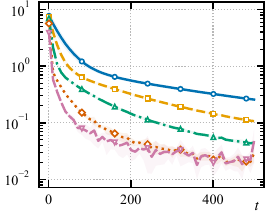}\label{fig:e1-stationarity}}\hfill
  \subfloat[Polyak: $\frac{1}{n}\|\mathbf{J}\mathbf{x}^t-\mathbf{x}^t\|^2$]{\includegraphics[width=0.242\linewidth]{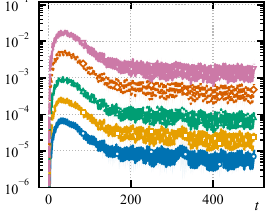}\label{fig:e1-model-consensus}}\hfill
\subfloat[Polyak: {\scriptsize $\frac{1}{|\mathcal{E}_t|}\sum\limits_{k\in\mathcal{E}_t}\|\overline{\nabla\mathbf{f}}(\mathbf{x}^k)-\bar{\nu}^k\|^2$}]{\includegraphics[width=0.242\linewidth]{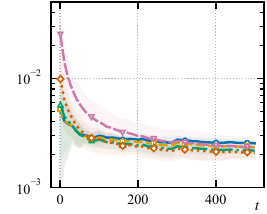}\label{fig:e1-gradient-estimation}}\\ [-8pt]
  \subfloat[Nesterov: Loss]{\includegraphics[width=0.242\linewidth]{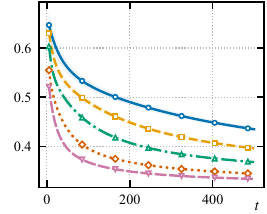}\label{fig:e1-nesterov-loss}}\hfill
  \subfloat[Nesterov: $\|\mathbf{G}^{\alpha}(\mathbf{x}^t)\|^2$]{\includegraphics[width=0.242\linewidth]{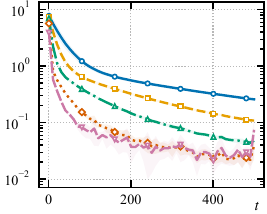}\label{fig:e1-nesterov-stationarity}}\hfill
  \subfloat[Nesterov: $\frac{1}{n}\|\mathbf{J}\mathbf{x}^t-\mathbf{x}^t\|^2$]{\includegraphics[width=0.242\linewidth]{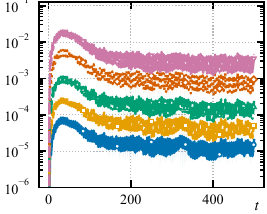}\label{fig:e1-nesterov-model-consensus}}\hfill
  \subfloat[Nesterov: {\tiny $\frac{1}{|\mathcal{E}_t|}\!\sum\limits_{k\in\mathcal{E}_t}\!\|\overline{\nabla\mathbf{f}}(\mathbf{x}^k)\!-\!\bar{\nu}^k\!\|^2$}]{\includegraphics[width=0.242\linewidth]{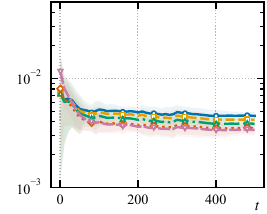}\label{fig:e1-nesterov-gradient-estimation}}
  \caption{Effect of stepsize $\alpha$ on DEPOSITUM.}
  \label{fig:e1}
\end{figure*}
\begin{figure}[t]
  \centering
  \includegraphics[width=0.98\columnwidth]{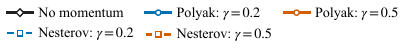}\\[-4pt]
  \subfloat[Loss]{\includegraphics[width=0.49\columnwidth]{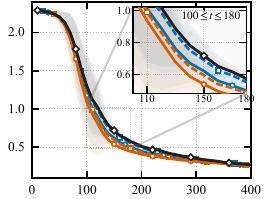}\label{fig:e2-loss}}\hspace{-0.005\columnwidth}%
  \subfloat[Test accuracy]{\includegraphics[width=0.49\columnwidth]{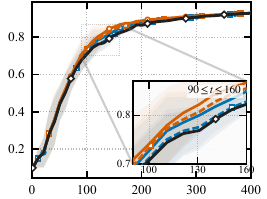}\label{fig:e2-accuracy}}
  \caption{Effect of the momentum parameter $\gamma$ on DEPOSITUM.}
  \label{fig:e2}
\end{figure}

We evaluate DEPOSITUM on a tabular binary-classification task, a convolutional neural network task, and decentralized parameter-efficient instruction fine-tuning of a language model.
The experiments first characterize the effects of the stepsize, momentum parameter, communication period, and regularization weight; they then compare DEPOSITUM with decentralized baselines under graph topologies of increasing connectivity, examine the effect of the client count, and assess its performance on the LLM fine-tuning task.

Unless otherwise stated, curves, points, and tabulated results report the mean over five independent runs, with shaded bands and error bars indicating one standard deviation.
Within each run, the compared methods use the same data partition and model initialization whenever applicable.
Detailed experimental settings, including datasets, models, data partitions, communication topologies, mixing weights, and hyperparameters, are provided in the supplementary material.\footnote{The reference implementation of DEPOSITUM is available at \url{https://github.com/Zhouxyz/DEPOSITUM}.}

\begin{figure*}[t]
  \centering
  \includegraphics[width=0.49\linewidth]{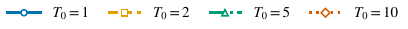}\\[-8pt]
  \subfloat[Polyak: Loss vs Commun.]{\includegraphics[width=0.242\linewidth]{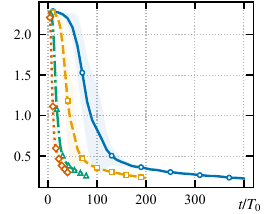}\label{fig:e3-loss}}\hfill
  \subfloat[Polyak: $\frac{1}{n}\|\mathbf{J}\mathbf{x}^t-\mathbf{x}^t\|^2$]{\includegraphics[width=0.242\linewidth]{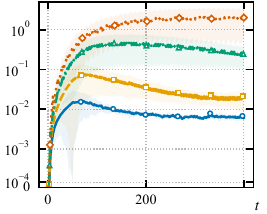}\label{fig:e3-model-consensus}}\hfill
  \subfloat[Nesterov: Loss vs Commun.]{\includegraphics[width=0.242\linewidth]{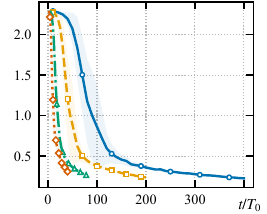}\label{fig:e3-nesterov-loss}}\hfill
  \subfloat[Nesterov: $\frac{1}{n}\|\mathbf{J}\mathbf{x}^t-\mathbf{x}^t\|^2$]{\includegraphics[width=0.242\linewidth]{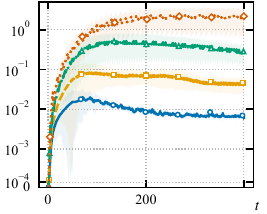}\label{fig:e3-nesterov-model-consensus}}
  \caption{Effect of the communication period $T_0$ on DEPOSITUM.}
  \label{fig:e3}
\end{figure*}
\begin{figure*}[t]
  \centering
  \includegraphics[width=0.6\linewidth]{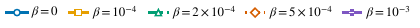}\\[-8pt]
  \subfloat[Polyak: Sparsity]{\includegraphics[width=0.242\linewidth]{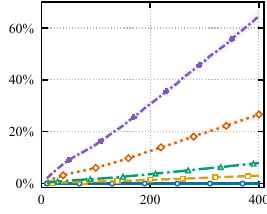}\label{fig:e3-sparsity}}\hfill
  \subfloat[Polyak: Sparsity vs Accuracy]{\includegraphics[width=0.242\linewidth]{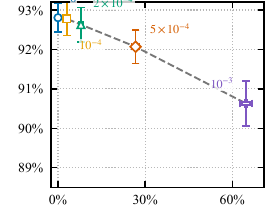}\label{fig:e3-accuracy}}\hfill
  \subfloat[Nesterov: Sparsity]{\includegraphics[width=0.242\linewidth]{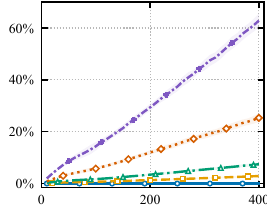}\label{fig:e3-nesterov-sparsity}}\hfill
  \subfloat[Nesterov: Sparsity vs Accuracy]{\includegraphics[width=0.242\linewidth]{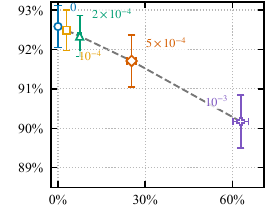}\label{fig:e3-nesterov-accuracy}}
  \caption{Effect of the regularization weight $\beta$ on DEPOSITUM.}
  \label{fig:e3-beta}
\end{figure*}
 \begin{figure*}[t]
  \centering
  \includegraphics[width=0.85\linewidth]{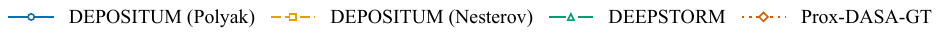}\\[-8pt]
  \subfloat[Ring: Loss]{\includegraphics[width=0.242\linewidth]{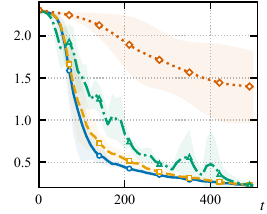}}\hfill
  \subfloat[2-hop Ring: Loss]{\includegraphics[width=0.242\linewidth]{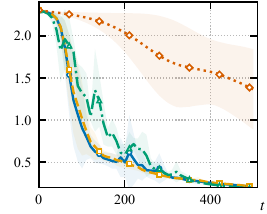}}\hfill
  \subfloat[3-hop Ring: Loss]{\includegraphics[width=0.242\linewidth]{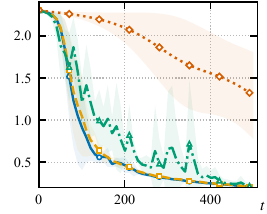}}\hfill
  \subfloat[Complete: Loss]{\includegraphics[width=0.242\linewidth]{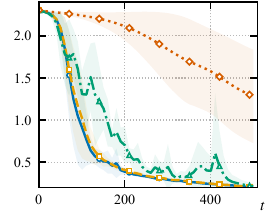}}\\[-5pt]
  \subfloat[Ring: $\frac{1}{n}\|\mathbf{J}\mathbf{x}^t-\mathbf{x}^t\|^2$]{\includegraphics[width=0.242\linewidth]{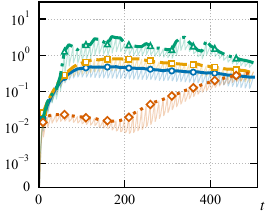}}\hfill
  \subfloat[2-hop Ring: $\frac{1}{n}\|\mathbf{J}\mathbf{x}^t-\mathbf{x}^t\|^2$]{\includegraphics[width=0.242\linewidth]{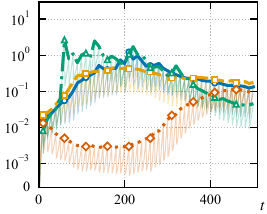}}\hfill
  \subfloat[3-hop Ring: $\frac{1}{n}\|\mathbf{J}\mathbf{x}^t-\mathbf{x}^t\|^2$]{\includegraphics[width=0.242\linewidth]{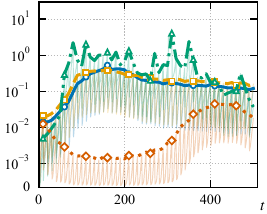}}\hfill
  \subfloat[Complete: $\frac{1}{n}\|\mathbf{J}\mathbf{x}^t-\mathbf{x}^t\|^2$]{\includegraphics[width=0.242\linewidth]{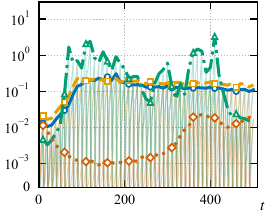}}
  \caption{Comparison under different graph topologies. In the model consensus error panels, faint curves show the per-iteration mean, while thick curves average the within-run maximum over each communication phase.}
  \label{fig:e4}
\end{figure*}
 \begin{figure*}[htbp]
  \centering
  \includegraphics[width=0.5\linewidth]{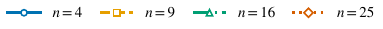}\\[-8pt]
  \subfloat[Polyak: Loss]{\includegraphics[width=0.242\linewidth]{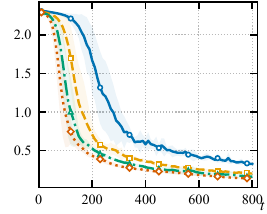}}\hfill
  \subfloat[Polyak: {\scriptsize $\frac{1}{n|\mathcal{E}_t|}\sum\limits_{k\in\mathcal{E}_t}\|\mathbf{G}^{\alpha}(\mathbf{x}^k)\|^2$}]{\includegraphics[width=0.242\linewidth]{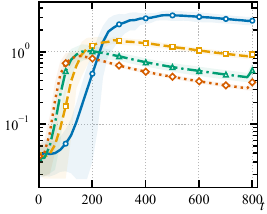}}\hfill
  \subfloat[Nesterov: Loss]{\includegraphics[width=0.242\linewidth]{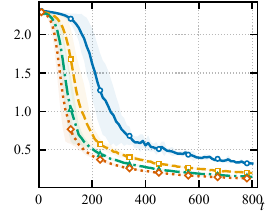}}\hfill
  \subfloat[Nesterov: {\scriptsize $\frac{1}{n\!|\mathcal{E}_t|}\sum\limits_{k\in\mathcal{E}_t}\|\mathbf{G}^{\alpha}(\mathbf{x}^k)\|^2$}]{\includegraphics[width=0.242\linewidth]{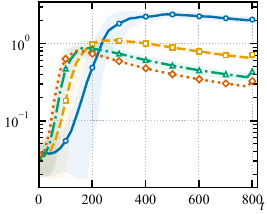}}\hfill
  \caption{Effect of the client count $n$ on DEPOSITUM.}
  \label{fig:e5}
\end{figure*}


\textbf{Effect of Stepsize $\alpha$.}
We consider a linear binary classifier trained on the A9A dataset using an independent and identically distributed (IID) partition across $10$ clients connected by a Ring graph.
The regularizer is the $\ell_1$ norm, and we fix the regularization weight at $\beta=10^{-5}$, the momentum parameter at $\gamma=0.5$, the communication period at $T_0=5$, and the batch size at $64$.

Fig. \ref{fig:e1} compares $\alpha\in\{0.005,0.01,0.02,0.05,0.1\}$ for two momentum variants over $500$ iterations.
For both momentum variants, increasing the stepsize accelerates the reduction in training loss, but increases the model consensus error.
The squared norm of the proximal-gradient mapping $\|\mathbf{G}^{\alpha}(\mathbf{x}^t)\|^2$ also decreases as the stepsize increases, although its trajectory becomes less stable for larger stepsizes.
The averaged gradient-estimation error does not exhibit a clear monotone dependence on the stepsize.

\textbf{Effect of Momentum Parameter $\gamma$.}
Momentum also affects the training performance of DEPOSITUM.
For this experiment, we train a convolutional neural network (CNN) on MNIST using an IID partition across $20$ clients connected by a Ring graph, and use the MCP regularizer.
We set $\alpha=0.05$, $\beta=10^{-5}$, $T_0=10$, and the batch size to $32$.

Fig. \ref{fig:e2} compares Polyak and Nesterov momentum with $\gamma\in\{0.2,0.5\}$ against the method without momentum.
Here and below, test accuracy is computed from the aggregated average of the local models.
Both momentum variants improve the early decrease in loss and the early increase in test accuracy relative to the no-momentum variant.
Within the tested range, $\gamma=0.5$ converges faster than $\gamma=0.2$, and Polyak momentum with $\gamma=0.5$ gives the fastest trajectory in the highlighted interval.
The differences become small later in training, and the five settings reach similar test accuracy.

\textbf{Effect of Communication Period $T_0$.}
In DEPOSITUM, the communication period specifies the number of local iterations between communication phases.
We train a CNN on MNIST using $20$ clients connected by a Ring graph.
Samples are partitioned according to $\operatorname{Dir}(1)$, which represents moderate label heterogeneity; smaller Dirichlet concentration parameters produce more heterogeneous partitions.
We use the $\ell_1$ regularizer with regularization weight $\beta=10^{-5}$, set $\alpha=0.05$ and $\gamma=0.5$, use a batch size of $32$, and compare $T_0\in\{1,2,5,10\}$ over $400$ iterations.

As $T_0$ increases, Fig.~\ref{fig:e3}(a) and (c) shift the loss trajectories toward fewer communication phases, whereas Fig.~\ref{fig:e3}(b) and (d) show substantially larger model consensus errors.
Increasing $T_0$ therefore reduces the number of communication phases, but increases the model consensus error and moderately degrades final task performance.
For example, after $400$ iterations, $T_0=2$ uses half as many communication phases as $T_0=1$, with mean test-accuracy decreases of $0.233$ and $0.403$ percentage points under Polyak and Nesterov momentum, respectively; the accuracy loss becomes larger for $T_0=5$ and $T_0=10$.
Overall, this tradeoff can be favorable in large-scale networks with limited bandwidth, where reducing the number of communication phases is important.

\textbf{Effect of Regularization Weight $\beta$.}
The regularization weight $\beta$ controls the strength of the $\ell_1$ regularizer and, through its proximal operator, the sparsity of the client models.
We fix $T_0=5$ and retain the dataset, model, network, data partition, stepsize, momentum parameter, and batch size from the communication-period experiment.
We compare $\beta\in\{0,10^{-4},2\times10^{-4},5\times10^{-4},10^{-3}\}$ over 400 iterations for both Polyak and Nesterov momentum, and measure sparsity by the mean fraction of zero parameters across the client models.

Fig. \ref{fig:e3-beta}(a) and (c) show that increasing $\beta$ produces a monotone and well-separated increase in sparsity under both momentum forms.
With $\beta=0$, the client models remain almost entirely dense throughout training.
Fig. \ref{fig:e3-beta}(b) and (d) show the corresponding accuracy--sparsity trade-off.
Relative to $\beta=0$, $\beta=5\times10^{-4}$ yields $27.242\%$ sparsity at the final iteration with a $0.763$ percentage-point decrease in mean test accuracy, whereas $\beta=10^{-3}$ yields $65.387\%$ sparsity with a $2.060$ percentage-point decrease.
Nesterov momentum exhibits a comparable trade-off, with corresponding sparsities of $26.041\%$ and $63.612\%$ and accuracy decreases of $0.867$ and $2.797$ percentage points, respectively.
Thus, under either momentum form, a modest degradation in test accuracy can yield substantially sparser client models, which is favorable for model transmission.

\textbf{Effect of Graph Topology.}
Theorem \ref{t1} implies that network connectivity affects performance.
We train a CNN on MNIST using 10 clients with samples partitioned according to $\operatorname{Dir}(1)$ and compare Ring, 2-hop Ring, 3-hop Ring, and Complete topologies, whose connectivity increases in this order.
All methods use symmetric doubly stochastic Metropolis mixing, the $\ell_1$ regularizer with regularization weight $\beta=10^{-5}$, 500 iterations, and a communication period of $T_0=10$.
For DEPOSITUM, we set $\alpha=0.05$ and $\gamma=0.5$ and evaluate both Polyak and Nesterov momentum.
For a controlled comparison under the same neighbor-mixing frequency, we use periodic local-update variants of DEEPSTORM \cite{mancino2023proximal} and Prox-DASA-GT \cite{xiao2023one} with the same communication period as DEPOSITUM.
The stepsize of each baseline is selected to obtain stable performance under the common communication schedule, with the complete settings provided in the supplementary material.
Within each run, all methods and topologies use the same client partition and model initialization.

Fig. \ref{fig:e4}(a)--(d) reports test loss, while Fig. \ref{fig:e4}(e)--(h) reports the model consensus error under different graph topologies.
For both momentum forms of DEPOSITUM, increasing connectivity generally reduces the model consensus error and accelerates the decrease in test loss.
Under Complete mixing, each communication phase restores model consensus to numerical precision, while the intervening local iterations produce periodic deviations.
Across the four topologies, both DEPOSITUM variants achieve lower test-loss trajectories than the two baselines.
In particular, DEEPSTORM exhibits pronounced late-stage fluctuations, whereas DEPOSITUM remains more stable under the same periodic local-update schedule.
These results show that DEPOSITUM can effectively exploit local updates while maintaining robust task performance.

\textbf{Effect of Client Count $n$.}
Corollary \ref{c1} establishes linear speedup with respect to $n$ under the stated parameter schedules and after a network-dependent transient.
To examine the corresponding client-count trend, we train a CNN on MNIST with IID partitions, the MCP regularizer, regularization weight $\beta=10^{-5}$, and a Complete topology.
We set $T=800$, $T_0=10$, and compare $n\in\{4,9,16,25\}$ using
\[
B_n=\lceil\sqrt n\rceil,\quad
\alpha_n=0.05\sqrt{\frac{n}{16}},\quad
1-\gamma_n=\sqrt{\frac{n}{T+1}}.
\]

For both momentum choices, Fig. \ref{fig:e5}(a) and (c) show that increasing $n$ accelerates the decrease in loss.
Fig. \ref{fig:e5}(b) and (d) report the cumulative mean of the normalized squared norm of the exact proximal-gradient mapping, which attains lower values in the later stage for larger client counts.
The consistent trends under Polyak and Nesterov momentum show that increasing the number of clients improves the optimization performance in this setting.

\begin{table*}[thb]
\centering
\caption{Response perplexity (PPL) on the Dolly-15k test tasks. PPL measures how well a model predicts the target response tokens; lower values are better. ``Overall'' pools all test response tokens, while the remaining columns report PPL for detailed task categories. Results report the mean $\pm$ sample standard deviation over five independent runs.}
\label{tab:llm}
\setlength{\tabcolsep}{3.2pt}
\resizebox{\textwidth}{!}{%
\begin{tabular}{lccccccccc}
\toprule
Algorithm & Overall & Brainstorm. & Class. & Closed QA & Creative & General QA & Info. Ext. & Open QA & Summ. \\
\midrule
DEPOSITUM-Polyak
& $9.045\!\pm\!0.143$
& $10.496\!\pm\!0.842$
& $\mathbf{3.291\!\pm\!0.195}$
& $\mathbf{4.775\!\pm\!0.530}$
& $14.031\!\pm\!1.571$
& $11.140\!\pm\!0.263$
& $\mathbf{5.573\!\pm\!0.860}$
& $9.279\!\pm\!0.329$
& $\mathbf{5.238\!\pm\!0.934}$ \\
DEPOSITUM-Nesterov
& $\mathbf{9.045\!\pm\!0.142}$
& $\mathbf{10.495\!\pm\!0.842}$
& $\mathbf{3.291\!\pm\!0.195}$
& $4.776\!\pm\!0.528$
& $\mathbf{14.028\!\pm\!1.570}$
& $\mathbf{11.140\!\pm\!0.261}$
& $5.575\!\pm\!0.861$
& $\mathbf{9.278\!\pm\!0.329}$
& $5.239\!\pm\!0.934$ \\
DEEPSTORM
& $9.095\!\pm\!0.141$
& $10.557\!\pm\!0.846$
& $3.324\!\pm\!0.195$
& $4.903\!\pm\!0.549$
& $14.045\!\pm\!1.559$
& $11.175\!\pm\!0.265$
& $5.727\!\pm\!0.939$
& $9.313\!\pm\!0.325$
& $5.325\!\pm\!0.952$ \\
Prox-DASA-GT
& $9.277\!\pm\!0.143$
& $10.878\!\pm\!0.890$
& $3.510\!\pm\!0.190$
& $5.142\!\pm\!0.594$
& $14.166\!\pm\!1.554$
& $11.289\!\pm\!0.271$
& $6.002\!\pm\!1.002$
& $9.421\!\pm\!0.349$
& $5.471\!\pm\!0.967$ \\
\midrule
GUT ($\beta=0$)
& $9.051\!\pm\!0.142$
& $10.498\!\pm\!0.843$
& $3.296\!\pm\!0.194$
& $4.805\!\pm\!0.537$
& $14.025\!\pm\!1.565$
& $11.144\!\pm\!0.265$
& $5.606\!\pm\!0.880$
& $9.280\!\pm\!0.333$
& $5.260\!\pm\!0.958$ \\
D-PSGD ($\beta=0$)
& ${9.062\!\pm\!0.142}$
& ${10.511\!\pm\!0.841}$
& ${3.303\!\pm\!0.195}$
& $4.841\!\pm\!0.534$
& ${14.026\!\pm\!1.566}$
& ${11.150\!\pm\!0.262}$
& $5.649\!\pm\!0.895$
& ${9.287\!\pm\!0.326}$
& $5.285\!\pm\!0.946$ \\
DSGT ($\beta=0$)
& $9.080\!\pm\!0.145$
& $10.531\!\pm\!0.842$
& $3.304\!\pm\!0.171$
& $4.806\!\pm\!0.526$
& $14.064\!\pm\!1.530$
& $11.178\!\pm\!0.268$
& $5.617\!\pm\!0.849$
& $9.317\!\pm\!0.388$
& $5.281\!\pm\!0.951$ \\
\bottomrule
\end{tabular}%
}
\end{table*}

\textbf{LLM Instruction Fine-Tuning.}
We further evaluate DEPOSITUM on decentralized parameter-efficient instruction fine-tuning with Qwen2.5-0.5B Base and rank-8 low-rank adaptation (LoRA) modules on Dolly-15k.
We use three IID clients connected by a Complete topology, a mini-batch size of 4 per client, a maximum sequence length of 192, and 500 training iterations.
Within each run, all methods use the same stratified train, validation, and test split as well as the same IID client partition.
The four rows above the horizontal rule in Table \ref{tab:llm} use the MCP regularizer with regularization weight $\beta=10^{-5}$, whereas the others use the smooth setting $h\equiv0$.

Table \ref{tab:llm} reports response-token perplexity on the held-out test data for each Dolly-15k task category.
In the composite-objective block, DEPOSITUM with either Polyak or Nesterov momentum achieves the lowest reported mean PPL in every task category, while DEEPSTORM remains competitive and Prox-DASA-GT yields higher mean PPL.
In the smooth block, DEPOSITUM-Polyak remains competitive with GUT \cite{aketi2023global}, D-PSGD \cite{lian2017can}, and DSGT \cite{pu2021distributed}, which are primarily designed for smooth optimization problems.
These results show that DEPOSITUM is effective for decentralized parameter-efficient instruction fine-tuning under both the composite and smooth settings.

\section{Conclusion}\label{sec:conclusion}
We investigate DNCFL with smooth nonconvex local functions and a common nonsmooth weakly convex regularizer.
The proposed DEPOSITUM integrates proximal updates, gradient tracking of momentum estimates, and periodic local computation, while supporting both Polyak and Nesterov momentum.
Without assuming bounded gradient heterogeneity or mean-squared smoothness, we establish an $\mathcal{O}(1/\epsilon^2)$ iteration complexity for reaching an expected \(\epsilon\)-stationary point.
With $B=\lceil\sqrt n\rceil$ and appropriate stepsize and momentum schedules, the averaged stationarity measure is $\mathcal{O}(1/\sqrt{nT})$ for $T=\Omega(n^2)$ after a network-dependent transient.
Experiments across different datasets, models, regularizers, data partitions, network topologies, and client populations support the theoretical findings and demonstrate the competitive performance of DEPOSITUM.
The current guarantees assume a fixed connected undirected network and noiseless information exchange; extending them to time-varying or directed networks and imperfect communication, including additive communication noise, remains future work.

\bibliographystyle{IEEEtran}       
\bibliography{ref}   

\begin{IEEEbiographynophoto}
{Yuan Zhou} received the B.S. degree in cyberspace security from Southeast University, Nanjing, China, in 2021.
He is currently pursuing the Ph.D. degree in School of Cyber Science and Engineering, Southeast University, Nanjing, China.
His current research focuses on distributed optimization, smart grid and federated learning.
\end{IEEEbiographynophoto}
\vspace{-15pt}
\begin{IEEEbiographynophoto}
{Xinli Shi}(Senior Member) received the B.S. degree in software engineering, the M.S. degree in applied mathematics and the Ph.D. degree in control science and engineering from Southeast University, Nanjing, China, in 2013, 2016 and 2019, respectively.
He was the recipient of the Outstanding Ph.D. Degree Thesis Award from Jiangsu Province, China.
He is currently an associate professor at Southeast University.
He is a recipient of the Australian Research Council Discovery Early Career Researcher Award.
	
His current research interests include distributed optimization, reinforcement learning, and network control systems.
\end{IEEEbiographynophoto}
\vspace{-15pt}
\begin{IEEEbiographynophoto}
{Xuelong Li} (Fellow, IEEE) is the CTO and Chief Scientist of China Telecom, where he founded the Institute of Artificial Intelligence (TeleAI) of China Telecom.
\end{IEEEbiographynophoto}
\vspace{-15pt}
\begin{IEEEbiographynophoto}
{Jiachen Zhong} received the B.S. degree in network engineering from Nanjing Agricultural University, Nanjing, China, in 2022.
He is currently pursuing the M.S. degree in School of Cyber Science and Engineering, Southeast University, Nanjing, China.
His current research focuses on distributed optimization and federated learning.
\end{IEEEbiographynophoto}
\vspace{-15pt}
\begin{IEEEbiographynophoto}
{Guanghui Wen} (Senior Member, IEEE) received the Ph.D. degree in mechanical systems and control from Peking University, Beijing, China, in 2012.

Currently, he is a Young Endowed Chair Professor with the Department of Systems Science, Southeast University, Nanjing, China.
His current research interests include autonomous intelligent systems, complex networked systems, distributed control and optimization, resilient control, and distributed reinforcement learning.

Prof. Wen was the recipient of the National Science Fund for Distinguished Young Scholars, and Australian Research Council Discovery Early Career Researcher Award.
He is a reviewer for American Mathematical Review and is an active reviewer for many journals.
He currently serves as an Associate Editor of the IEEE Transactions on Industrial Informatics, the IEEE Transactions on Neural Networks and Learning Systems, the IEEE Transactions on Intelligent Vehicles, the IEEE Journal of Emerging and Selected Topics in Industrial Electronics, the IEEE Transactions on Systems, Man and Cybernetics: Systems, the IEEE Open Journal of the Industrial Electronics Society, and the Asian Journal of Control.
Prof. Wen has been named a Highly Cited Researcher by Clarivate Analytics since 2018.
He is an IET Fellow.

\end{IEEEbiographynophoto}
\vspace{-15pt}
\begin{IEEEbiographynophoto}
{Jinde Cao} (Fellow, IEEE) received the B.S. degree in mathematics from Anhui Normal University, Wuhu, China in 1986, the M.S. degree from Yunnan University, Kunming, China, and the Ph.D. degree from Sichuan University, Chengdu, China, both in applied mathematics, 1989, and 1998, respectively.
He was a Postdoctoral Research Fellow at the Department of Automation and Computer-Aided Engineering, Chinese University of Hong Kong, Hong Kong, China from 2001 to 2002.

Professor Cao is an Endowed Chair Professor, the Dean of Science Department and the Director of the Research Center for Complex Systems and Network Sciences at Southeast University (SEU).
He is also the Director of the National Center for Applied Mathematics at SEU-Jiangsu of China and the Director of the Jiangsu Provincial Key Laboratory of Networked Collective Intelligence of China.
He is also Honorable Professor of Institute of Mathematics and Mathematical Modeling, Almaty, Kazakhstan.

Prof. Cao was a recipient of the National Innovation Award of China, IETI Annual Scientific Award, Obada Prize and the Highly Cited Researcher Award in Engineering, Computer Science, and Mathematics by Clarivate Analytics.
He is elected as a member of Russian Academy of Sciences, a member of the Academia Europaea (Academy of Europe), a member of Russian Academy of Engineering, a member of the European Academy of Sciences and Arts, a member of the Lithuanian Academy of Sciences, a fellow of African Academy of Sciences, and a fellow of Pakistan Academy of Sciences.
\end{IEEEbiographynophoto}

\vfill

\clearpage
\onecolumn
\setcounter{lem}{0}
\setcounter{thm}{0}
\setcounter{assum}{0}
\setcounter{rem}{0}
\setcounter{prop}{0}
\setcounter{defn}{0}
\setcounter{cor}{0}
\numberwithin{equation}{section}

\begin{center}
{\LARGE\bfseries Supplementary Material for\\[4pt]
``Decentralized Nonconvex Composite Federated Learning with Gradient Tracking and Momentum''}
\end{center}
\vspace{1em}
\appendices
\section{Technical Preliminaries}
This section collects the algebraic, weak-convexity, and mixing properties used in the subsequent proofs.

\begin{lem}\label{lem:VectorEquation}
    For all $x,y\in\mathbb{R}^{d}$, the following identities and inequalities hold:
    \begin{equation}\label{lem:VectorEquation1}
        \langle x,y\rangle=\frac{1}{2}{\Vert x \Vert}^2+\frac{1}{2}{\Vert y \Vert}^2-\frac{1}{2}{\Vert x-y \Vert}^2; \tag*{i.}
    \end{equation}
    \begin{equation}\label{lem:VectorEquation2}
        \langle x,y\rangle\leqslant \frac{1}{2\tau}{\Vert x \Vert}^2+\frac{\tau}{2}{\Vert y \Vert}^2, \quad \tau>0; \tag*{ii.}
    \end{equation}
    \begin{equation}\label{lem:VectorEquation3}
        \begin{split}
            {\Vert x+y \Vert}^2\leqslant\tau^{-1}{{\Vert x \Vert}^2}+{(1-\tau)}^{-1}{{\Vert y \Vert}^2}, \quad 0<\tau<1; \\
            {\Vert x+y \Vert}^2\leqslant(1+\tau){\Vert x \Vert}^2+(1+\tau^{-1}){\Vert y \Vert}^2, \quad \tau>0;
        \end{split}\tag*{iii.}
    \end{equation} 
    \begin{equation}\label{lem:VectorEquation4}
    -{\Vert x \Vert}^2\leqslant {\Vert x-y \Vert}^2-\frac{1}{2}{\Vert y \Vert}^2. \tag*{iv.}   
    \end{equation}
    \end{lem}

\begin{lem}\label{lem:WeaklyConvex}
    Let $h:\mathbb{R}^d\rightarrow\mathbb{R}\cup\{+\infty\}$ be proper, closed, and $\rho$-weakly convex. Then:
    \begin{equation}\label{lem:WeaklyConvexi}
        h(y)\leqslant h(x)+\langle v, y-x \rangle+\frac{\rho}{2}{\Vert y-x \Vert}^2,
        \quad x,y\in\operatorname{dom}h,\ v\in\partial h(y); \tag*{i.}
    \end{equation}
    \begin{equation}\label{lem:WeaklyConvexii}
        h(\bar{x})\leqslant\frac{1}{n}\sum_{i=1}^nh(x_i)+\frac{\rho}{2n^2}\sum_{i=1}^{n-1}\sum_{j=i+1}^n{\Vert x_i-x_j \Vert}^2,
        \quad x_i\in\operatorname{dom}h,\ \bar{x}=\frac{1}{n}\sum_{i=1}^nx_i; \tag*{ii.}
    \end{equation}
\begin{equation}\label{lem:WeaklyConvexiii}
    \Vert \textbf{prox}_{h}^{\tau^{-1}}\{x\}-\textbf{prox}_{h}^{\tau^{-1}}\{y\}\Vert\leqslant \frac{1}{1-\tau\rho} \Vert x-y\Vert,
    \quad x,y\in\mathbb{R}^{d},\ \tau>0,\ \tau\rho<1.  \tag*{iii.}
    \end{equation}    
\end{lem}

Here, $\partial h$ denotes the limiting subdifferential. With $\psi(\cdot)=h(\cdot)+\frac{\rho}{2}{\Vert \cdot\Vert}^2$, weak convexity implies that $\psi$ is convex and $\partial h(y)=\partial\psi(y)-\rho y$, which yields Lemma \ref{lem:WeaklyConvex}.\ref{lem:WeaklyConvexi} \cite{chen2021distributed}. Lemma \ref{lem:WeaklyConvex}.\ref{lem:WeaklyConvexii} is a special case of \cite[Lemma II.1]{chen2021distributed}. Lemma \ref{lem:WeaklyConvex}.\ref{lem:WeaklyConvexiii} follows by applying the nonexpansiveness of a convex proximal operator to an equivalent convexified proximal problem; see \cite[Theorem 6.42]{beck2017first} and \cite[Lemma II.8]{chen2021distributed}. When $\rho=0$, these results reduce to their standard convex counterparts.

Let $\alpha>0$ and $\beta>0$ satisfy $\alpha\beta\rho<1$. The corresponding proximal subproblem is strongly convex and hence has a unique minimizer. Moreover, multiplying its objective by $\beta$ does not change the minimizer, so
\begin{equation*}
\textbf{prox}_{h}^{1/(\alpha\beta)}\{u\}
=\arg\min_z\left\{h(z)+\frac{1}{2\alpha\beta}\Vert z-u\Vert^2\right\}
=\arg\min_z\left\{\beta h(z)+\frac{1}{2\alpha}\Vert z-u\Vert^2\right\}.
\end{equation*}
Thus, the algorithm applies a proximal-gradient step of length $\alpha$ to $f+\beta h$, and the regularization term $\beta h$ is $\beta\rho$-weakly convex.

For $T_0\in\mathbb N_+$, define the communication schedule and the scheduled mixing matrix by
\begin{equation*}
\mathcal T=\{kT_0:k=0,1,\ldots\},
\qquad
\mathbf W^t=\begin{cases}
\mathbf W, & t\in\mathcal T,\\
\mathbf I_{nd}, & t\notin\mathcal T.
\end{cases}
\end{equation*}
Thus, each proof block starts with one mixing step followed by at most $T_0-1$ local steps. Define
\begin{equation*}
\lambda_t=\begin{cases}
\lambda, & t\in\mathcal T,\\
1, & t\notin\mathcal T.
\end{cases}
\end{equation*}

\begin{lem}\label{lem:MatrixW}
    For the time-varying matrix $\mathbf{W}^t$, it holds that
    \begin{equation}\label{lem:MatrixW1}
        \mathbf{J}\mathbf{W}^t=\mathbf{W}^t\mathbf{J}=\mathbf{J}\mathbf{W}=\mathbf{W}\mathbf{J}=\mathbf{J}; \tag*{i.}
    \end{equation}
    \begin{equation}\label{lem:MatrixW2}
        \Vert\mathbf{W}^t-\mathbf{J}\Vert\leqslant\lambda_t; \tag*{ii.}
    \end{equation}
    \begin{equation}\label{lem:MatrixW3}
       \Vert \mathbf{W}^t\mathbf{x}-\mathbf{J}\mathbf{x} \Vert \leqslant \lambda_t \Vert \mathbf{x}-\mathbf{J}\mathbf{x} \Vert, \quad \mathbf{x}\in\mathbb{R}^{nd}. \tag*{iii.}
    \end{equation}
\end{lem}
\begin{IEEEproof}
    Part (i) follows from the double stochasticity of $\mathbf W$ and the definition of $\mathbf W^t$. For part (ii), $\Vert\mathbf W-\mathbf J\Vert=\lambda$ at communication iterations, whereas $\Vert\mathbf I_{nd}-\mathbf J\Vert\leqslant1$ at local iterations. If $\mathbf{W}^t=\mathbf{I}_{nd}$, part (iii) holds directly. If $\mathbf{W}^t=\mathbf W$, then
        \begin{equation*}
        \Vert \mathbf{W}\mathbf{x}-\mathbf{J}\mathbf{x} \Vert=\Vert (\mathbf{W}-\mathbf{J})(\mathbf{x}-\mathbf{J}\mathbf{x}) \Vert \leqslant \Vert \mathbf{W}-\mathbf{J}\Vert\Vert\mathbf{x}-\mathbf{J}\mathbf{x}\Vert=\lambda \Vert \mathbf{x}-\mathbf{J}\mathbf{x} \Vert,
    \end{equation*} 
    which proves part (iii).
\end{IEEEproof}

Next, we can split \eqref{equ:Update4} into two separate equations:
\begin{subequations}
\begin{align}
\mathbf{x}^{t+\frac{1}{2}}&=\textbf{prox}_{\mathbf{h}}^{1/(\alpha\beta)}\{\mathbf{x}^t-\alpha\mathbf{y}^{t}\}=\mathbf{x}^t-\alpha\underbrace{\left[\frac{1}{\alpha}\left(\mathbf{x}^t-\textbf{prox}_{\mathbf{h}}^{1/(\alpha\beta)}\{\mathbf{x}^t-\alpha\mathbf{y}^{t}\}\right)\right]}_{\mathbf{G}^{\alpha}(\mathbf{x}^t,\mathbf{y}^{t})}=\mathbf{x}^t-\alpha \mathbf{G}^{\alpha}(\mathbf{x}^t,\mathbf{y}^{t}), \label{equ:Adopt}\\
\mathbf{x}^{t+1}&=\mathbf{W}^t\mathbf{x}^{t+\frac{1}{2}}. \label{equ:Combine} 
\end{align}
\end{subequations}
Multiplying both sides of \eqref{equ:Combine} by $\mathbf{J}$, we have
$$
\mathbf{J}\mathbf{x}^{t+1}=\mathbf{J}\mathbf{W}^t\mathbf{x}^{t+\frac{1}{2}}=\mathbf{J}\mathbf{x}^{t+\frac{1}{2}},
$$
where Lemma \ref{lem:MatrixW}.\ref{lem:MatrixW1} gives $\mathbf{J}(\mathbf{I}_{nd}-\mathbf{W}^t)=\mathbf{0}$.
Substituting \eqref{equ:Adopt} into \eqref{equ:Combine} and taking averages yields
\begin{equation}\label{equ:ProximalGradientEqu2}
    \bar{x}^{t+1}=\bar{x}^{t+\frac{1}{2}}=\bar{x}^{t}-\frac{\alpha}{n}\sum_{i=1}^n G^{\alpha}(x_i^t,y_i^{t}). 
\end{equation}
Consistently, the exact proximal-gradient mapping for $f+\beta h$ is
\begin{equation*}
G^{\alpha}(x)=\frac{1}{\alpha}\left[x-\textbf{prox}_{h}^{1/(\alpha\beta)}\{x-\alpha\nabla f(x)\}\right].
\end{equation*}
The mapping $G^{\alpha}(x_i^t,y_i^t)$ and the proximal-gradient mapping $G^{\alpha}(x_i^t)$ satisfy
\begin{align}
    {\Vert G^{\alpha}(x_i^t,y_i^t)-G^{\alpha}(x_i^t)\Vert}^2 ={}&{\left\Vert \frac{1}{\alpha}\textbf{prox}_{h}^{1/(\alpha\beta)}\{{x}_i^t-\alpha y_i^t\}-\frac{1}{\alpha}\textbf{prox}_{h}^{1/(\alpha\beta)}\{{x}_i^t-\alpha\nabla f({x}_i^{t})\}\right\Vert}^2 \notag \\
    {\leqslant}&\frac{{\Vert y_i^t-\nabla f({x}_i^{t})\Vert}^2}{(1-\alpha\beta\rho)^2}, \label{equ:ProximalGradientRelation}
\end{align}
where Lemma \ref{lem:WeaklyConvex}.\ref{lem:WeaklyConvexiii} is used in the last inequality.

For completeness, we restate the initialization and remaining algorithmic updates used throughout the subsequent proofs.
Given a common initial model $x_0$, each client forms an initial mini-batch gradient and initializes
\begin{align*}
x_i^0=x_0, \qquad \mu_i^0=\nu_i^0=g_i^0=\frac{1}{B}\sum_{b=1}^B\nabla F_i(x_0,\xi_{i,b}^0), \qquad
y_i^0=\bar g^0=\frac{1}{n}\sum_{j=1}^n g_j^0.
\end{align*}
At iteration $t$, the clients first update their models and evaluate fresh stochastic gradients:
\begin{align}
\mathbf{x}^{t+1}
    &=\mathbf{W}^t\textbf{prox}_{\mathbf h}^{1/(\alpha\beta)}
\{\mathbf{x}^t-\alpha\mathbf{y}^t\},
\tag{\ref{equ:Update4}}\\
g_i^{t+1}
&=\frac{1}{B}\sum_{b=1}^B
\nabla F_i(x_i^{t+1},\xi_{i,b}^{t+1}),
\qquad i\in\mathcal V. \notag
\end{align}
Here, $\alpha$ is the stepsize used in the proximal update, whereas $\beta$ controls the regularization strength through the proximal operator.
For Polyak momentum, the momentum estimates are updated by
\begin{equation}
\bm\nu^{t+1}=\gamma\bm\nu^t+(1-\gamma)\mathbf g^{t+1}.
\tag{\ref{equ:Update1}}
\end{equation}
For Nesterov momentum, they are updated by
\begin{align}
\bm\mu^{t+1}&=\gamma\bm\mu^t+(1-\gamma)\mathbf g^{t+1},
\tag{\ref{equ:Update2}}\\
\bm\nu^{t+1}&=\gamma\bm\mu^{t+1}+(1-\gamma)\mathbf g^{t+1}.
\tag{\ref{equ:Update3}}
\end{align}
Finally, the gradient-tracking variables are updated according to
\begin{equation}
\mathbf y^{t+1}=\mathbf W^t
(\mathbf y^t+\bm\nu^{t+1}-\bm\nu^t).
\tag{\ref{equ:Update5}}
\end{equation}
Since $\mathbf J\mathbf W^t=\mathbf J$ and
$\mathbf J\mathbf y^0=\mathbf J\bm\nu^0$, these updates preserve
$\mathbf J\mathbf y^t=\mathbf J\bm\nu^t$ for every $t\geqslant0$.

For later use, define
          \begin{align*}
     \delta_1=\left\{ \begin{array}{ll}
    \lambda(1-\lambda)\left[(1-\alpha\beta\rho)^2-\lambda^{\frac{1}{T_0}}\right], & 0<\lambda<1 \\
    \frac{T_0^{T_0}(1-\alpha\beta\rho)^{2T_0+2}}{(1+T_0)^{T_0+1}}, & \lambda=0 
    \end{array} \right.,\ 
     \delta_2=\left\{ \begin{array}{ll}
    \lambda(1-\lambda)(1-\lambda^{\frac{1}{T_0}}), & 0<\lambda<1 \\
    \frac{T_0^{T_0}}{(1+T_0)^{T_0+1}}, & \lambda=0 
    \end{array} \right..
\end{align*}
Under $0\leqslant\alpha\beta\rho<1-\lambda^{1/(2T_0)}$, both $\delta_1$ and $\delta_2$ are positive.
\section{Proof of Proposition \ref{prop1}}
We recall the stationarity measure
\begin{equation*}
\mathbf{s}(\mathbf{x},\bar\nu)
=\Vert\mathbf{G}^{\alpha}(\mathbf{x})\Vert^2
+L^2\Vert\mathbf{J}\mathbf{x}-\mathbf{x}\Vert^2
+n\Vert\overline{\nabla\mathbf f}(\mathbf{x})-\bar\nu\Vert^2.
\end{equation*}
\begin{prop}
   Suppose that Assumptions \ref{ass:AssumptionFunction} and \ref{ass:AssumptionMixingMatrix} hold. If $\alpha L\leqslant\frac{1}{16}$ and $\alpha\beta\rho\leqslant\frac{1}{48}$, then
\begin{equation}\tag{\ref{equ:Proposition1}}
        \begin{split}
\sum_{t=0}^T\mathbf{s}(\mathbf{x}^t,\bar{\nu}^{t})\leqslant&-\sum_{t=0}^T{\Vert \mathbf{G}^{\alpha}(\mathbf{x}^t,\mathbf{y}^{t})\Vert}^2+\frac{8n}{\alpha}\left[\phi({x}_{0})-\phi^*\right]+142n\sum_{t=0}^T{\Vert \overline{\nabla \mathbf{f}}(\mathbf{x}^{t})-\bar{\nu}^{t}\Vert}^2 \\
&+\frac{5}{\alpha^2}\sum_{t=0}^T{\Vert \mathbf{J}\mathbf{x}^{t}-\mathbf{x}^{t}\Vert}^2+11\sum_{t=0}^T{\Vert \mathbf{J}\mathbf{y}^{t}-\mathbf{y}^{t}\Vert}^2. 
\end{split}
\end{equation}
\end{prop}

\begin{IEEEproof}
Since $\mathbf{J}\mathbf{y}^0=\mathbf{J}\bm{\nu}^0$ and $\mathbf{J}\mathbf{W}^t=\mathbf{J}$, \eqref{equ:Update5} implies $\mathbf{J}\mathbf{y}^t=\mathbf{J}\bm{\nu}^t$, and therefore $\bar y^t=\bar\nu^t$ for every $t\geqslant0$. For agent $i$, the condition $\alpha\beta\rho<1$ makes the proximal subproblem in \eqref{equ:Adopt} strongly convex. Its optimality condition is
\begin{equation*}
0\in\partial h(x_i^{t+\frac{1}{2}})
+\frac{1}{\alpha\beta}(x_i^{t+\frac{1}{2}}-x_i^t)
+\frac{1}{\beta}y_i^t.
\end{equation*}
Hence,
\begin{equation*}
h'(x_i^{t+\frac{1}{2}})
=-\frac{1}{\alpha\beta}(x_i^{t+\frac{1}{2}}-x_i^t)-\frac{1}{\beta}y_i^t
\in\partial h(x_i^{t+\frac{1}{2}}).
\end{equation*}
Multiplying Lemma \ref{lem:WeaklyConvex}.\ref{lem:WeaklyConvexi} by $\beta$ and substituting this subgradient yields
\begin{align}
\beta h(x_i^{t+\frac{1}{2}}){\leqslant}& \beta h(x)+\beta\langle h'(x_i^{t+\frac{1}{2}}), x_i^{t+\frac{1}{2}}-x\rangle+\frac{\beta\rho}{2}{\Vert x_i^{t+\frac{1}{2}}-x\Vert}^2 \notag\\
    {=}&\beta h(x)+\frac{\beta\rho}{2}{\Vert x_i^{t+\frac{1}{2}}-x\Vert}^2-\langle y_i^t, x_i^{t+\frac{1}{2}}-x\rangle-\frac{1}{\alpha}\langle x_i^{t+\frac{1}{2}}-x_i^{t}, x_i^{t+\frac{1}{2}}-x\rangle  \notag\\
    \leqslant& \beta h(x)-\langle \bar{\nu}^{t}, x_i^{t+\frac{1}{2}}-{x}\rangle+\frac{\alpha}{2}{\Vert \bar{\nu}^{t}-y_i^t\Vert}^2+\frac{\beta\rho}{2}{\Vert x_i^{t+\frac{1}{2}}-x\Vert}^2+\frac{1}{2\alpha}{\Vert x_i^t-x\Vert}^2-\frac{1}{2\alpha}{\Vert x_i^{t+\frac{1}{2}}-x_i^{t}\Vert}^2. \label{equ:HRelationLocal}
\end{align}
The last inequality in \eqref{equ:HRelationLocal} follows from
\begin{align*}
-\langle y_i^t, x_i^{t+\frac{1}{2}}-{x}\rangle=& -\langle \bar{\nu}^{t}, x_i^{t+\frac{1}{2}}-{x}\rangle +\langle \bar{\nu}^{t}-y_i^t, x_i^{t+\frac{1}{2}}-{x}\rangle \\
    \leqslant& -\langle \bar{\nu}^{t}, x_i^{t+\frac{1}{2}}-{x}\rangle+\frac{\alpha}{2}{\Vert \bar{\nu}^{t}-y_i^t\Vert}^2+\frac{1}{2\alpha}{\Vert x_i^{t+\frac{1}{2}}-{x}\Vert}^2,
    \end{align*}
     \begin{equation*}
        -\frac{1}{\alpha}\langle x_i^{t+\frac{1}{2}}-x_i^{t}, x_i^{t+\frac{1}{2}}-x\rangle=\frac{1}{2\alpha}{\Vert x_i^t-x\Vert}^2-\frac{1}{2\alpha}{\Vert x_i^{t+\frac{1}{2}}-x_i^t\Vert}^2-\frac{1}{2\alpha}{\Vert x_i^{t+\frac{1}{2}}-x\Vert}^2, 
\end{equation*}
where the inequality uses Lemma \ref{lem:VectorEquation}.\ref{lem:VectorEquation2}, and the identity uses Lemma \ref{lem:VectorEquation}.\ref{lem:VectorEquation1}.
Setting $x=\bar{x}^t$ and averaging \eqref{equ:HRelationLocal} over all agents gives a bound in terms of the local proximal points. Applying weak convexity once more relates this bound to $\beta h(\bar{x}^{t+1})$:
\begin{align}
 &\beta h(\bar{x}^{t+1})\overset{\eqref{equ:ProximalGradientEqu2}}{=}\beta h(\bar{x}^{t+\frac{1}{2}})\notag\\
 \overset{(a)}{\leqslant}& \frac{\beta}{n}\sum_{i=1}^n h(x_i^{t+\frac{1}{2}})+\frac{\beta\rho}{2n^2}\sum_{i=1}^{n-1}\sum_{j=i+1}^n{\Vert x_i^{t+\frac{1}{2}}-x_j^{t+\frac{1}{2}} \Vert}^2 \label{equ:HSuccessiveRelation}\\
 \overset{(b)}{\leqslant}& \beta h(\bar{x}^t)-\langle \bar{\nu}^{t}, \bar{x}^{t+1}-\bar{x}^t\rangle+\frac{\alpha}{2n}{\Vert \mathbf{J}\mathbf{y}^{t}-\mathbf{y}^{t}\Vert}^2+\frac{1}{2\alpha n}{\Vert \mathbf{J}\mathbf{x}^{t}-\mathbf{x}^{t}\Vert}^2 -\frac{1}{2\alpha n}{\Vert \mathbf{x}^{t+\frac{1}{2}}-\mathbf{x}^{t}\Vert}^2+\frac{\beta\rho}{n}{\Vert \mathbf{x}^{t+\frac{1}{2}}-\mathbf{J}\mathbf{x}^{t}\Vert}^2\notag\\
 \overset{(c)}{\leqslant}& \beta h(\bar{x}^t)-\langle \bar{\nu}^{t}, \bar{x}^{t+1}-\bar{x}^t\rangle+\frac{\alpha}{2n}{\Vert \mathbf{J}\mathbf{y}^{t}-\mathbf{y}^{t}\Vert}^2+\frac{1+4\alpha\beta\rho}{2\alpha n}{\Vert \mathbf{J}\mathbf{x}^{t}-\mathbf{x}^{t}\Vert}^2-\frac{1-4\alpha\beta\rho}{2\alpha n}{\Vert \mathbf{x}^{t+\frac{1}{2}}-\mathbf{x}^{t}\Vert}^2,\notag
\end{align}
Here, $(a)$ follows from Lemma \ref{lem:WeaklyConvex}.\ref{lem:WeaklyConvexii}. To justify $(b)$, first rewrite the pairwise correction in $(a)$ as a consensus error:
\begin{align*}
&\sum_{i=1}^{n-1}\sum_{j=i+1}^n
\Vert x_i^{t+\frac{1}{2}}-x_j^{t+\frac{1}{2}}\Vert^2 =\frac{1}{2}\sum_{i=1}^n\sum_{j=1}^n
\Vert x_i^{t+\frac{1}{2}}-x_j^{t+\frac{1}{2}}\Vert^2\\
={}&\frac{1}{2}\sum_{i=1}^n\sum_{j=1}^n
\left(\Vert x_i^{t+\frac{1}{2}}\Vert^2
+\Vert x_j^{t+\frac{1}{2}}\Vert^2
-2\left\langle x_i^{t+\frac{1}{2}},x_j^{t+\frac{1}{2}}\right\rangle\right)\\
={}&n\sum_{i=1}^n\Vert x_i^{t+\frac{1}{2}}\Vert^2
-\big\Vert\sum_{i=1}^n x_i^{t+\frac{1}{2}}\big\Vert^2\\
={}&n\sum_{i=1}^n
\Vert x_i^{t+\frac{1}{2}}-\bar{x}^{t+\frac{1}{2}}\Vert^2\\
={}&n\Vert(\mathbf{I}_{nd}-\mathbf{J})
\mathbf{x}^{t+\frac{1}{2}}\Vert^2,
\end{align*}
where $\mathbf{J}\mathbf{x}^{t+\frac{1}{2}}=[\bar{x}^{t+\frac{1}{2}};\ldots;\bar{x}^{t+\frac{1}{2}}]$ yields the final stacked expression. 
Expanding the centered squared norms verifies the penultimate equality:
\begin{align*}
&n\sum_{i=1}^n\Vert x_i^{t+\frac{1}{2}}
-\bar{x}^{t+\frac{1}{2}}\Vert^2=n\sum_{i=1}^n\Vert x_i^{t+\frac{1}{2}}\Vert^2
-2n\Big\langle\sum_{i=1}^n x_i^{t+\frac{1}{2}},
\bar{x}^{t+\frac{1}{2}}\Big\rangle
+n^2\Vert\bar{x}^{t+\frac{1}{2}}\Vert^2\\
={}&n\sum_{i=1}^n\Vert x_i^{t+\frac{1}{2}}\Vert^2
-n^2\Vert\bar{x}^{t+\frac{1}{2}}\Vert^2\\
={}&n\sum_{i=1}^n\Vert x_i^{t+\frac{1}{2}}\Vert^2
-\big\Vert\sum_{i=1}^n x_i^{t+\frac{1}{2}}\big\Vert^2,
\end{align*}
where the last two equalities use
$\sum_{i=1}^n x_i^{t+\frac{1}{2}}=n\bar{x}^{t+\frac{1}{2}}$.
Thus, the pairwise correction in $(a)$ becomes
\begin{equation*}
\frac{\beta\rho}{2n^2}\sum_{i=1}^{n-1}\sum_{j=i+1}^n
\Vert x_i^{t+\frac{1}{2}}-x_j^{t+\frac{1}{2}}\Vert^2
=\frac{\beta\rho}{2n}\Vert(\mathbf{I}_{nd}-\mathbf{J})
\mathbf{x}^{t+\frac{1}{2}}\Vert^2.
\end{equation*}
To compare this correction with the weak-convexity term in the averaged local bound, we use the fact that $\mathbf{J}$ and $\mathbf{I}_{nd}-\mathbf{J}$ are orthogonal projections:
\begin{align*}
&\Vert\mathbf{x}^{t+\frac{1}{2}}-\mathbf{J}\mathbf{x}^{t}\Vert^2 
={}\Vert(\mathbf{I}_{nd}-\mathbf{J})
\mathbf{x}^{t+\frac{1}{2}}+\mathbf{J}(\mathbf{x}^{t+\frac{1}{2}}-\mathbf{x}^{t})\Vert^2  \\
={}&\Vert(\mathbf{I}_{nd}-\mathbf{J})
\mathbf{x}^{t+\frac{1}{2}}\Vert^2 +\Vert\mathbf{J}(\mathbf{x}^{t+\frac{1}{2}}-\mathbf{x}^{t})\Vert^2+2\underbrace{\langle(\mathbf{I}_{nd}-\mathbf{J})\mathbf{x}^{t+\frac{1}{2}}, \mathbf{J}(\mathbf{x}^{t+\frac{1}{2}}-\mathbf{x}^{t})\rangle}_{=0} \\
\geqslant{}&\Vert(\mathbf{I}_{nd}-\mathbf{J})
\mathbf{x}^{t+\frac{1}{2}}\Vert^2.
\end{align*}
Hence, the pairwise correction is bounded as
\begin{equation*}
\frac{\beta\rho}{2n^2}\sum_{i=1}^{n-1}\sum_{j=i+1}^n
\Vert x_i^{t+\frac{1}{2}}-x_j^{t+\frac{1}{2}}\Vert^2
\leqslant\frac{\beta\rho}{2n}\Vert
\mathbf{x}^{t+\frac{1}{2}}-\mathbf{J}\mathbf{x}^{t}\Vert^2.
\end{equation*}
Averaging \eqref{equ:HRelationLocal} with $x=\bar{x}^t$ contributes another
$\frac{\beta\rho}{2n}\Vert\mathbf{x}^{t+\frac{1}{2}}-\mathbf{J}\mathbf{x}^{t}\Vert^2$.
Adding the two terms gives the coefficient $\beta\rho/n$; all other terms in $(b)$ follow directly from the averaged local bound. For $(c)$, we use
\begin{equation*}
    \frac{\beta\rho}{n}{\Vert \mathbf{x}^{t+\frac{1}{2}}-\mathbf{J}\mathbf{x}^{t}\Vert}^2
    \leqslant\frac{2\beta\rho}{n}{\Vert \mathbf{x}^{t+\frac{1}{2}}-\mathbf{x}^{t}\Vert}^2
    +\frac{2\beta\rho}{n}{\Vert \mathbf{J}\mathbf{x}^{t}-\mathbf{x}^{t}\Vert}^2.
\end{equation*}

For the smooth component $f$, $L$-smoothness gives
\begin{equation}\label{equ:FSuccessiveRelation}
    f(\bar{x}^{t+1})\leqslant f(\bar{x}^{t})+\langle\nabla f(\bar{x}^{t}), \bar{x}^{t+1}-\bar{x}^{t}\rangle+\frac{L}{2}{\Vert \bar{x}^{t+1}-\bar{x}^{t}\Vert}^2.
\end{equation}
Combining \eqref{equ:FSuccessiveRelation} with \eqref{equ:HSuccessiveRelation} and substituting \eqref{equ:Adopt} and \eqref{equ:ProximalGradientEqu2} gives
\begin{align*}
\phi(\bar{x}^{t+1})-\phi(\bar{x}^{t})\leqslant&\big\langle  \bar{\nu}^{t}-\nabla f(\bar{x}^{t}), \frac{\alpha}{n}\sum_{i=1}^n G^{\alpha}(x_i^t,y_i^t)\big\rangle-\frac{\alpha-4\alpha^2\beta\rho}{2n}{\Vert \mathbf{G}^{\alpha}(\mathbf{x}^t,\mathbf{y}^{t})\Vert}^2 \\
&+\frac{1+4\alpha\beta\rho}{2\alpha n}{\Vert \mathbf{J}\mathbf{x}^{t}-\mathbf{x}^{t}\Vert}^2+\frac{\alpha}{2n}{\Vert \mathbf{J}\mathbf{y}^{t}-\mathbf{y}^{t}\Vert}^2+\frac{L\alpha^2}{2}{\left\Vert \frac{1}{n}\sum_{i=1}^n G^{\alpha}(x_i^t,y_i^t)\right\Vert}^2 \notag\\
\leqslant&\frac{1+4\alpha\beta\rho}{2\alpha n}{\Vert \mathbf{J}\mathbf{x}^{t}-\mathbf{x}^{t}\Vert}^2+\frac{\alpha}{2n}{\Vert \mathbf{J}\mathbf{y}^{t}-\mathbf{y}^{t}\Vert}^2+\frac{\alpha}{2\eta}{\Vert \nabla f(\bar{x}^{t})-\bar{\nu}^{t}\Vert}^2 \notag\\
&-\frac{\alpha}{4n}{\Vert \mathbf{G}^{\alpha}(\mathbf{x}^t,\mathbf{y}^{t})\Vert}^2-\frac{\alpha-2L\alpha^2-8\alpha^2\beta\rho-2\alpha\eta}{4n}{\Vert \mathbf{G}^{\alpha}(\mathbf{x}^t,\mathbf{y}^{t})\Vert}^2,\notag
\end{align*}
The last inequality uses Jensen's inequality,
${\Vert \frac{1}{n}\sum_{i=1}^n G^{\alpha}(x_i^t,y_i^t)\Vert}^2\leqslant\frac{1}{n}\sum_{i=1}^n{\Vert G^{\alpha}(x_i^t,y_i^t)\Vert}^2=\frac{1}{n}{\Vert \mathbf{G}^{\alpha}(\mathbf{x}^t,\mathbf{y}^{t})\Vert}^2$, together with Lemma \ref{lem:VectorEquation}.\ref{lem:VectorEquation2}.
Set $\eta=\frac{1}{16}$. Since $\alpha L\leqslant\frac{1}{16}$ and $\alpha\beta\rho\leqslant\frac{1}{48}$,
\begin{equation*}
\alpha-2L\alpha^2-8\alpha^2\beta\rho-2\alpha\eta
\geqslant\frac{7\alpha}{12}>\frac{\alpha}{2},
\qquad
\frac{1+4\alpha\beta\rho}{2\alpha n}\leqslant\frac{13}{24\alpha n}.
\end{equation*}
Consequently,
\begin{align}
    \phi(\bar{x}^{t+1})-\phi(\bar{x}^{t})\leqslant&\frac{13}{24\alpha n}{\Vert \mathbf{J}\mathbf{x}^{t}-\mathbf{x}^{t}\Vert}^2+\frac{\alpha}{2n}{\Vert \mathbf{J}\mathbf{y}^{t}-\mathbf{y}^{t}\Vert}^2+{8\alpha}{\Vert \nabla f(\bar{x}^{t})-\bar{\nu}^{t}\Vert}^2 \notag \\
    &-\frac{\alpha}{8n}{\Vert \mathbf{G}^{\alpha}(\mathbf{x}^t,\mathbf{y}^{t})\Vert}^2-\frac{\alpha}{4n}{\Vert \mathbf{G}^{\alpha}(\mathbf{x}^t,\mathbf{y}^{t})\Vert}^2,\label{equ:PhiSuccessiveRelationEuq1}
\end{align}
We bound the last term in \eqref{equ:PhiSuccessiveRelationEuq1} as follows:
\begin{align}
-\frac{\alpha}{4n}{\Vert \mathbf{G}^{\alpha}(\mathbf{x}^t,\mathbf{y}^{t})\Vert}^2\overset{(a)}{\leqslant}&\frac{\alpha}{4n}{\Vert \mathbf{G}^{\alpha}(\mathbf{x}^t,\mathbf{y}^{t})-\mathbf{G}^{\alpha}(\mathbf{x}^t)\Vert}^2-\frac{\alpha}{8n}{\Vert \mathbf{G}^{\alpha}(\mathbf{x}^t)\Vert}^2 \label{equ:PhiSuccessiveRelationEuq2}\\
\overset{\eqref{equ:ProximalGradientRelation}}{\leqslant}&\frac{\alpha}{4n(1-\alpha\beta\rho)^2}\sum_{i=1}^n{\Vert y_i^{t}-\nabla f({x}_i^{t})\Vert}^2-\frac{\alpha}{8n}{\Vert \mathbf{G}^{\alpha}(\mathbf{x}^t)\Vert}^2 \notag\\
 ={}&\frac{\alpha}{4n(1-\alpha\beta\rho)^2}\sum_{i=1}^n{\Vert y_i^{t}-\bar{\nu}^{t}+\bar{\nu}^{t}-\nabla f(\bar{x}^{t})+\nabla f(\bar{x}^{t})-\nabla f({x}_i^{t})\Vert}^2-\frac{\alpha}{8n}{\Vert \mathbf{G}^{\alpha}(\mathbf{x}^t)\Vert}^2 \notag\\
    \overset{(b)}{\leqslant}&\frac{3\alpha L^2}{4n(1-\alpha\beta\rho)^2}{\Vert \mathbf{J}\mathbf{x}^{t}-\mathbf{x}^{t}\Vert}^2+\frac{3\alpha}{4n(1-\alpha\beta\rho)^2}{\Vert \mathbf{J}\mathbf{y}^{t}-\mathbf{y}^{t}\Vert}^2 \notag\\
    &+\frac{3\alpha}{4(1-\alpha\beta\rho)^2}{\Vert \nabla f(\bar{x}^{t})-\bar{\nu}^{t}\Vert}^2-\frac{\alpha}{8n}{\Vert \mathbf{G}^{\alpha}(\mathbf{x}^t)\Vert}^2, \notag
\end{align}
where $(a)$ follows from Lemma \ref{lem:VectorEquation}.\ref{lem:VectorEquation4}, while $(b)$ uses $\Vert a+b+c\Vert^2\leqslant3(\Vert a\Vert^2+\Vert b\Vert^2+\Vert c\Vert^2)$ and the $L$-smoothness of $f=\frac{1}{n}\sum_{i=1}^n f_i$.
To express the remaining average-gradient discrepancy in terms of consensus errors and the gradient-estimation error, we use
\begin{align}
{\Vert \nabla f(\bar{x}^{t})-\bar{\nu}^{t}\Vert}^2\leqslant&2{\Vert \nabla f(\bar{x}^{t})-\overline{\nabla \mathbf{f}}(\mathbf{x}^{t})\Vert}^2+2{\Vert \overline{\nabla \mathbf{f}}(\mathbf{x}^{t})-\bar{\nu}^{t}\Vert}^2 \label{equ:PhiSuccessiveRelationEuq3}\\
=&2{\left\Vert \frac{1}{n}\sum_{i=1}^n\nabla f_i(\bar{x}^{t})-\frac{1}{n}\sum_{i=1}^n\nabla f_i({x}_i^t)\right\Vert}^2+2{\Vert \overline{\nabla \mathbf{f}}(\mathbf{x}^{t})-\bar{\nu}^{t}\Vert}^2 \notag\\
\leqslant&\frac{2L^2}{n}{\Vert \mathbf{J}\mathbf{x}^{t}-\mathbf{x}^{t}\Vert}^2+2{\Vert  \overline{\nabla \mathbf{f}}(\mathbf{x}^{t})-\bar{\nu}^{t}\Vert}^2. \notag
\end{align}
Substituting \eqref{equ:PhiSuccessiveRelationEuq2} and \eqref{equ:PhiSuccessiveRelationEuq3} into \eqref{equ:PhiSuccessiveRelationEuq1} and summing from $t=0$ to $T$, we obtain
\begin{align*}
&\sum_{t=0}^T{\Vert \mathbf{G}^{\alpha}(\mathbf{x}^t)\Vert}^2+\sum_{t=0}^T{\Vert \mathbf{G}^{\alpha}(\mathbf{x}^t,\mathbf{y}^{t})\Vert}^2 \\
\leqslant& \frac{8n}{\alpha}\sum_{t=0}^T[\phi(\bar{x}^{t})-\phi(\bar{x}^{t+1})]+\left(\frac{13}{3\alpha^2}+128L^2+\frac{18L^2}{(1-\alpha\beta\rho)^2}\right)\sum_{t=0}^T{\Vert \mathbf{J}\mathbf{x}^{t}-\mathbf{x}^{t}\Vert}^2\\
&+n\left(128+\frac{12}{(1-\alpha\beta\rho)^2}\right)\sum_{t=0}^T{\Vert \overline{\nabla \mathbf{f}}(\mathbf{x}^{t})-\bar{\nu}^{t}\Vert}^2 +\left(4+\frac{6}{(1-\alpha\beta\rho)^2}\right)\sum_{t=0}^T{\Vert \mathbf{J}\mathbf{y}^{t}-\mathbf{y}^{t}\Vert}^2. 
    \end{align*}
By the definition of $\mathbf{s}$, adding
$L^2\sum_{t=0}^T\Vert\mathbf{J}\mathbf{x}^t-\mathbf{x}^t\Vert^2+n\sum_{t=0}^T\Vert\overline{\nabla\mathbf{f}}(\mathbf{x}^t)-\bar{\nu}^{t}\Vert^2$
to both sides gives the accumulated stationarity measure. The conditions
$\alpha L\leqslant1/16$ and $\alpha\beta\rho\leqslant1/48$ imply
$L^2\leqslant1/(256\alpha^2)$ and $(1-\alpha\beta\rho)^{-2}\leqslant(48/47)^2$. Therefore,
\begin{align*}
\frac{13}{3\alpha^2}+129L^2+\frac{18L^2}{(1-\alpha\beta\rho)^2}
&\leqslant\frac{1}{\alpha^2}\left[\frac{13}{3}+\frac{1}{256}\left(129+18\left(\frac{48}{47}\right)^2\right)\right]
<\frac{5}{\alpha^2},\\
n\left(129+\frac{12}{(1-\alpha\beta\rho)^2}\right)
&\leqslant n\left[129+12\left(\frac{48}{47}\right)^2\right]<142n,\\
4+\frac{6}{(1-\alpha\beta\rho)^2}
&\leqslant4+6\left(\frac{48}{47}\right)^2<11.
\end{align*}
Finally, $\bar{x}^0=x_0$ and Assumption \ref{ass:AssumptionFunction}.\ref{ass:AssumptionFunctioni} give
\begin{equation*}
\sum_{t=0}^T[\phi(\bar{x}^{t})-\phi(\bar{x}^{t+1})]
=\phi(x_0)-\phi(\bar{x}^{T+1})
\leqslant\phi(x_0)-\phi^*.
\end{equation*}
Substituting this telescoping bound and collecting terms yields \eqref{equ:Proposition1}.
\end{IEEEproof}

\section{Proof of Lemma \ref{lem:GradientEstimation}}
Under Assumptions \ref{ass:AssumptionFunction} and
\ref{ass:AssumptionStochastic}, for $\gamma\in[0,1)$ and either momentum
choice, the gradient-estimation error satisfies
\begin{equation*}
\begin{aligned}
\sum_{t=0}^{T}\mathbb E\Vert
\overline{\nabla\mathbf f}(\mathbf x^t)-\bar\nu^t\Vert^2\leqslant\frac{\gamma^2L^2}{n(1-\gamma)^2}
\sum_{t=0}^{T}\mathbb E\Vert\mathbf x^{t+1}-\mathbf x^t\Vert^2
+\left[\frac{1}{1-\gamma}+\frac{5T}{4}(1-\gamma)\right]
\frac{\sigma^2}{nB}.
\end{aligned}
\tag{\ref{equ:PropositionP1}}
\end{equation*}
We prove this common bound separately for the two momentum choices.

\subsection{Polyak Momentum}
\begin{IEEEproof}
For $t\geqslant0$, the Polyak update gives
\begin{align}
&\mathbb{E}{\Vert \overline{\nabla \mathbf{f}}(\mathbf{x}^{t+1})-\bar{\nu}^{t+1}\Vert}^2 \notag\\
={}&\gamma^2\mathbb{E}{\Vert \overline{\nabla \mathbf{f}}(\mathbf{x}^{t+1})-\overline{\nabla \mathbf{f}}(\mathbf{x}^{t})+\overline{\nabla \mathbf{f}}(\mathbf{x}^{t})-\bar{\nu}^{t}\Vert}^2
+(1-\gamma)^2\mathbb{E}{\Vert \overline{\nabla \mathbf{f}}(\mathbf{x}^{t+1})-\bar{g}^{t+1}\Vert}^2 \notag\\
\leqslant{}&\gamma\mathbb{E}{\Vert \overline{\nabla \mathbf{f}}(\mathbf{x}^{t})-\bar{\nu}^{t}\Vert}^2
+\frac{\gamma^2L^2}{n(1-\gamma)}\mathbb{E}{\Vert \mathbf{x}^{t+1}-\mathbf{x}^t\Vert}^2
+\frac{(1-\gamma)^2\sigma^2}{nB}. \label{equ:ProPProofEuq1}
\end{align}
The fresh mini-batch at iteration $t+1$ is conditionally independent of the
preceding iterates, so the cross term involving
$\bar g^{t+1}-\overline{\nabla\mathbf f}(\mathbf x^{t+1})$ vanishes. The
last inequality follows from Lemma
\ref{lem:VectorEquation}.\ref{lem:VectorEquation3}, Assumption
\ref{ass:AssumptionFunction}.\ref{ass:AssumptionFunctionii}, and the
mini-batch variance bound stated after Assumption
\ref{ass:AssumptionStochastic}.

Summing \eqref{equ:ProPProofEuq1} from $t=0$ to $T-1$ and dropping the
nonpositive terminal term gives
\begin{align*}
(1-\gamma)\sum_{t=0}^T\mathbb{E}{\Vert
\overline{\nabla \mathbf{f}}(\mathbf{x}^{t})-\bar{\nu}^{t}\Vert}^2
\leqslant{}\frac{\gamma^2L^2}{n(1-\gamma)}
\sum_{t=0}^{T-1}\mathbb{E}{\Vert \mathbf{x}^{t+1}-\mathbf{x}^t\Vert}^2
+\frac{T(1-\gamma)^2\sigma^2}{nB}
+\mathbb{E}{\Vert\overline{\nabla \mathbf{f}}(\mathbf{x}^{0})-\bar{\nu}^{0}\Vert}^2.
\end{align*}
The initialization $\bar\nu^0=\bar g^0$ implies
$\mathbb{E}\Vert\overline{\nabla\mathbf f}(\mathbf x^0)-\bar\nu^0\Vert^2
\leqslant\sigma^2/(nB)$. Extending the nonnegative model-movement sum to
$t=T$, dividing by $1-\gamma$, and using
$T(1-\gamma)\leqslant\frac{5T}{4}(1-\gamma)$ proves
\eqref{equ:PropositionP1} for Polyak momentum.
\end{IEEEproof}

\subsection{Nesterov Momentum}
\begin{IEEEproof}
The Nesterov case first controls the error of $\bar\mu^t$. For $t\geqslant1$,
conditional unbiasedness of the fresh mini-batch at $\mathbf x^t$ gives
\begin{align}
\mathbb{E}\Vert\overline{\nabla\mathbf f}(\mathbf x^t)-\bar\mu^{t}\Vert^2 ={}&\gamma^2\mathbb{E}\Vert\overline{\nabla\mathbf f}(\mathbf x^t)-\overline{\nabla\mathbf f}(\mathbf x^{t-1})+\overline{\nabla\mathbf f}(\mathbf x^{t-1})-\bar\mu^{t-1}\Vert^2
+(1-\gamma)^2\mathbb{E}\Vert\overline{\nabla\mathbf f}(\mathbf x^t)-\bar g^t\Vert^2 \notag\\
\leqslant{}&\gamma\mathbb{E}\Vert\overline{\nabla\mathbf f}(\mathbf x^{t-1})-\bar\mu^{t-1}\Vert^2
+\frac{\gamma^2L^2}{n(1-\gamma)}\mathbb{E}\Vert\mathbf x^t-\mathbf x^{t-1}\Vert^2
+\frac{(1-\gamma)^2\sigma^2}{nB}. \label{equ:ProPProofEuq2}
\end{align}
Summing \eqref{equ:ProPProofEuq2} from $t=1$ to $T$ and dropping the
nonpositive terminal term yields
\begin{align}
\sum_{t=0}^{T-1}\mathbb{E}\Vert\overline{\nabla\mathbf f}(\mathbf x^t)-\bar\mu^{t}\Vert^2
\leqslant{}\frac{1}{1-\gamma}\mathbb{E}\Vert\overline{\nabla\mathbf f}(\mathbf x^0)-\bar\mu^0\Vert^2 +\frac{\gamma^2L^2}{n(1-\gamma)^2}\sum_{t=1}^{T}\mathbb{E}\Vert\mathbf x^t-\mathbf x^{t-1}\Vert^2
+\frac{T(1-\gamma)\sigma^2}{nB}. \label{equ:ProPProofMuSum}
\end{align}
Next, \eqref{equ:Update2}--\eqref{equ:Update3} imply
$\bar\nu^{t}=\gamma^2\bar\mu^{t-1}+(1-\gamma^2)\bar g^t$. Thus, for
$t\geqslant1$, the same conditional-independence argument gives
\begin{align*}
\mathbb{E}\Vert\overline{\nabla\mathbf f}(\mathbf x^t)-\bar\nu^{t}\Vert^2
\leqslant{}\gamma^3\mathbb{E}\Vert
\overline{\nabla\mathbf f}(\mathbf x^{t-1})-\bar\mu^{t-1}\Vert^2
+\frac{\gamma^4L^2}{n(1-\gamma)}\mathbb{E}\Vert\mathbf x^t-\mathbf x^{t-1}\Vert^2
+\frac{(1-\gamma^2)^2\sigma^2}{nB}.
\end{align*}
Summing this inequality from $t=1$ to $T$, substituting
\eqref{equ:ProPProofMuSum}, and using
$\bar\mu^0=\bar\nu^0=\bar g^0$ give
\begin{align*}
\sum_{t=0}^T\mathbb{E}\Vert\overline{\nabla\mathbf f}(\mathbf x^t)-\bar\nu^{t}\Vert^2\leqslant{}\frac{\gamma^4L^2}{n(1-\gamma)^2}
\sum_{t=1}^{T}\mathbb{E}\Vert\mathbf x^t-\mathbf x^{t-1}\Vert^2
+\left[\frac{1}{1-\gamma}+T(1-\gamma)(1+\gamma-\gamma^2)\right]\frac{\sigma^2}{nB}.
\end{align*}
Here, $1+\gamma^3/(1-\gamma)\leqslant1/(1-\gamma)$ and
$\gamma^3(1-\gamma)+(1-\gamma^2)^2=(1-\gamma)(1+\gamma-\gamma^2)$ collect
the initial and per-iteration variance terms. Finally,
$\gamma^4\leqslant\gamma^2$ and $1+\gamma-\gamma^2\leqslant5/4$ for
$\gamma\in[0,1)$. Extending the model-movement sum by its nonnegative final
term proves \eqref{equ:PropositionP1} for Nesterov momentum.
\end{IEEEproof}

\section{Proof of Lemma \ref{lem:CoupledRelations}}
Suppose that Assumptions \ref{ass:AssumptionFunction},
\ref{ass:AssumptionMixingMatrix}, and \ref{ass:AssumptionStochastic} hold. 
Let $\{(\bm\mu^t,\bm\nu^t,\mathbf x^t,\mathbf y^t)\}_{t\geqslant0}$ be generated by Algorithm \ref{alg1}.
Let $\alpha>0$, $0\leqslant\gamma<1$, and $0\leqslant\alpha\beta\rho<1-\lambda^{1/(2T_0)}$. 
Then
\begin{equation*}
\sum_{t=0}^{T}\Vert\mathbf x^{t+1}-\mathbf x^t\Vert^2
\leqslant\alpha^2\sum_{t=0}^{T}\Vert\mathbf G^\alpha(\mathbf x^t,\mathbf y^t)\Vert^2
+4\sum_{t=0}^{T+1}\Vert\mathbf J\mathbf x^t-\mathbf x^t\Vert^2.
\tag{\ref{equ:PropositionSuccessiveX}}
\end{equation*}
\begin{equation*}
\sum_{t=0}^{T+1}\Vert\mathbf J\mathbf x^t-\mathbf x^t\Vert^2
\leqslant\frac{T_0\alpha^2}{\delta_1}
\sum_{t=0}^{T}\Vert\mathbf J\mathbf y^t-\mathbf y^t\Vert^2.
\tag{\ref{equ:PropositionX}}
\end{equation*}
\begin{equation*}
\sum_{t=0}^{T}\mathbb E\Vert\bm\nu^{t+1}-\bm\nu^t\Vert^2
\leqslant2L^2\sum_{t=0}^{T}\mathbb E\Vert\mathbf x^{t+1}-\mathbf x^t\Vert^2
+\frac{2n(1-\gamma)[1+8(T+1)(1-\gamma)]\sigma^2}{B}.
\tag{\ref{equ:MomentumDifference}}
\end{equation*}
\begin{equation*}
\sum_{t=0}^{T+1}\mathbb E\Vert\mathbf J\mathbf y^t-\mathbf y^t\Vert^2
\leqslant\frac{T_0}{\delta_2}
\sum_{t=0}^{T}\mathbb E\Vert\bm\nu^{t+1}-\bm\nu^t\Vert^2.
\tag{\ref{equ:PropositionY}}
\end{equation*}

We prove the four inequalities separately.
\subsection{Proof of \eqref{equ:PropositionSuccessiveX}}
\begin{IEEEproof}
The successive-iterate relation follows directly from the model update. Because $\mathbf{J}$ is symmetric and idempotent, $\mathbf{J}$ and $\mathbf{I}_{nd}-\mathbf{J}$ are orthogonal projections. Hence, for $t\geqslant 0$,
\begin{align*}
&\Vert \mathbf{x}^{t+1}-\mathbf{x}^{t}\Vert^2=\Vert \mathbf{J}(\mathbf{x}^{t+1}-\mathbf{x}^{t})+(\mathbf{I}_{nd}-\mathbf{J})(\mathbf{x}^{t+1}-\mathbf{x}^{t})\Vert^2 \\
={}&\Vert \mathbf{J}(\mathbf{x}^{t+1}-\mathbf{x}^{t})\Vert^2
+\Vert (\mathbf{I}_{nd}-\mathbf{J})(\mathbf{x}^{t+1}-\mathbf{x}^{t})\Vert^2.
\end{align*}
Moreover, \eqref{equ:Adopt}, \eqref{equ:Combine}, and Lemma \ref{lem:MatrixW}.\ref{lem:MatrixW1} imply that
\begin{equation*}
\mathbf{J}(\mathbf{x}^{t+1}-\mathbf{x}^{t})=\mathbf{J}(\mathbf{x}^{t+\frac{1}{2}}-\mathbf{x}^{t}) =-\alpha\mathbf{J}\mathbf{G}^{\alpha}(\mathbf{x}^{t},\mathbf{y}^{t}).
\end{equation*}
Therefore, using the nonexpansiveness of an orthogonal projection and Lemma \ref{lem:VectorEquation}.\ref{lem:VectorEquation3} with $\tau=1$, we obtain
\begin{align*}
\Vert \mathbf{x}^{t+1}-\mathbf{x}^{t}\Vert^2
\leqslant{}&\alpha^2\Vert \mathbf{G}^{\alpha}(\mathbf{x}^{t},\mathbf{y}^{t})\Vert^2 +2\Vert \mathbf{J}\mathbf{x}^{t+1}-\mathbf{x}^{t+1}\Vert^2+2\Vert \mathbf{J}\mathbf{x}^{t}-\mathbf{x}^{t}\Vert^2.
\end{align*}
Summing from $t=0$ to $T$ gives coefficient $4$ for the interior consensus error terms and coefficient $2$ for the two endpoints. Bounding the endpoint coefficients by $4$ yields \eqref{equ:PropositionSuccessiveX}.
\end{IEEEproof}

\subsection{Proof of \eqref{equ:PropositionX}}
\begin{IEEEproof}
    For $t\geqslant1$ and $1>\lambda>0$, it holds that
\begin{align}
    &{\Vert \mathbf{J}\mathbf{x}^{t}-\mathbf{x}^{t}\Vert}^2\overset{\eqref{equ:Update4}}{=}{\Vert (\mathbf{J}\mathbf{W}^{t-1}-\mathbf{W}^{t-1})\textbf{prox}_{\mathbf h}^{1/(\alpha\beta)}\{\mathbf{x}^{t-1}-\alpha\mathbf{y}^{t-1}\}\Vert}^2 \notag \\
    \overset{(a)}{=}&{\Vert (\mathbf{J}-\mathbf{W}^{t-1})[\textbf{prox}_{\mathbf h}^{1/(\alpha\beta)}\{\mathbf{J}\mathbf{x}^{t-1}-\alpha\mathbf{J}\mathbf{y}^{t-1}\}-\textbf{prox}_{\mathbf h}^{1/(\alpha\beta)}\{\mathbf{x}^{t-1}-\alpha\mathbf{y}^{t-1}\}]\Vert}^2 \notag\\
    \overset{(b)}{\leqslant}&\Vert\mathbf{W}^{t-1}-\mathbf{J}\Vert^2{\Vert \textbf{prox}_{\mathbf h}^{1/(\alpha\beta)}\{\mathbf{J}\mathbf{x}^{t-1}-\alpha\mathbf{J}\mathbf{y}^{t-1}\}-\textbf{prox}_{\mathbf h}^{1/(\alpha\beta)}\{\mathbf{x}^{t-1}-\alpha\mathbf{y}^{t-1}\}\Vert}^2 \notag\\
    \overset{(c)}{\leqslant}& \frac{\Vert\mathbf{W}^{t-1}-\mathbf{J}\Vert^2}{(1-\alpha\beta\rho)^2}{\Vert\mathbf{J}{\mathbf{x}}^{t-1}-\mathbf{x}^{t-1}-\alpha\mathbf{J}\mathbf{y}^{t-1}+\alpha\mathbf{y}^{t-1} \Vert}^2 \notag\\
    \leqslant&\frac{(1+c)\Vert\mathbf{W}^{t-1}-\mathbf{J}\Vert^2}{(1-\alpha\beta\rho)^2}{\Vert\mathbf{J}\mathbf{x}^{t-1}-{\mathbf{x}}^{t-1} \Vert}^2+\frac{(1+c)\alpha^2\Vert\mathbf{W}^{t-1}-\mathbf{J}\Vert^2}{c(1-\alpha\beta\rho)^2}{\Vert\mathbf{J}\mathbf{y}^{t-1}-\mathbf{y}^{t-1} \Vert}^2 \notag\\
    \overset{(d)}{\leqslant}&\lambda^2{\left[\frac{1+c}{(1-\alpha\beta\rho)^2}\right]}^{t-r_tT_0}{\Vert\mathbf{J}\mathbf{x}^{r_tT_0}-{\mathbf{x}}^{r_tT_0} \Vert}^2+\frac{(1+c)\alpha^2}{c(1-\alpha\beta\rho)^2}\sum_{l=r_tT_0+1}^{t}\left[\frac{1+c}{(1-\alpha\beta\rho)^2}\right]^{t-l}{\Vert\mathbf{J}\mathbf{y}^{l-1}-\mathbf{y}^{l-1}\Vert}^2 \notag\\
     \overset{(e)}{\leqslant}&\lambda^2{\left[\frac{1+c}{(1-\alpha\beta\rho)^2}\right]}^{T_0}{\Vert\mathbf{J}\mathbf{x}^{r_tT_0}-{\mathbf{x}}^{r_tT_0} \Vert}^2+\frac{\alpha^2}{c}{\left[\frac{1+c}{(1-\alpha\beta\rho)^2}\right]}^{T_0+1}\sum_{l=r_tT_0+1}^{t}{\Vert\mathbf{J}\mathbf{y}^{l-1}-\mathbf{y}^{l-1}\Vert}^2 \notag \\
     \overset{(f)}{=}&\lambda{\Vert\mathbf{J}\mathbf{x}^{r_tT_0}-{\mathbf{x}}^{r_tT_0} \Vert}^2+\frac{\alpha^2\sum_{l=r_tT_0+1}^{t}{\Vert\mathbf{J}\mathbf{y}^{l-1}-\mathbf{y}^{l-1}\Vert}^2}{\lambda\left[(1-\alpha\beta\rho)^2-\lambda^{\frac{1}{T_0}}\right]},   \label{equ:ProXProofEqu1}
\end{align}
where $(a)$ follows from Lemma \ref{lem:MatrixW}.\ref{lem:MatrixW1} and the fact that $$(\mathbf{J}-\mathbf{W}^{t-1})\textbf{prox}_{\mathbf h}^{1/(\alpha\beta)}\{\mathbf{J}\mathbf{x}^{t-1}-\alpha\mathbf{J}\mathbf{y}^{t-1}\}=\mathbf{0}.$$
 Lemma \ref{lem:MatrixW}.\ref{lem:MatrixW2} and Lemma \ref{lem:WeaklyConvex}.\ref{lem:WeaklyConvexiii} are utilized in $(b)$ and $(c)$, respectively.
Moreover, $(d)$ is derived by iterating the previous equation up to $r_tT_0$, where $r_t=\lfloor (t-1)/T_0\rfloor$, while $(e)$ relies on $T_0\geqslant t-r_tT_0$.
The last equation in $(f)$ is obtained by setting $c=(1-\alpha\beta\rho)^2\lambda^{-\frac{1}{T_0}}-1>0$, which follows from $\alpha\beta\rho<1-\lambda^{\frac{1}{2T_0}}$.
Similarly, we have
\begin{equation}
    {\Vert \mathbf{J}\mathbf{x}^{r_tT_0}-\mathbf{x}^{r_tT_0}\Vert}^2 \leqslant\lambda{\Vert\mathbf{J}\mathbf{x}^{(r_t-1)T_0}-{\mathbf{x}}^{(r_t-1)T_0} \Vert}^2+\frac{\alpha^2\sum_{l=(r_t-1)T_0+1}^{r_tT_0}{\Vert\mathbf{J}\mathbf{y}^{l-1}-\mathbf{y}^{l-1}\Vert}^2}{\lambda\left[(1-\alpha\beta\rho)^2-\lambda^{\frac{1}{T_0}}\right]}. \label{equ:ProXProofEqu2}
\end{equation}
 Then, substituting \eqref{equ:ProXProofEqu2} into \eqref{equ:ProXProofEqu1} and iterating it, we can obtain
 \begin{equation}
    {\Vert \mathbf{J}\mathbf{x}^{t}-\mathbf{x}^{t}\Vert}^2 \leqslant \frac{\alpha^2\sum_{l=1}^{t}\lambda^{\left\lfloor\frac{t-1}{T_0}\right\rfloor-\left\lfloor\frac{l-1}{T_0}\right\rfloor}{\Vert\mathbf{J}\mathbf{y}^{l-1}-\mathbf{y}^{l-1}\Vert}^2}{\lambda\left[(1-\alpha\beta\rho)^2-\lambda^{\frac{1}{T_0}}\right]}, \label{equ:ProXProofEqu3}
 \end{equation}
 The exponent counts the number of complete communication blocks separating iterations $l$ and $t$.
 The original first term ${\Vert \mathbf{J}\mathbf{x}^{0}-\mathbf{x}^{0}\Vert}^2=0$ in \eqref{equ:ProXProofEqu3} is eliminated.
 Next, summing \eqref{equ:ProXProofEqu3} from $t=1$ to $T+1$, we obtain:
 \begin{align*}
    \lambda\left[(1-\alpha\beta\rho)^2-\lambda^{\frac{1}{T_0}}\right]\sum_{t=1}^{T+1}{\Vert \mathbf{J}\mathbf{x}^{t}-\mathbf{x}^{t}\Vert}^2\leqslant&{\alpha^2\sum_{t=1}^{T+1}\sum_{l=1}^{t}\lambda^{\left\lfloor\frac{t-1}{T_0}\right\rfloor-\left\lfloor\frac{l-1}{T_0}\right\rfloor}{\Vert\mathbf{J}\mathbf{y}^{l-1}-\mathbf{y}^{l-1}\Vert}^2} \\
    \overset{(a)}{=}&  {\alpha^2\sum_{l=1}^{T+1}\sum_{t=l}^{T+1}\lambda^{\left\lfloor\frac{t-1}{T_0}\right\rfloor-\left\lfloor\frac{l-1}{T_0}\right\rfloor}{\Vert\mathbf{J}\mathbf{y}^{l-1}-\mathbf{y}^{l-1}\Vert}^2} \\
    \overset{(b)}{\leqslant}& \frac{T_0\alpha^2\sum_{t=0}^{T}{\Vert\mathbf{J}\mathbf{y}^{t}-\mathbf{y}^{t}\Vert}^2}{1-\lambda},\notag
\end{align*}
 where $(a)$ exchanges the order of summation. For each fixed $l$, every nonnegative exponent in the inner sum appears at most $T_0$ times. Thus, $(b)$ follows from $\sum_{k=0}^{+\infty}\lambda^k=(1-\lambda)^{-1}$.
Due to the initial values assigned to $\mathbf{x}^0$ and $\mathbf{y}^0$, we have ${\Vert\mathbf{J}\mathbf{x}^0-\mathbf{x}^0\Vert}^2={\Vert\mathbf{J}\mathbf{y}^0-\mathbf{y}^0\Vert}^2=0$, which further leads to \eqref{equ:PropositionX} for $\lambda\in(0,1)$.

 On the other hand, when $\mathbf{W}=\mathbf{J}$ and $\lambda=0$, the state immediately after the communication at iteration $r_tT_0$ satisfies $\mathbf J\mathbf x^{r_tT_0+1}-\mathbf x^{r_tT_0+1}=\mathbf0$. For the intervening local states $t\geqslant r_tT_0+2$, we have $\mathbf W^{t-1}=\mathbf I_{nd}$, and a derivation similar to that of \eqref{equ:ProXProofEqu1} gives
\begin{align}
    &{\Vert \mathbf{J}\mathbf{x}^{t}-\mathbf{x}^{t}\Vert}^2={\Vert (\mathbf{J}-\mathbf{I}_{nd})\textbf{prox}_{\mathbf h}^{1/(\alpha\beta)}\{\mathbf{x}^{t-1}-\alpha\mathbf{y}^{t-1}\}\Vert}^2 \notag \\
    \leqslant& \frac{1}{(1-\alpha\beta\rho)^2}{\Vert\mathbf{J}{\mathbf{x}}^{t-1}-\mathbf{x}^{t-1}-\alpha\mathbf{J}\mathbf{y}^{t-1}+\alpha\mathbf{y}^{t-1} \Vert}^2\leqslant\frac{1+c}{(1-\alpha\beta\rho)^2}{\Vert\mathbf{J}\mathbf{x}^{t-1}-{\mathbf{x}}^{t-1} \Vert}^2+\frac{(1+c)\alpha^2}{c(1-\alpha\beta\rho)^2}{\Vert\mathbf{J}\mathbf{y}^{t-1}-\mathbf{y}^{t-1} \Vert}^2 \notag\\
    \leqslant&\frac{(1+c)\alpha^2}{c(1-\alpha\beta\rho)^2}\sum_{l=r_tT_0+1}^{t}\left[\frac{1+c}{(1-\alpha\beta\rho)^2}\right]^{t-l}{\Vert\mathbf{J}\mathbf{y}^{l-1}-\mathbf{y}^{l-1}\Vert}^2\leqslant\frac{\alpha^2}{c}{\left[\frac{1+c}{(1-\alpha\beta\rho)^2}\right]}^{T_0+1}\sum_{l=r_tT_0+1}^{t}{\Vert\mathbf{J}\mathbf{y}^{l-1}-\mathbf{y}^{l-1}\Vert}^2 \notag\\
    {=}&\frac{\alpha^2}{(1-\alpha\beta\rho)^{2T_0+2}}{\frac{(1+T_0)^{T_0+1}}{T_0^{T_0}}}\sum_{l=r_tT_0+1}^{t}{\Vert\mathbf{J}\mathbf{y}^{l-1}-\mathbf{y}^{l-1}\Vert}^2,\label{equ:ProXProofEqu4}
\end{align}
where $c=\frac{1}{T_0}$ is substituted in the last equation.
For the communication output $t=r_tT_0+1$, the left-hand side of \eqref{equ:ProXProofEqu4} is zero, so the same upper bound holds trivially. Hence, \eqref{equ:ProXProofEqu4} is valid throughout each proof block.
This choice of $c$ minimizes $c^{-1}{(1+c)^{T_0+1}}$.
Summing \eqref{equ:ProXProofEqu4} from $t=1$ to $T+1$, we have
 \begin{align*}
    \sum_{t=1}^{T+1}{\Vert \mathbf{J}\mathbf{x}^{t}-\mathbf{x}^{t}\Vert}^2
    \leqslant{}&\frac{\alpha^2(1+T_0)^{T_0+1}}
    {T_0^{T_0}(1-\alpha\beta\rho)^{2T_0+2}}
    \sum_{t=1}^{T+1}\sum_{l=r_tT_0+1}^{t}{\Vert\mathbf{J}\mathbf{y}^{l-1}-\mathbf{y}^{l-1}\Vert}^2\\
    \leqslant{}&\frac{\alpha^2(1+T_0)^{T_0+1}}
    {T_0^{T_0-1}(1-\alpha\beta\rho)^{2T_0+2}}
    \sum_{t=0}^{T}{\Vert\mathbf{J}\mathbf{y}^{t}-\mathbf{y}^{t}\Vert}^2.
\end{align*}
which implies that \eqref{equ:PropositionX} holds for $\lambda=0$.
\end{IEEEproof}

\subsection{Proof of \eqref{equ:MomentumDifference}}
\begin{IEEEproof}
At iteration $t$, $\mathbf x^t$ is determined before the fresh mini-batch is sampled. Conditional unbiasedness and the mini-batch variance bound stated after Assumption \ref{ass:AssumptionStochastic} therefore give
\begin{equation*}
\mathbb E[\mathbf g^t-\nabla\mathbf f(\mathbf x^t)\mid\text{the preceding iterates}]=\mathbf0,
\qquad
\mathbb E\Vert\mathbf g^t-\nabla\mathbf f(\mathbf x^t)\Vert^2\leqslant\frac{n\sigma^2}{B}.
\end{equation*}
For $k>l$, condition on all randomness revealed before the mini-batch at iteration $k$ is sampled. The term $\mathbf g^l-\nabla\mathbf f(\mathbf x^l)$ is then fixed, whereas the conditional mean of $\mathbf g^k-\nabla\mathbf f(\mathbf x^k)$ is zero. The tower property therefore gives
\begin{equation*}
\mathbb E\left\langle
\mathbf g^k-\nabla\mathbf f(\mathbf x^k),
\mathbf g^l-\nabla\mathbf f(\mathbf x^l)
\right\rangle=0,\qquad k>l.
\end{equation*}
By symmetry, the same equality holds for $k<l$, so stochastic-gradient errors from different iterations are orthogonal in expectation.

For Polyak momentum, subtracting \eqref{equ:Update1} at two consecutive iterations gives, for $t\geqslant1$,
\begin{align*}
\bm\nu^{t+1}-\bm\nu^t
={}&\gamma(\bm\nu^t-\bm\nu^{t-1})
+(1-\gamma)(\mathbf g^{t+1}-\mathbf g^t)\\
={}&\gamma^t(\bm\nu^1-\bm\nu^0)
+(1-\gamma)\sum_{k=1}^{t}\gamma^{t-k}
(\mathbf g^{k+1}-\mathbf g^k).
\end{align*}
Moreover, the initialization $\bm\nu^0=\mathbf g^0$ implies
\begin{equation*}
\bm\nu^1-\bm\nu^0
=\gamma\bm\nu^0+(1-\gamma)\mathbf g^1-\bm\nu^0
=(1-\gamma)(\mathbf g^1-\mathbf g^0).
\end{equation*}
Substituting this equality into the preceding expansion and including the case $t=0$, we obtain
\begin{equation}\label{equ:MomentumDifferencePolyak}
\bm\nu^{t+1}-\bm\nu^t
=(1-\gamma)\sum_{k=0}^{t}\gamma^{t-k}(\mathbf g^{k+1}-\mathbf g^k).
\end{equation}
Replacing each $\mathbf g^k$ in \eqref{equ:MomentumDifferencePolyak} by $\nabla\mathbf f(\mathbf x^k)$, the resulting term satisfies
\begin{align*}
\sum_{t=0}^{T}\mathbb E\left\Vert
(1-\gamma)\sum_{k=0}^{t}\gamma^{t-k}
[\nabla\mathbf f(\mathbf x^{k+1})-\nabla\mathbf f(\mathbf x^k)]
\right\Vert^2
\leqslant{}&(1-\gamma)\sum_{t=0}^{T}\sum_{k=0}^{t}\gamma^{t-k}
\mathbb E\Vert\nabla\mathbf f(\mathbf x^{k+1})-\nabla\mathbf f(\mathbf x^k)\Vert^2\\
\leqslant{}&L^2(1-\gamma)\sum_{k=0}^{T}\sum_{t=k}^{T}\gamma^{t-k}
\mathbb E\Vert\mathbf x^{k+1}-\mathbf x^k\Vert^2\\
\leqslant{}&L^2\sum_{k=0}^{T}\mathbb E\Vert\mathbf x^{k+1}-\mathbf x^k\Vert^2,
\end{align*}
where the first inequality follows from the convexity of the squared norm and the fact that $(1-\gamma)\sum_{k=0}^{t}\gamma^{t-k}\leqslant1$.
Subtracting the above deterministic-gradient term from \eqref{equ:MomentumDifferencePolyak} and collecting the coefficient of the stochastic-gradient error at each iteration yield
\begin{align*}
&\bm\nu^{t+1}-\bm\nu^t
-(1-\gamma)\sum_{k=0}^{t}\gamma^{t-k}
[\nabla\mathbf f(\mathbf x^{k+1})-\nabla\mathbf f(\mathbf x^k)]\\
={}&(1-\gamma)\sum_{k=0}^{t}\gamma^{t-k}
[\mathbf g^{k+1}-\nabla\mathbf f(\mathbf x^{k+1})]
-(1-\gamma)\sum_{k=0}^{t}\gamma^{t-k}
[\mathbf g^k-\nabla\mathbf f(\mathbf x^k)]\\
={}&(1-\gamma)[\mathbf g^{t+1}-\nabla\mathbf f(\mathbf x^{t+1})]
-(1-\gamma)^2\sum_{k=1}^{t}\gamma^{t-k}
[\mathbf g^k-\nabla\mathbf f(\mathbf x^k)]-(1-\gamma)\gamma^t[\mathbf g^0-\nabla\mathbf f(\mathbf x^0)].
\end{align*}
Indeed, the errors at iterations $t+1$ and $0$ appear only once, whereas the coefficient of the error at iteration $k$, $1\leqslant k\leqslant t$, is
$(1-\gamma)(\gamma^{t-k+1}-\gamma^{t-k})=-(1-\gamma)^2\gamma^{t-k}$.
Consequently, the orthogonality stated above and the mini-batch variance bound stated after Assumption \ref{ass:AssumptionStochastic} imply
\begin{align*}
&\sum_{t=0}^{T}\mathbb E\left\Vert
\bm\nu^{t+1}-\bm\nu^t
-(1-\gamma)\sum_{k=0}^{t}\gamma^{t-k}
[\nabla\mathbf f(\mathbf x^{k+1})-\nabla\mathbf f(\mathbf x^k)]
\right\Vert^2\\
\leqslant{}&\frac{n\sigma^2}{B}\left[(T+1)(1-\gamma)^2
+(1-\gamma)^4\sum_{t=1}^{T}\sum_{k=1}^{t}\gamma^{2(t-k)}
+(1-\gamma)^2\sum_{t=0}^{T}\gamma^{2t}\right]\\
\leqslant{}&\frac{n\sigma^2}{B}\left[(T+1)(1-\gamma)^2
+\frac{T(1-\gamma)^4}{1-\gamma^2}
+\frac{(1-\gamma)^2}{1-\gamma^2}\right]\\
\leqslant{}&\frac{n[\,(1-\gamma)+2(T+1)(1-\gamma)^2\,]\sigma^2}{B}.
\end{align*}
Applying Lemma \ref{lem:VectorEquation}.\ref{lem:VectorEquation3} with $\tau=1$ to the deterministic-gradient and stochastic-gradient terms therefore gives
\begin{align*}
&\sum_{t=0}^{T}\mathbb E\Vert\bm\nu^{t+1}-\bm\nu^t\Vert^2\\
\leqslant{}&2\sum_{t=0}^{T}\mathbb E\big\Vert
(1-\gamma)\sum_{k=0}^{t}\gamma^{t-k}
[\nabla\mathbf f(\mathbf x^{k+1})-\nabla\mathbf f(\mathbf x^k)]
\big\Vert^2+2\sum_{t=0}^{T}\mathbb E\big\Vert
\bm\nu^{t+1}-\bm\nu^t
-(1-\gamma)\sum_{k=0}^{t}\gamma^{t-k}
[\nabla\mathbf f(\mathbf x^{k+1})-\nabla\mathbf f(\mathbf x^k)]
\big\Vert^2\\
\leqslant{}&2L^2\sum_{t=0}^{T}\mathbb E\Vert\mathbf x^{t+1}-\mathbf x^t\Vert^2 +\frac{2n(1-\gamma)[1+2(T+1)(1-\gamma)]\sigma^2}{B},
\end{align*}
which is no larger than \eqref{equ:MomentumDifference}.

For Nesterov momentum, \eqref{equ:Update2} and the initialization $\bm\mu^0=\mathbf g^0$ have the same recursion and initial value as the Polyak update. Hence, the derivation leading to \eqref{equ:MomentumDifferencePolyak} gives
\begin{equation*}
\bm\mu^{t+1}-\bm\mu^t
=(1-\gamma)\sum_{k=0}^{t}\gamma^{t-k}(\mathbf g^{k+1}-\mathbf g^k).
\end{equation*}
Thus, the two estimates derived above also give
\begin{align*}
&\sum_{t=0}^{T}\mathbb E\left\Vert
(1-\gamma)\sum_{k=0}^{t}\gamma^{t-k}
[\nabla\mathbf f(\mathbf x^{k+1})-\nabla\mathbf f(\mathbf x^k)]
\right\Vert^2
\leqslant L^2\sum_{t=0}^{T}\mathbb E\Vert\mathbf x^{t+1}-\mathbf x^t\Vert^2,\\
&\sum_{t=0}^{T}\mathbb E\left\Vert
\bm\mu^{t+1}-\bm\mu^t
-(1-\gamma)\sum_{k=0}^{t}\gamma^{t-k}
[\nabla\mathbf f(\mathbf x^{k+1})-\nabla\mathbf f(\mathbf x^k)]
\right\Vert^2\leqslant
\frac{n[\,(1-\gamma)+2(T+1)(1-\gamma)^2\,]\sigma^2}{B}.
\end{align*}
The initialization $\bm\mu^0=\bm\nu^0=\mathbf g^0$ also implies
$\bm\nu^0=\gamma\bm\mu^0+(1-\gamma)\mathbf g^0$. Therefore, \eqref{equ:Update3} has the same form at $t=0$ as at the subsequent iterations. Subtracting this relation at two consecutive iterations gives
\begin{align*}
\bm\nu^{t+1}-\bm\nu^t
={}&[\gamma\bm\mu^{t+1}+(1-\gamma)\mathbf g^{t+1}]
-[\gamma\bm\mu^t+(1-\gamma)\mathbf g^t]\\
={}&\gamma(\bm\mu^{t+1}-\bm\mu^t)
+(1-\gamma)(\mathbf g^{t+1}-\mathbf g^t).
\end{align*}
By the convexity of the squared norm and $\gamma+(1-\gamma)=1$, its deterministic-gradient component satisfies
\begin{align*}
&\sum_{t=0}^{T}\mathbb E\left\Vert
\gamma(1-\gamma)\sum_{k=0}^{t}\gamma^{t-k}
[\nabla\mathbf f(\mathbf x^{k+1})-\nabla\mathbf f(\mathbf x^k)]
+(1-\gamma)[\nabla\mathbf f(\mathbf x^{t+1})-\nabla\mathbf f(\mathbf x^t)]
\right\Vert^2\\
\leqslant{}&\gamma\sum_{t=0}^{T}\mathbb E\left\Vert
(1-\gamma)\sum_{k=0}^{t}\gamma^{t-k}
[\nabla\mathbf f(\mathbf x^{k+1})-\nabla\mathbf f(\mathbf x^k)]
\right\Vert^2 +(1-\gamma)\sum_{t=0}^{T}\mathbb E\Vert
\nabla\mathbf f(\mathbf x^{t+1})-\nabla\mathbf f(\mathbf x^t)
\Vert^2\\
\leqslant{}&\gamma L^2\sum_{t=0}^{T}\mathbb E\Vert\mathbf x^{t+1}-\mathbf x^t\Vert^2
+(1-\gamma)L^2\sum_{t=0}^{T}\mathbb E\Vert\mathbf x^{t+1}-\mathbf x^t\Vert^2\\
={}&L^2\sum_{t=0}^{T}\mathbb E\Vert\mathbf x^{t+1}-\mathbf x^t\Vert^2.
\end{align*}
Subtracting the deterministic-gradient component from the exact difference relation gives
\begin{align*}
&\bm\nu^{t+1}-\bm\nu^t
-\gamma(1-\gamma)\sum_{k=0}^{t}\gamma^{t-k}
[\nabla\mathbf f(\mathbf x^{k+1})-\nabla\mathbf f(\mathbf x^k)]
-(1-\gamma)[\nabla\mathbf f(\mathbf x^{t+1})-\nabla\mathbf f(\mathbf x^t)]\\
={}&\gamma\Big(\bm\mu^{t+1}-\bm\mu^t
-(1-\gamma)\sum_{k=0}^{t}\gamma^{t-k}
[\nabla\mathbf f(\mathbf x^{k+1})-\nabla\mathbf f(\mathbf x^k)]\Big) +(1-\gamma)\big([\mathbf g^{t+1}-\nabla\mathbf f(\mathbf x^{t+1})]
-[\mathbf g^t-\nabla\mathbf f(\mathbf x^t)]\big).
\end{align*}
For $t=0$, the right-hand side reduces to
\begin{align*}
&(1-\gamma)(1+\gamma)\big([\mathbf g^1-\nabla\mathbf f(\mathbf x^1)]
-[\mathbf g^0-\nabla\mathbf f(\mathbf x^0)]\big),
\end{align*}
whereas, for $t\geqslant1$, collecting the coefficients of the stochastic-gradient errors gives
\begin{align*}
&(1-\gamma)(1+\gamma)[\mathbf g^{t+1}-\nabla\mathbf f(\mathbf x^{t+1})]\\
&\quad -(1-\gamma)[1+\gamma(1-\gamma)]
[\mathbf g^t-\nabla\mathbf f(\mathbf x^t)]\\
&\quad -(1-\gamma)^2\sum_{k=1}^{t-1}\gamma^{t-k+1}
[\mathbf g^k-\nabla\mathbf f(\mathbf x^k)]
-(1-\gamma)\gamma^{t+1}[\mathbf g^0-\nabla\mathbf f(\mathbf x^0)].
\end{align*}
The stochastic-gradient errors at different iterations are orthogonal. Therefore, applying the variance bound to the preceding two expressions and summing over $t$ yields
\begin{align*}
&\sum_{t=0}^{T}\mathbb E\left\Vert
\bm\nu^{t+1}-\bm\nu^t
-\gamma(1-\gamma)\sum_{k=0}^{t}\gamma^{t-k}
[\nabla\mathbf f(\mathbf x^{k+1})-\nabla\mathbf f(\mathbf x^k)]
-(1-\gamma)[\nabla\mathbf f(\mathbf x^{t+1})-\nabla\mathbf f(\mathbf x^t)]
\right\Vert^2\\
\leqslant{}&\frac{n\sigma^2}{B}\Bigg\{
2(1-\gamma)^2(1+\gamma)^2
+T(1-\gamma)^2\big[(1+\gamma)^2+[1+\gamma(1-\gamma)]^2\big] +(1-\gamma)^4\sum_{t=1}^{T}\sum_{k=1}^{t-1}\gamma^{2(t-k+1)}
+(1-\gamma)^2\sum_{t=1}^{T}\gamma^{2(t+1)}\Bigg\}\\
\leqslant{}&\frac{n\sigma^2}{B}\Bigg\{
\frac{(1-\gamma)\gamma^4}{1+\gamma}
+(1-\gamma)^2\Bigg[2(1+\gamma)^2 +T\left((1+\gamma)^2+[1+\gamma(1-\gamma)]^2
+\frac{(1-\gamma)\gamma^4}{1+\gamma}\right)\Bigg]\Bigg\}\\
\leqslant{}&\frac{n(1-\gamma)[1+8(T+1)(1-\gamma)]\sigma^2}{B},
\end{align*}
where the second inequality uses
$\sum_{t=1}^{T}\sum_{k=1}^{t-1}\gamma^{2(t-k+1)}
\leqslant T\gamma^4/(1-\gamma^2)$ and
$\sum_{t=1}^{T}\gamma^{2(t+1)}\leqslant\gamma^4/(1-\gamma^2)$,
and the last one follows from $(1+\gamma)^2\leqslant4$,
$\gamma(1-\gamma)\leqslant1/4$, and
$(1-\gamma)\gamma^4/(1+\gamma)\leqslant1$.
Finally, decomposing $\bm\nu^{t+1}-\bm\nu^t$ into the deterministic-gradient component and the remaining stochastic component, as in the Polyak case, gives
\begin{align*}
\sum_{t=0}^{T}\mathbb E\Vert\bm\nu^{t+1}-\bm\nu^t\Vert^2
\leqslant{}&2L^2\sum_{t=0}^{T}\mathbb E\Vert\mathbf x^{t+1}-\mathbf x^t\Vert^2
+\frac{2n(1-\gamma)[1+8(T+1)(1-\gamma)]\sigma^2}{B},
\end{align*}
which proves \eqref{equ:MomentumDifference} for Nesterov momentum as well.
\end{IEEEproof}

\subsection{Proof of \eqref{equ:PropositionY}}
\begin{IEEEproof}
For $t\geqslant1$ and $1>\lambda>0$, the update of $\mathbf{y}$ gives
\begin{align}
    &\mathbb{E}{\Vert\mathbf{J}\mathbf{y}^{t}-{\mathbf{y}}^{t} \Vert}^2\overset{\eqref{equ:Update5}}{=}\mathbb{E}{\left\Vert(\mathbf{W}^{t-1}-\mathbf{J})\left[\mathbf{y}^{t-1}+\bm\nu^{t}-\bm\nu^{t-1}\right]\right\Vert}^2 \label{equ:ProYProofEqu2}\\
    \overset{(a)}{\leqslant}& (1+c)\Vert\mathbf W^{t-1}-\mathbf J\Vert^2\mathbb{E}{\Vert\mathbf{J}\mathbf{y}^{t-1}-{\mathbf{y}}^{t-1}\Vert}^2+\frac{(1+c)\Vert\mathbf W^{t-1}-\mathbf J\Vert^2}{c}\mathbb{E}{\Vert\bm\nu^{t}-\bm\nu^{t-1} \Vert}^2  \notag\\
    \overset{(b)}{\leqslant}&\lambda^2(1+c)^{t-r_tT_0}\mathbb{E}{\Vert\mathbf{J}\mathbf{y}^{r_tT_0}-{\mathbf{y}}^{r_tT_0} \Vert}^2
    +\frac{1+c}{c}\sum_{l=r_tT_0+1}^{t}(1+c)^{t-l}\mathbb{E}{\Vert\bm\nu^{l}-\bm\nu^{l-1}\Vert}^2 \notag\\
\overset{(c)}{\leqslant}&\lambda^2(1+c)^{T_0}\mathbb{E}{\Vert\mathbf{J}\mathbf{y}^{r_tT_0}-{\mathbf{y}}^{r_tT_0} \Vert}^2
    +\frac{(1+c)^{T_0+1}}{c}\sum_{l=r_tT_0+1}^{t}\mathbb{E}{\Vert\bm\nu^{l}-\bm\nu^{l-1}\Vert}^2 \notag\\
    \overset{(d)}{=}&\lambda\mathbb{E}{\Vert\mathbf{J}\mathbf{y}^{r_tT_0}-{\mathbf{y}}^{r_tT_0} \Vert}^2
    +\frac{\sum_{l=r_tT_0+1}^{t}\mathbb{E}{\Vert\bm\nu^{l}-\bm\nu^{l-1}\Vert}^2}{\lambda(1-\lambda^{\frac{1}{T_0}})},\notag
\end{align}
where $(a)$ is deduced by Lemmas \ref{lem:VectorEquation}.\ref{lem:VectorEquation3}, \ref{lem:MatrixW}.\ref{lem:MatrixW2} and \ref{lem:MatrixW}.\ref{lem:MatrixW3}, $(b)$ is obtained by iterating the previous inequality forward $t-r_tT_0-1$ times, while $(c)$ and $(d)$ are derived from $T_0\geqslant t-r_tT_0$ and setting $c=\lambda^{-\frac{1}{T_0}}-1$, respectively.
The deduction of \eqref{equ:ProYProofEqu2} is similar to that of \eqref{equ:ProXProofEqu1}.
Following the same analysis approach, for $r_t\geqslant1$ we can deduce 
\begin{equation}
\mathbb{E}{\Vert\mathbf{J}\mathbf{y}^{r_tT_0}-{\mathbf{y}}^{r_tT_0} \Vert}^2\leqslant\lambda\mathbb{E}{\Vert\mathbf{J}\mathbf{y}^{(r_t-1)T_0}-{\mathbf{y}}^{(r_t-1)T_0} \Vert}^2+\frac{\sum_{l=(r_t-1)T_0+1}^{r_tT_0}\mathbb{E}{\Vert\bm\nu^{l}-\bm\nu^{l-1}\Vert}^2}{\lambda(1-\lambda^{\frac{1}{T_0}})}.\label{equ:ProYProofEqu3}
\end{equation}
Substituting \eqref{equ:ProYProofEqu3} into \eqref{equ:ProYProofEqu2} and iterating the result, we can obtain
\begin{equation}
    \mathbb{E}{\Vert\mathbf{J}\mathbf{y}^{t}-{\mathbf{y}}^{t} \Vert}^2\leqslant\frac{1}{\lambda(1-\lambda^{\frac{1}{T_0}})}\sum_{l=1}^{t}\lambda^{\left\lfloor\frac{t-1}{T_0}\right\rfloor-\left\lfloor\frac{l-1}{T_0}\right\rfloor}\mathbb{E}{\Vert\bm\nu^{l}-\bm\nu^{l-1}\Vert}^2. \label{equ:ProYProofEqu4}
\end{equation}
The exponent has the same block-counting interpretation as in \eqref{equ:ProXProofEqu3}.
Similar to \eqref{equ:ProXProofEqu2}, the term ${\Vert\mathbf{J}\mathbf{y}^{0}-{\mathbf{y}}^{0}\Vert}^2$ in \eqref{equ:ProYProofEqu4} vanishes because $y_1^0=y_2^0=\cdots=y_n^0=\bar g^0$.
Summing \eqref{equ:ProYProofEqu4} from $t=1$ to $T+1$, we have
\begin{align*}
    \lambda(1-\lambda^{\frac{1}{T_0}})\sum_{t=1}^{T+1}\mathbb{E}{\Vert\mathbf{J}\mathbf{y}^{t}-{\mathbf{y}}^{t} \Vert}^2 \leqslant&\sum_{l=1}^{T+1}\sum_{t=l}^{T+1}\lambda^{\left\lfloor\frac{t-1}{T_0}\right\rfloor-\left\lfloor\frac{l-1}{T_0}\right\rfloor}\mathbb{E}{\Vert\bm\nu^{l}-\bm\nu^{l-1}\Vert}^2 \\
    \leqslant&\frac{T_0}{1-\lambda}\sum_{l=1}^{T+1}\mathbb{E}{\Vert\bm\nu^{l}-\bm\nu^{l-1}\Vert}^2.
\end{align*}
Since $\mathbf{J}\mathbf{y}^0-\mathbf{y}^0=\mathbf{0}$ and $\delta_2=\lambda(1-\lambda)(1-\lambda^{1/T_0})$, this proves \eqref{equ:PropositionY} for $\lambda\in(0,1)$.

 In the case of $\lambda=0$, the state immediately after the communication at iteration $r_tT_0$ satisfies $\mathbf J\mathbf y^{r_tT_0+1}-\mathbf y^{r_tT_0+1}=\mathbf0$. For the intervening local states $t\geqslant r_tT_0+2$, we have the following recursion similar to \eqref{equ:ProYProofEqu2}:
\begin{align}
    &\mathbb{E}{\Vert\mathbf{J}\mathbf{y}^{t}-{\mathbf{y}}^{t} \Vert}^2{=}\mathbb{E}{\Vert(\mathbf{I}_{nd}-\mathbf{J})[\mathbf{y}^{t-1}+\bm\nu^{t}-\bm\nu^{t-1}] \Vert}^2 \notag\\
    {\leqslant}& (1+c)\mathbb{E}{\Vert\mathbf{J}\mathbf{y}^{t-1}-{\mathbf{y}}^{t-1}\Vert}^2+\frac{1+c}{c}\mathbb{E}{\Vert\bm\nu^{t}-\bm\nu^{t-1} \Vert}^2  \notag\\
    {\leqslant}&\frac{1+c}{c}\sum_{l=r_tT_0+1}^{t}(1+c)^{t-l}\mathbb{E}{\Vert\bm\nu^{l}-\bm\nu^{l-1}\Vert}^2 \notag\\
\leqslant&\frac{(1+c)^{T_0+1}}{c}\sum_{l=r_tT_0+1}^{t}\mathbb{E}{\Vert\bm\nu^{l}-\bm\nu^{l-1}\Vert}^2 \notag\\
=&\frac{(1+T_0)^{T_0+1}}{T_0^{T_0}}\sum_{l=r_tT_0+1}^{t}\mathbb{E}{\Vert\bm\nu^{l}-\bm\nu^{l-1}\Vert}^2, \label{equ:ProYProofEqu5} 
\end{align}
where the last equation is also obtained by setting $c=\frac{1}{T_0}$. 
For the communication output $t=r_tT_0+1$, the left-hand side of \eqref{equ:ProYProofEqu5} is zero, and hence the same upper bound holds trivially. Therefore, \eqref{equ:ProYProofEqu5} is valid throughout each proof block.
Summing \eqref{equ:ProYProofEqu5} over the proof blocks gives
\begin{align*}
    \sum_{t=1}^{T+1}\mathbb{E}{\Vert\mathbf{J}\mathbf{y}^{t}-{\mathbf{y}}^{t} \Vert}^2
    \leqslant\frac{(1+T_0)^{T_0+1}}{T_0^{T_0-1}}\sum_{l=1}^{T+1}\mathbb{E}{\Vert\bm\nu^{l}-\bm\nu^{l-1}\Vert}^2,
\end{align*}
which is \eqref{equ:PropositionY} for $\lambda=0$ because $T_0/\delta_2=(1+T_0)^{T_0+1}/T_0^{T_0-1}$.
\end{IEEEproof}

\section{Proof of Lemma \ref{lem:TrackingConsensusBound}}
Suppose that Assumptions \ref{ass:AssumptionFunction},
\ref{ass:AssumptionMixingMatrix}, and \ref{ass:AssumptionStochastic} hold,
$\alpha>0$, $\gamma\in[0,1)$, and either momentum choice is used. If
$0\leqslant\alpha\beta\rho<1-\lambda^{1/(2T_0)}$ and
$\alpha^2<\delta_1\delta_2/(8L^2T_0^2)$, then
\begin{equation*}
\begin{aligned}
\sum_{t=0}^{T+1}\mathbb E\Vert
\mathbf J\mathbf y^t-\mathbf y^t\Vert^2\leqslant
\frac{2T_0\delta_1\alpha^2L^2}
{\delta_1\delta_2-8T_0^2(\alpha L)^2}
\sum_{t=0}^{T}\mathbb E\Vert
\mathbf G^\alpha(\mathbf x^t,\mathbf y^t)\Vert^2+\frac{2nT_0\delta_1(1-\gamma)
[1+8(T+1)(1-\gamma)]\sigma^2}
{B[\delta_1\delta_2-8T_0^2(\alpha L)^2]}.
\end{aligned}
\tag{\ref{equ:PropositionYG1}}
\end{equation*}
\begin{IEEEproof}
Combining \eqref{equ:PropositionY}, \eqref{equ:MomentumDifference}, and \eqref{equ:PropositionSuccessiveX} first gives
\begin{align*}
\sum_{t=0}^{T+1}\mathbb{E}{\Vert \mathbf{J}\mathbf{y}^{t}-\mathbf{y}^{t}\Vert}^2
\leqslant\frac{2T_0\alpha^2L^2}{\delta_2}
\sum_{t=0}^{T}\mathbb{E}{\Vert \mathbf{G}^{\alpha}(\mathbf{x}^{t},\mathbf{y}^{t})\Vert}^2+\frac{8T_0L^2}{\delta_2}
\sum_{t=0}^{T+1}\mathbb{E}{\Vert \mathbf{J}\mathbf{x}^{t}-\mathbf{x}^{t}\Vert}^2
+\frac{2nT_0(1-\gamma)[1+8(T+1)(1-\gamma)]\sigma^2}{B\delta_2}.
\end{align*}
By \eqref{equ:PropositionX},
\begin{equation*}
    \sum_{t=0}^{T+1}\mathbb E{\Vert \mathbf{J}\mathbf{x}^{t}-\mathbf{x}^{t}\Vert}^2\leqslant\frac{T_0\alpha^2}{\delta_1}\sum_{t=0}^{T}\mathbb E{\Vert\mathbf{J}\mathbf{y}^{t}-\mathbf{y}^{t}\Vert}^2\leqslant\frac{T_0\alpha^2}{\delta_1}\sum_{t=0}^{T+1}\mathbb E{\Vert\mathbf{J}\mathbf{y}^{t}-\mathbf{y}^{t}\Vert}^2. 
\end{equation*}
Substituting this inequality into the preceding bound gives
\begin{align*}
&\sum_{t=0}^{T+1}\mathbb{E}{\Vert \mathbf{J}\mathbf{y}^{t}-\mathbf{y}^{t}\Vert}^2
\leqslant\frac{8T_0^2\alpha^2L^2}{\delta_1\delta_2}
\sum_{t=0}^{T+1}\mathbb{E}{\Vert \mathbf{J}\mathbf{y}^{t}-\mathbf{y}^{t}\Vert}^2\\
&\qquad+\frac{2T_0\alpha^2L^2}{\delta_2}
\sum_{t=0}^{T}\mathbb{E}{\Vert \mathbf{G}^{\alpha}(\mathbf{x}^{t},\mathbf{y}^{t})\Vert}^2
+\frac{2nT_0(1-\gamma)[1+8(T+1)(1-\gamma)]\sigma^2}{B\delta_2}.
\end{align*}
The condition $\alpha^2<\frac{\delta_1\delta_2}{8L^2T_0^2}$ makes the coefficient of the consensus error term on the right-hand side strictly smaller than one. Moving that term to the left-hand side and dividing by $1-\frac{8T_0^2\alpha^2L^2}{\delta_1\delta_2}$ yields \eqref{equ:PropositionYG1}.
\end{IEEEproof}

\section{Proof of Theorem \ref{t1}}
\begin{thm}
Suppose that Assumptions \ref{ass:AssumptionFunction}, \ref{ass:AssumptionMixingMatrix} and \ref{ass:AssumptionStochastic} hold. For either Polyak or Nesterov momentum, let
\begin{align*}
0<\alpha\leqslant\min\left\{\frac{1}{16L},\frac{1-\gamma}{24L}\right\},
\qquad \beta>0,
\qquad \alpha\beta\rho\leqslant\frac{1}{48},
\qquad
\alpha\beta\rho<1-\lambda^{\frac{1}{2T_0}},
\end{align*}
and assume that
\begin{equation}\label{equ:TheoremStepsize}
\alpha^2\leqslant
\frac{3\delta_1\delta_2}{8T_0L^2(11\delta_1+9T_0)}.
\end{equation}
Then
\begin{align}
\frac{1}{n(T+1)}\sum_{t=0}^{T}\mathbb E[\mathbf s(\mathbf x^t,\bar\nu^t)]
\leqslant{}&\frac{8[\phi(x_0)-\phi^*]}{\alpha(T+1)}
+\frac{142\left[\frac{1}{1-\gamma}+\frac{5T}{4}(1-\gamma)\right]\sigma^2}{nB(T+1)} \notag\\
&+\frac{2T_0(11\delta_1+6T_0)(1-\gamma)[1+8(T+1)(1-\gamma)]\sigma^2}
{B(T+1)[\delta_1\delta_2-8T_0^2(\alpha L)^2]}.
\label{equ:SupplementTheorem}
\end{align}
\end{thm}
\begin{IEEEproof}
Applying Lemma \ref{lem:GradientEstimation} to Proposition \ref{prop1}, and then
using \eqref{equ:PropositionSuccessiveX} and \eqref{equ:PropositionX} from Lemma \ref{lem:CoupledRelations}, gives
for either momentum choice
\begin{align}
\sum_{t=0}^T\mathbb E[\mathbf s(\mathbf x^t,\bar\nu^t)] \leqslant&-\left[1-\frac{142\alpha^2\gamma^2L^2}{(1-\gamma)^2}\right]
\sum_{t=0}^T\mathbb E\Vert\mathbf G^{\alpha}(\mathbf x^t,\mathbf y^t)\Vert^2 \notag\\
&\qquad+\left[11+\frac{5T_0}{\delta_1}
+\frac{568T_0\alpha^2\gamma^2L^2}{\delta_1(1-\gamma)^2}\right]
\sum_{t=0}^{T+1}\mathbb E\Vert\mathbf J\mathbf y^t-\mathbf y^t\Vert^2 \notag\\
&\qquad+\frac{8n}{\alpha}[\phi(x_0)-\phi^*]
+142\left[\frac{1}{1-\gamma}+\frac{5T}{4}(1-\gamma)\right]\frac{\sigma^2}{B}.
\label{equ:SupplementTheoremProof}
\end{align}

The condition $\alpha L\leqslant(1-\gamma)/24$ implies
\begin{equation*}
\frac{142\alpha^2L^2}{(1-\gamma)^2}<\frac14,
\qquad
\frac{568\alpha^2L^2}{(1-\gamma)^2}<1.
\end{equation*}
Consequently, \eqref{equ:SupplementTheoremProof} is bounded by
\begin{align*}
-\frac34\sum_{t=0}^T\mathbb E\Vert\mathbf G^{\alpha}(\mathbf x^t,\mathbf y^t)\Vert^2
+\left(11+\frac{6T_0}{\delta_1}\right)
\sum_{t=0}^{T+1}\mathbb E\Vert\mathbf J\mathbf y^t-\mathbf y^t\Vert^2 +\frac{8n}{\alpha}[\phi(x_0)-\phi^*]
+142\left[\frac{1}{1-\gamma}+\frac{5T}{4}(1-\gamma)\right]\frac{\sigma^2}{B}.
\end{align*}
Applying Lemma \ref{lem:TrackingConsensusBound}, the coefficient of the accumulated surrogate proximal-gradient term is nonpositive whenever
\begin{equation*}
-\frac34+
\left(11+\frac{6T_0}{\delta_1}\right)
\frac{2T_0\delta_1\alpha^2L^2}
{\delta_1\delta_2-8T_0^2(\alpha L)^2}\leqslant0.
\end{equation*}
After rearrangement, this condition is exactly \eqref{equ:TheoremStepsize}. Dropping the resulting nonpositive term and dividing by $n(T+1)$ proves \eqref{equ:SupplementTheorem}.
\end{IEEEproof}

\begin{cor}\label{cor:SupplementLinear}
Fix $\beta>0$. Let $T+1\geqslant n$, set $B=\lceil\sqrt n\rceil$, and choose
\begin{equation*}
1-\gamma=\sqrt{\frac{n}{T+1}},
\qquad
\alpha=\frac{\sqrt n}{24L\sqrt{T+1}},
\end{equation*}
subject to the conditions of the theorem above. Then, for either Polyak or
Nesterov momentum,
\begin{align}
\frac{1}{n(T+1)}\sum_{t=0}^{T}\mathbb E[\mathbf s(\mathbf x^t,\bar\nu^t)]
\leqslant\frac{192L[\phi(x_0)-\phi^*]}{\sqrt{n(T+1)}}
+\frac{320\sigma^2}{n\sqrt{T+1}} +\frac{T_0(11\delta_1+6T_0)\sigma^2}
{\delta_1\delta_2-\frac{nT_0^2}{72(T+1)}}
\left[\frac{2}{(T+1)^{3/2}}+\frac{16\sqrt n}{T+1}\right].
\label{equ:SupplementLinear}
\end{align}
For fixed problem and network parameters $L$, $\beta$, $\rho$, $T_0$, and
$\lambda<1$, and for sufficiently large $T$ satisfying the theorem
conditions, \eqref{equ:SupplementLinear} is $\mathcal O(1/\sqrt{nT})$ when
$T=\Omega(n^2)$, up to constants depending on these fixed parameters.
Consequently, there exists $\eta\in\{0,\ldots,T\}$ such that
\begin{equation*}
\frac{1}{n}\mathbb E[\mathbf s(\mathbf x^\eta,\bar\nu^\eta)]
=\mathcal O\left(\frac{1}{\sqrt{nT}}\right).
\end{equation*}
In this post-transient regime, expected $\epsilon$-stationarity requires
$\mathcal O(1/(n\epsilon^2))$ iterations.
\end{cor}
\begin{IEEEproof}
Substituting the prescribed $\alpha$ and $\gamma$ into
\eqref{equ:SupplementTheorem} gives
\begin{equation*}
8T_0^2(\alpha L)^2=\frac{nT_0^2}{72(T+1)}.
\end{equation*}
Moreover, $B=\lceil\sqrt n\rceil\geqslant\sqrt n$ and $T\leqslant T+1$.
Applying these two inequalities to the gradient-estimation term gives
\begin{align*}
\frac{142}{nB(T+1)}
\left[\frac{1}{1-\gamma}+\frac{5T}{4}(1-\gamma)\right]\sigma^2\leqslant
\left(\frac{142}{n^2\sqrt{T+1}}
+\frac{355}{2n\sqrt{T+1}}\right)\sigma^2
\leqslant\frac{320\sigma^2}{n\sqrt{T+1}}.
\end{align*}
The two parts of the tracking consensus term are bounded in the same way by
\begin{equation*}
\frac{T_0(11\delta_1+6T_0)\sigma^2}
{\delta_1\delta_2-\frac{nT_0^2}{72(T+1)}}
\left[\frac{2}{(T+1)^{3/2}}
+\frac{16\sqrt n}{T+1}\right].
\end{equation*}
Together with the objective-gap term, these bounds prove
\eqref{equ:SupplementLinear}. For fixed problem and network parameters, its
last term is $\mathcal O(\sqrt n/T)$ and is therefore
$\mathcal O(1/\sqrt{nT})$ once $T=\Omega(n^2)$. Finally, at least one term in
an average is no larger than that average; choosing its index as $\eta$
gives the stated pointwise guarantee. Solving
$1/\sqrt{nT}=\mathcal O(\epsilon)$ gives the iteration complexity.
\end{IEEEproof}

\section{Experimental Details}
This section provides detailed experimental settings and additional results that complement the experiments in the main manuscript.

\subsection{Experiment Settings}
\textbf{Problem.}
The experiments solve \eqref{equ:OriginalProblem}, where the local objective at client $i$ is
\[
f_i(x_i)=\frac{1}{N_i}\sum_{j=1}^{N_i}l\bigl(g(x_i,a_{i,j}),b_{i,j}\bigr).
\]
Here, $g$ denotes the model parameterized by $x_i$, $l$ is the cross-entropy loss, and $N_i$ is the number of training samples at client $i$.
The regularizer $h$ is instantiated as the $\ell_1$ norm, MCP, or SCAD, as specified in the corresponding experiment.

\textbf{Datasets.}
We use A9A, MNIST, Fashion MNIST (FMNIST), and CIFAR-10.
They comprise a tabular binary-classification dataset and image-classification datasets of handwritten digits, clothing items, and natural color images, respectively.
We use the standard training and test splits of these datasets; Table \ref{table1} summarizes their data formats, split sizes, class counts, and input dimensions.

\textbf{Models.}
We evaluate Linear, MLP, CNN, and ResNet-18 models as appropriate for each dataset.
The Linear model contains a single trainable layer. The MLP has three fully connected layers, each followed by a ReLU activation.
The CNN has two convolutional layers, each followed by ReLU activation and max pooling. ResNet-18 uses residual blocks with skip connections.
Table \ref{table2} lists the resulting trainable-parameter counts, which vary with the input dimension and number of classes.

\begin{table}[htbp]
\centering
\caption{An Overview of Employed Datasets.}\label{table1}
\setlength{\tabcolsep}{2pt}
\begin{tabular}{c|cccc}
\toprule
Datasets &  Data Format & Train/Test Samples & Classes & Sizes \\
\midrule
A9A & Tabular data& 32,561/16,281 & 2 & 123 \\
MNIST/FMNIST & Grayscale images& 60,000/10,000 & 10 & 1*28*28 \\
CIFAR-10 & Color images& 50,000/10,000 & 10 & 3*32*32 \\
\bottomrule
\end{tabular}
\end{table}
\begin{table}[htbp]
\centering
 \begin{threeparttable}
\caption{Model Parameter Counts on Datasets}\label{table2}
\begin{tabular}{c|cccc}
\toprule
Models&  Linear & MLP & CNN & ResNet-18    \\
\midrule
A9A &248 &24,258 &--- &--- \\
{MNIST/FMNIST}  & {7850} & {109,386} & {268,362}& {11,175,370}   \\
{CIFAR-10}  & 30,730 & 402,250 & 268,650& 11,181,642   \\

\bottomrule
\end{tabular}
\end{threeparttable}
\end{table}
\begin{figure}[htbp]
  \centering
  \subfloat[Ring]{%
    \begin{tikzpicture}[scale=1.35, every node/.style={circle,draw=blue!65!black,fill=blue!25,minimum size=7pt,inner sep=0pt}]
      \foreach \i [evaluate=\i as \a using 90-36*(\i-1)] in {1,...,10}
        \node (v\i) at (\a:1.25) {};
      \foreach \i/\j in {1/2,2/3,3/4,4/5,5/6,6/7,7/8,8/9,9/10,10/1}
        \draw[blue!70!black,line width=.75pt] (v\i)--(v\j);
    \end{tikzpicture}}\hspace{.025\linewidth}
  \subfloat[2-hop Ring]{%
    \begin{tikzpicture}[scale=1.35, every node/.style={circle,draw=blue!65!black,fill=blue!25,minimum size=7pt,inner sep=0pt}]
      \foreach \i [evaluate=\i as \a using 90-36*(\i-1)] in {1,...,10}
        \node (v\i) at (\a:1.25) {};
      \foreach \i/\j in {1/3,2/4,3/5,4/6,5/7,6/8,7/9,8/10,9/1,10/2}
        \draw[orange!85!black,line width=.65pt] (v\i)--(v\j);
      \foreach \i/\j in {1/2,2/3,3/4,4/5,5/6,6/7,7/8,8/9,9/10,10/1}
        \draw[blue!70!black,line width=.75pt] (v\i)--(v\j);
    \end{tikzpicture}}\hspace{.025\linewidth}
  \subfloat[3-hop Ring]{%
    \begin{tikzpicture}[scale=1.35, every node/.style={circle,draw=blue!65!black,fill=blue!25,minimum size=7pt,inner sep=0pt}]
      \foreach \i [evaluate=\i as \a using 90-36*(\i-1)] in {1,...,10}
        \node (v\i) at (\a:1.25) {};
      \foreach \i/\j in {1/4,2/5,3/6,4/7,5/8,6/9,7/10,8/1,9/2,10/3}
        \draw[green!55!black,line width=.55pt] (v\i)--(v\j);
      \foreach \i/\j in {1/3,2/4,3/5,4/6,5/7,6/8,7/9,8/10,9/1,10/2}
        \draw[orange!85!black,line width=.65pt] (v\i)--(v\j);
      \foreach \i/\j in {1/2,2/3,3/4,4/5,5/6,6/7,7/8,8/9,9/10,10/1}
        \draw[blue!70!black,line width=.75pt] (v\i)--(v\j);
    \end{tikzpicture}}\hspace{.025\linewidth}
  \subfloat[Complete]{%
    \begin{tikzpicture}[scale=1.35, every node/.style={circle,draw=blue!65!black,fill=blue!25,minimum size=7pt,inner sep=0pt}]
      \foreach \i [evaluate=\i as \a using 90-36*(\i-1)] in {1,...,10}
        \node (v\i) at (\a:1.25) {};
      \foreach \i/\j in {1/2,1/3,1/4,1/5,1/6,1/7,1/8,1/9,1/10,2/3,2/4,2/5,2/6,2/7,2/8,2/9,2/10,3/4,3/5,3/6,3/7,3/8,3/9,3/10,4/5,4/6,4/7,4/8,4/9,4/10,5/6,5/7,5/8,5/9,5/10,6/7,6/8,6/9,6/10,7/8,7/9,7/10,8/9,8/10,9/10}
        \draw[blue!70!black,opacity=.28,line width=.45pt] (v\i)--(v\j);
    \end{tikzpicture}}
  \caption{Communication topologies with 10 clients.}\label{fig:graph}
\end{figure}
 \begin{figure}[htbp]
	\centering
\includegraphics[width=0.8\linewidth]{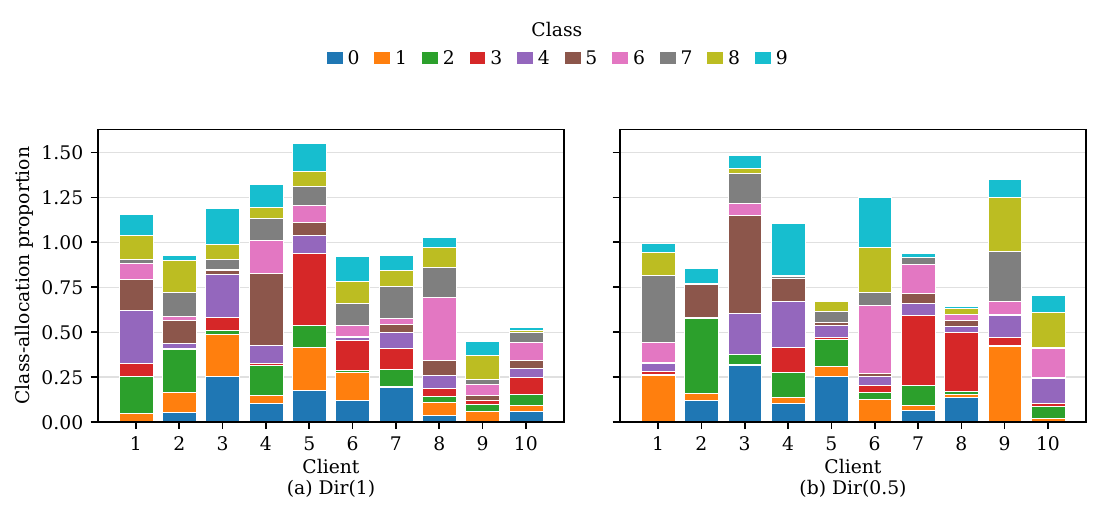}
\vspace{-5pt}
\caption{The proportions of samples corresponding to 10 classes of labels distributed across 10 clients under different Dirichlet distributions. 
The $x$-axis of this bar chart represents each client, while the $y$-axis shows the proportion of samples from each class assigned to that client. 
The colors indicate different labels, and the total proportion for each class across clients sums to 1.
With $Dir(0.5)$ (b), the distributions are more uneven across clients, indicating greater data heterogeneity, whereas $Dir(1)$ (a) produces more balanced distributions.
In the case of uniformly distributed (IID) local data, each bar would reach a height of 1, evenly divided into 10 colors.}\label{fig:dir}
\end{figure}

\textbf{Networks.} We consider Ring, 2-hop Ring, 3-hop Ring, and Complete communication topologies; Fig. \ref{fig:graph} illustrates these graphs for 10 clients.
In a $k$-hop ring, each client communicates with the clients within $k$ hops in either direction.
Thus, the Ring, 2-hop Ring, and 3-hop Ring have degrees 2, 4, and 6, respectively, whereas the Complete topology connects every pair of clients.

\textbf{Partitions.}
For a non-IID partition, the allocation proportions for class $k$ are sampled as $p_k\sim\operatorname{Dir}(\theta\mathbf{1}_n)$, where $p_{ki}$ is the proportion assigned to client $i$.
The concentration parameter $\theta>0$ controls label skew: larger values yield partitions closer to IID, whereas smaller values yield greater heterogeneity.
Fig. \ref{fig:dir} illustrates this partitioning scheme.

\textbf{Baselines.}
For the decentralized topology comparison, we use DEEPSTORM \cite{mancino2023proximal} and Prox-DASA-GT \cite{xiao2023one}.
The additional task-level comparison also includes the centralized FL methods FedMiD \cite{yuan2021federated}, FedDR \cite{tran2021feddr}, and FedADMM \cite{wang2022fedadmm}.

\textbf{Implementation.}
Experiments were run on a platform with an Intel\textregistered\ Xeon\textregistered\ Gold 6242R CPU, 512 GB RAM, and three NVIDIA Tesla P100 GPUs under Ubuntu 20.04.
The software environment used Python 3.8 and PyTorch 1.13.

\subsection{Additional Comparison with Centralized FL Methods}
Table \ref{table:accuracy} compares DEPOSITUM with the centralized FL methods FedMiD \cite{yuan2021federated}, FedDR \cite{tran2021feddr}, and FedADMM \cite{wang2022fedadmm} on MNIST, Fashion-MNIST, and CIFAR-10.
We evaluate MLP, CNN, and ResNet-18 models under IID, $\operatorname{Dir}(1)$, and $\operatorname{Dir}(0.5)$ data partitions, where the smaller Dirichlet concentration represents stronger data heterogeneity.
DEPOSITUM uses 10 clients, a complete topology, a mini-batch size of 32, the SCAD regularizer, stepsize $\alpha=0.1$, regularization weight $\beta=1$, momentum parameter $\gamma=0.2$, and communication period $T_0=5$.
Each client objective is weighted by its proportion of the training samples to account for unequal local dataset sizes.
Each DEPOSITUM run uses 1000 iterations, with one communication phase every five local iterations.
FedMiD, FedDR, and FedADMM use 1000 communication rounds and five local mini-batch updates per client per round.
The final iterate of each method is evaluated on the test set.
All methods use the same fixed client partition, and we report the mean and sample standard deviation over five independent runs.
DEPOSITUM attains the highest mean accuracy in the most settings.

\begin{table}[t]
 \renewcommand\arraystretch{1.27}
		\centering
  \begin{threeparttable}
\caption{Classification accuracy (\%; mean $\pm$ sample standard deviation over five independent runs) for the comparison with centralized FL methods. All methods use the same fixed client partition.
}\label{table:accuracy}
		\setlength{\tabcolsep}{7pt}
		\begin{tabular}{c|c|c|ccccc}
			\toprule
			Dataset & Model & Partition & \shortstack{DEPOSITUM-Polyak} & \shortstack{DEPOSITUM-Nesterov} & FedMiD & FedDR & FedADMM\\
   \midrule
     \multirow{6}{*}{MNIST}& \multirow{3}{*}{MLP}  & IID &$94.20\pm0.13$&\textbf{$94.24\pm0.15$}&$84.65\pm0.45$&$91.13\pm0.15$&$89.49\pm0.11$\\
 & & $\operatorname{Dir}(1)$ &$94.26\pm0.13$&\textbf{$94.29\pm0.12$}&$84.45\pm0.37$&$91.24\pm0.05$&$89.48\pm0.14$\\
 & & $\operatorname{Dir}(0.5)$ &$94.25\pm0.09$&\textbf{$94.29\pm0.08$}&$84.18\pm0.56$&$91.24\pm0.09$&$89.27\pm0.17$\\
 \cline{2-8}
 & \multirow{3}{*}{CNN} & IID &\textbf{$97.99\pm0.11$}&$97.95\pm0.14$&$91.17\pm0.38$&$96.42\pm0.17$&$94.79\pm0.07$\\
 & & $\operatorname{Dir}(1)$ &\textbf{$97.95\pm0.12$}&$97.88\pm0.08$&$90.77\pm0.24$&$96.52\pm0.20$&$94.23\pm0.41$\\
 & & $\operatorname{Dir}(0.5)$ &\textbf{$97.86\pm0.04$}&$97.83\pm0.09$&$90.28\pm0.31$&$96.51\pm0.11$&$93.66\pm0.29$\\
\hline
    \multirow{9}{*}{FMNIST}& \multirow{3}{*}{MLP} & IID &\textbf{$83.28\pm0.17$}&$82.97\pm0.18$&$70.22\pm1.12$&$81.13\pm0.39$&$78.64\pm0.29$\\
 & & $\operatorname{Dir}(1)$ &\textbf{$83.06\pm0.19$}&$82.63\pm0.40$&$70.94\pm0.81$&$80.53\pm0.25$&$78.25\pm0.23$\\
 & & $\operatorname{Dir}(0.5)$ &\textbf{$82.99\pm0.16$}&$82.81\pm0.17$&$70.74\pm0.93$&$80.19\pm0.13$&$77.46\pm0.49$\\
 \cline{2-8}
 & \multirow{3}{*}{CNN} & IID &\textbf{$85.24\pm0.36$}&$85.16\pm0.38$&$74.67\pm0.54$&$82.92\pm0.67$&$81.17\pm0.59$\\
 & & $\operatorname{Dir}(1)$ &$84.69\pm1.06$&\textbf{$84.75\pm0.48$}&$74.35\pm0.43$&$81.62\pm0.34$&$79.89\pm0.71$\\
 & & $\operatorname{Dir}(0.5)$ &$84.46\pm1.47$&\textbf{$84.76\pm0.33$}&$73.44\pm0.62$&$81.13\pm0.47$&$78.78\pm0.59$\\
 \cline{2-8}
 & \multirow{3}{*}{ResNet-18} & IID &\textbf{$89.93\pm0.16$}&$89.67\pm0.18$&$87.55\pm0.08$&$89.83\pm0.27$&$89.25\pm0.31$\\
 & & $\operatorname{Dir}(1)$ &\textbf{$89.54\pm0.27$}&$89.36\pm0.18$&$86.99\pm0.09$&$89.41\pm0.17$&$88.71\pm0.33$\\
 & & $\operatorname{Dir}(0.5)$ &$88.48\pm0.64$&$88.29\pm0.31$&$86.76\pm0.09$&\textbf{$88.94\pm0.42$}&$88.13\pm0.49$\\
\hline
\multirow{6}{*}{CIFAR-10}& \multirow{3}{*}{CNN} & IID &\textbf{$51.92\pm0.69$}&$51.24\pm0.64$&$29.14\pm0.51$&$46.92\pm0.38$&$39.93\pm0.45$\\
 & & $\operatorname{Dir}(1)$ &\textbf{$50.55\pm1.23$}&$50.24\pm1.18$&$29.18\pm1.07$&$42.10\pm0.80$&$37.64\pm1.19$\\
 & & $\operatorname{Dir}(0.5)$ &$49.57\pm2.06$&\textbf{$50.10\pm1.14$}&$27.78\pm1.18$&$40.02\pm1.05$&$35.16\pm2.13$\\
\cline{2-8}
& \multirow{3}{*}{ResNet-18} & IID &\textbf{$65.76\pm1.51$}&$64.87\pm1.40$&$54.17\pm0.66$&$62.59\pm0.64$&$57.37\pm0.73$\\
 & & $\operatorname{Dir}(1)$ &\textbf{$64.05\pm1.20$}&$61.41\pm1.13$&$53.01\pm0.42$&$60.75\pm1.25$&$55.12\pm0.90$\\
& & $\operatorname{Dir}(0.5)$ &$59.13\pm1.74$&$59.24\pm1.16$&$50.44\pm1.12$&\textbf{$59.26\pm0.52$}&$52.26\pm1.67$\\
   \bottomrule
		\end{tabular}
     \end{threeparttable}
	\end{table}

\end{document}